\documentclass[twocolumn]{autart}

\usepackage[T1]{fontenc}

\newcommand{\vect}[1]{\mathbf{#1}}

\newcommand{\diffs}[3]{\frac{\partial^2 #1}{
\ifx#2#3 
\partial #2^2
\else
\partial #2 \partial #3
\fi
}}

\usepackage{amsmath,amssymb,amsfonts}
\usepackage{nicefrac}
\usepackage{cancel}
\usepackage{comment}

\newtheorem{ass}{Assumption}

\newcommand{\colvec}[2][.9]{%
	\scalebox{#1}{%
		\renewcommand{\arraystretch}{1}%
		$\begin{bmatrix}#2\end{bmatrix}$%
	}
}

\usepackage{graphicx}
\usepackage{subfig}
\usepackage{color}
\usepackage{multirow}
\usepackage{cite}
\usepackage{hyperref}
\usepackage{tikz}
\usetikzlibrary{calc}

\graphicspath{{images/}}

\DeclareMathOperator*{\argmin}{argmin}
\newcommand{\revision}[1]{{\textcolor{black}{#1}}}
\newcommand{\revisiontwo}[1]{\textcolor{black}{#1}}

\usepackage{enumitem}

\begin{document}

	\begin{frontmatter}
		
		\title{Coordinated Multi-Robot Trajectory Tracking Control\\ over Sampled Communication\thanksref{footnoteinfo}}
		
		\thanks[footnoteinfo]{This work Partially funded by: the European Commission project H2020 AERIAL-CORE (EC 871479), the ANR, Project ANR-17- CE33-0007 MuRoPhen and by University of Padova grant ``Magic'' SCHE\_SID17\_01.}
		
		\author[mas]{Enrica Rossi}\ead{renrica91@gmail.com},
		\author[inria]{Marco Tognon}\ead{marco.tognon@inria.fr},
		\author[unipd]{Luca Ballotta}\ead{ballotta@dei.unipd.it},
		\author[unipd]{Ruggero Carli}\ead{carlirug@dei.unipd.it},
		\author[laas]{Juan Cort\'es}\ead{juan.cortes@laas.fr},
		\author[laas,twente]{Antonio Franchi}\ead{a.franchi@utwente.nl},
		\author[unipd]{Luca Schenato}\ead{schenato@dei.unipd.it}
		
		\address[mas]{Mas Automazioni S.r.l., Torre di Mosto, Italy}
		\address[inria]{Inria, Universit\'e de Rennes, CNRS, IRISA, Rennes, France}
		\address[unipd]{Department of Information Engineering, University of Padova, Italy}
		\address[laas]{LAAS-CNRS, Universit\'e de Toulouse, CNRS, Toulouse, France}
		\address[twente]{Robotics and Mechatronics lab, University of Twente, Enschede, The Netherlands}
	

\begin{abstract}
	In this paper, 
	we propose an inverse-kinematics controller for a class of multi-robot systems in the scenario of sampled communication. 
	The goal is to make a group of robots perform trajectory tracking in a coordinated way when the sampling time of communications is 
	\revision{much larger than the sampling time of low-level controllers},
	disrupting theoretical convergence guarantees of standard control design \revision{in continuous time}.
	Given a desired trajectory in configuration space \revision{which is pre-computed offline}, 
	the proposed controller receives \revision{configuration} measurements, 
	\revision{possibly via wireless, to re-compute} velocity references for the robots, 
	which are tracked by a low-level controller.
	We propose \revisiontwo{joint design of a} 
	\revision{sampled proportional} feedback plus 
	\revision{a \revisiontwo{novel} continuous-time feedforward
		\revisiontwo{that linearizes the dynamics around the reference trajectory:
		this method} is amenable to 
		distributed communication implementation where only one broadcast transmission is needed per sample.
		Also, 
		we provide closed-form expressions for instability and stability regions 
		and convergence rate in terms of proportional gain $k$ and sampling period $T$.}
	We test the proposed control strategy via numerical simulations in the scenario 
	of cooperative aerial manipulation of a cable-suspended load using a realistic simulator (\emph{Fly-Crane}).
	Finally,
	we compare our proposed controller with centralized approaches that adapt the feedback gain online through smart heuristics,
	and show that it achieves comparable performance.
\end{abstract}

		
	\end{frontmatter}

	\begin{tikzpicture}[remember picture, overlay]
		\node[align=center] at (9,17.4) {%
			\footnotesize This article has been accepted for publication on Automatica.\\%
			Please cite the paper as: E. Rossi, M. Tognon, L. Ballotta, R. Carli, J. Cort\'es, A. Franchi, and L. Schenato,\\%
			"Coordinated Multi-Robot Trajectory Tracking Control over Sampled Communication”,\\%
			Automatica (to appear).};
	\end{tikzpicture}


\vspace{-1cm}
\section{Introduction}\label{sec:intro}
Unmanned Aerial Vehicles (UAVs) are used in the context of mobile robotics to perform surveillance, coverage, exploration, and transportation~\cite{2010-MazKonBerOll}.
Generally speaking, a group of robots allows to improve task performance with respect to (w.r.t.) the single-robot solution. 
Multiple robots can mitigate problems such as limited payload and time of flight~\cite{arai2002advances}. 
However, they require careful consideration of cooperation or coordination strategies to achieve common goals~\cite{erhart2015internal}.
Consider a group of UAVs that transports a load while avoiding obstacles. 
In this case, not only each UAV needs to avoid obstacles,
but the overall multi-robot system should move in a way such that the load avoids obstacles, as well~\cite{2017-CorEge}.

One of the most direct approaches to control multi-robot systems is the centralized kinematic/dynamic inversion~\cite{2013-SreKum,7759262,2013-ManDevRosCor,75,antonelli2006kinematic}. 
Because of its nature, failure of the central unit may cause the whole task to fail.
Distributed and decentralized approaches spread computations across the robots, guaranteeing greater robustness and flexibility w.r.t. a centralized one~\cite{2013-SabSecChoGas}. 
Although such approaches are preferable for multi-robot systems,
they often lack global information
\revision{such as} load state and parameters, 
or the total number of robots. 
This aspect increases the difficulty of the controller design and might even degrade performance.	
Examples of distributed control methods,
\revision{where robots explicitly exchange local data},
can be found for groups of ground~\cite{pereira2004decentralized,2006-DimKyr,2018q-FraPetRiz,ren2020fully},
underwater~\cite{conti2015innovative,simetti2016manipulation},
and aerial robots~\cite{2013-MelShoMicKum}. 
\revision{Conversely,
decentralized control strategies allow for no direct communication among robots,
with examples including ground or aerial manipulators}~\cite{sieber2018human,verginis2018communication}.
\revision{Also},
to reduce communication issues, 
communication-less approaches relying on a leader-follower paradigm were presented for
cooperative transportation and manipulation~\cite{2019l-GabTogPalFra, doi:10.1177/0278364919854131, 2016-WaSch, farivarnejad2018stability,tsiamis2015cooperative}.
In these cases, 
communication is \revision{implicitly given by} forces exerted on the load~\cite{2018h-TogGabPalFra}. 
However, force feedback may be insufficient for precise tracking, because it lacks pose information.
\revision{The latter} can be retrieved by communication among robots, 
e.g., \revision{making them exchange poses}, 
or installing a sensor on the load \revision{to broadcast} its pose.
This setup can also be extended to formation control problems where a group of robots \revision{needs to} complete a task~\cite{7468455, doi:10.1177/0278364912462493}.
For example, 
a common goal \revision{is} mapping or surveillance of an area while robots keep a certain 3D formation in order to, 
e.g.,
\revision{minimize overlaps among their} fields of view~\cite{5959179}.
In this case,
\revision{communication-based approaches let robots exchange group-level measurements},
\revision{such as} relative distances.
 
In applications,
limited bandwidth of wireless channels disrupts the assumption of continuous-measurement feedback.
As so,
specific strategies are needed to deal with sampled communication~\cite{liu2016formation, hayakawa2006formation, stramigioli2005sampled}.
However,
little work is currently available for multi-robot manipulation.

\subsection{Preview of Key Results}

\begin{figure}
	\centering
	\includegraphics[width=0.28\textwidth]{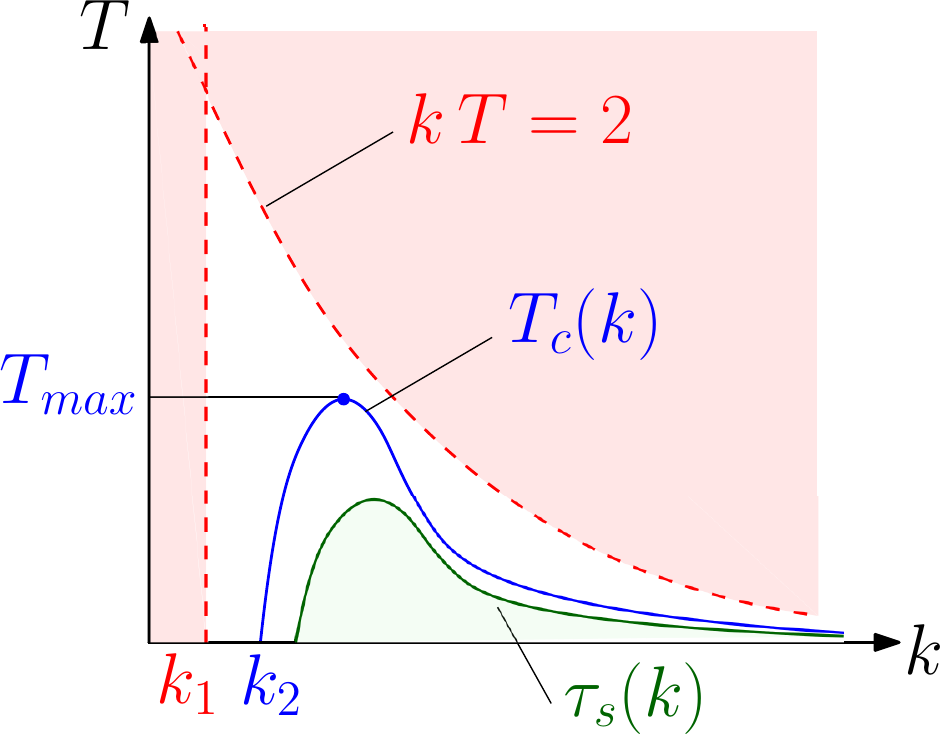}
	\caption{
		We propose a controller for trajectory tracking when measurements are transmitted every $ T $ seconds,
		\revision{with feedback proportional gain $ k $.}
		We show that,
		if $ (k,T) $ \revision{lies outside the dashed red curve},
		the tracking error \revision{is not $ \rho $-monotonically \revisiontwo{contractive}},
		while zero tracking error is guaranteed if $ (k,T) $ \revision{lies below the solid green} curve $ \tau_s(k) $.
		\revision{In particular,
		there exist sampling time $ T_{max} $ such that
		no gain $ k $ can guarantee convergence if $ T > T_{max} $
		and sampling time $ \tau_{CR} $ such that a stabilizing gain exists for any $ T < \tau_{CR} $}.
	}
	\label{fig:tauS_idea}
\end{figure}

In this paper, we aim to design a multi-robot \revision{distributed communication} controller for trajectory tracking 
when wireless communication induces non-negligible sampling of feedback measurements.
In this context,
a \revision{distributed} implementation is preferred because
(i) it reduces the overall communication burden
and related issues such as packet loss or latency and
(ii) it enhances system robustness and scalability.
\revision{Inspired by previous work~\cite{8743454} where point-stabilization was considered, 
	we propose a novel \emph{Sampled communication-aware Inverse-Kinematic controller for Multi-robot systems} (SIKM)
	to address the problem of trajectory tracking under sampled communication.}
Our contributions are summarized as follows.
\revision{
\begin{itemize}[leftmargin=*]
	\item We develop a \revision{distributed} SIKM controller for trajectory tracking that
	receives sampled measurements and \revision{re-computes reference} robot velocities along the trajectory,
	exploiting a \revision{novel continuous-time} feedforward term that \revision{allows} exact trajectory tracking even in the presence of sampled communication.
    \item We show that,
    differently from~\cite{8743454}, 
    there are a maximum sampling time $T_{max}$ and a minimum feedback gain $k_{min}$ 
    beyond which trajectory tracking cannot be achieved,
    as graphically depicted in red in Fig.~\ref{fig:tauS_idea}.
	\item We provide closed-form expressions, 
	whose coefficients can be numerically computed, 
	for the stability region (depicted in green in Fig.~\ref{fig:tauS_idea}) 
	and for the (exponential) convergence rate \revisiontwo{of the trajectory tracking error norm}
	in terms of communication sampling period $T$ and feedback gain $k$. 
	We consider stability in terms of \textit{$ \rho $-monotonic contractiveness},
	which,
	roughly speaking,
	ensures that the trajectory monotonically decreases the tracking error overtime 
	in the absence of external disturbances and is bounded away from singularities.
    \item We validated our strategy by testing the controller on a realistic dynamical simulator
    which replicates with high accuracy the experimental setup available at LAAS-CNRS Lab,
    \textit{Fly-Crane}~\cite{75},
    including dynamical inertial terms, 
    motor actuators, 
    sensor noise,
    and real-time embedded software implementation.
\end{itemize}
}

\subsection{Paper Outline}
In Section~\ref{sec:model_problem_form}
we introduce the class of considered multi-robot systems,
provide the kinematic model (Section~\ref{subsec:kin_model}),
and give \revision{an example} of real system (Section~\ref{subsec:examples}). 
In Section~\ref{sec:prob-formulation}
\revision{we review control architectures for trajectory tracking,
and present our proposed SIKM controller in Section~\ref{sec:proposed-controller}.
In Section~\ref{sec:stability-limitations-SIKM}
we derive fundamental stability limitations in terms of feedback gain and sampling time.
In Section~\ref{sec:stab_and_conv_rate}
we compute an upper bound for the convergence rate,
outline a numerical procedure to estimate it from data (Section~\ref{sec:param_estim}),
and explicitly find controller parameters that yield the fastest convergence (Section~\ref{sec:optimal-param}).}
In Section~\ref{sec:simulations}
we test our controller on a realistic simulator of the \textit{Fly-Crane},
showing that it outperforms \revision{standard designs under sampled communication}.
Final remarks are drawn in Section~\ref{sec:conclusion}.

\section{System Model and Problem Formulation}\label{sec:model_problem_form}

\subsection{Kinematics of Multi-Robot Systems}\label{subsec:kin_model}
In this section, 
we describe the kinematic model of a multi-robot system composed of $N$ robots that 
exchange state information with a common \textit{pivot}, 
labeled as $V$,
to fulfill a task.
As so, 
robots \revision{need not} communicate among themselves,
\revision{but they implicitly coordinate their motions by communicating with the pivot}.
The pivot may be an object to be manipulated, a robot to be escorted, or a vehicle in the space.
The kinematic model is given by
\begin{align*}
	\colvec{	\vect{p}_1 \;\, \dots \;\, \vect{p}_N
	}^\top = \vect{h(q)}=  \colvec{
		\vect{h}^{(1)}(\vect{q}_{1}, \vect{q}_V) \;\, \dots \,\; \vect{h}^{(N)}(\vect{q}_{N}, \vect{q}_V)}^\top.
\end{align*}
This function maps the Lagrangian coordinates of the system 
$\vect{q} = [\vect{q}_{1}^\top \, \cdots \, \vect{q}_{N}^\top\ \vect{q}_{V}^\top]^\top\in\mathbb{R}^{m}$ 
to the vector collecting the robots configurations $\vect{p} = \colvec{	\vect{p}_1^\top \;\, \dots \;\, \vect{p}_N^\top	}^\top \in \mathbb{R}^{n}$.
In particular, 
$\vect{p}_i\in\mathbb{R}^{n_i}$ is the position of the $i$th robot in space, 
$\vect{q}_i\in\mathbb{R}^{m_i}$ gathers the angles and/or distances between the pivot $V$ and the $i$th robot, 
and $\vect{q}_V\in\mathbb{R}^{m_V}$ represents the pose (position and orientation) of the pivot.
Notice that $m=\sum_{i=1}^N{m_i}+m_V$ and $n=\sum_{i=1}^N{n_i}$.
The differential kinematics of the system is
\begin{align}
	\dot{\vect{p}}&=\vect{A_q} \, \, \dot{\vect{q}}, \label{eq:jacobian1}
\end{align}
where the Jacobian $\vect{A_q} = \frac{\partial \vect{h(q)}}{\partial \vect{q}} \in \mathbb{R}^{n \times m}$ has structure
\begin{align}
	\vect{A_q} =
	\colvec{
		\begin{array}{ccc|c}
			\vect{A}^{(1)}_{\vect{q}_{1}} &  & \text{\fontsize{5mmm}{5mm}\selectfont$\vect{0}$} &  \vect{A}^{(1)}_{\vect{q}_{V}}\\
			& \ddots & &  \vdots\\
			\text{\fontsize{5mmm}{5mm}\selectfont$\vect{0}$} & & \vect{A}^{(N)}_{\vect{q}_{N}} &   \vect{A}^{(N)}_{\vect{q}_{V}}
		\end{array}
	}
	\label{eq:jacobian},
\end{align}
$ \vect{A}^{(i)}_{\vect{q}_{i}} =   \frac{\partial \vect{h}^{(i)}(\vect{q}_i,\vect{q}_V)}{\partial \vect{q}_i} \in \mathbb{R}^{n_i \times m_i}$ and 
$\vect{A}^{(i)}_{\vect{q}_{V}} =   \frac{\partial \vect{h}^{(i)}(\vect{q}_i,\vect{q}_V)}{\partial \vect{q}_V} \in \mathbb{R}^{n_i \times m_V}$.
We focus on the case $n=m$, corresponding to \emph{square systems}.
Let us make an example to justify this choice.
Consider the multi-robot system in Fig.~\ref{fig:ex_multi_robot_033},
\revision{where} robots are linked to a platform though rigid cables,
and assume that $\vect{A_q}$ is invertible and \revision{that
desired robot velocities $\vect{\dot{p}}^d$ are assigned \textit{a priori}}.
\revision{In this case},
there always exists a vector $\vect{\dot{q}}=\vect{A}^{-1}_{\vect{q}}\vect{\dot{p}}^d$ in 
configuration space that allows the robots to achieve the desired velocity.
\revision{The next section expands this example more in details.}
If $n \neq m$, 
the Jacobian is not square.
In particular, 
the system is redundant if $n>m$.
If $n<m$, 
there exist trajectories $\vect{\dot{q}}$ in the configuration space which are infeasible for any input $\vect{\dot{p}}^d$.
Such cases require
a dedicated analysis which goes beyond the scope of this paper.
We refer to the preprint~\cite{2021arXiv211201107R} for such an analysis,
that, 
with some attention, 
allows to integrate the control strategy studied here into more general systems.

\revision{
	Furthermore, 
	we consider a cascade control architecture where we design the proposed SIKM controller at the kinematic level.
	The latter provides the desired motor velocities which are tracked by low-level dynamics controllers 
	(see Fig.~\ref{fig:scheme_global_local_planner}),
	assuming that these have larger closed-loop bandwidth than the SIKM.
	This assumption is indeed validated by extensive simulations of the proposed architecture on a full dynamical model
	(including low-level dynamics as well as external disturbances) with a realistic simulator of the testbed \textit{Fly-Crane}.
}

\subsection{Examples of Square Systems}
\label{subsec:examples}

\begin{figure}[t]
	\centering 
	\subfloat[][\emph{$(N_{1c},N_{2c}, N)=(0,3,3)$}\label{fig:ex_multi_robot_033}] {\includegraphics[width=.22\textwidth]{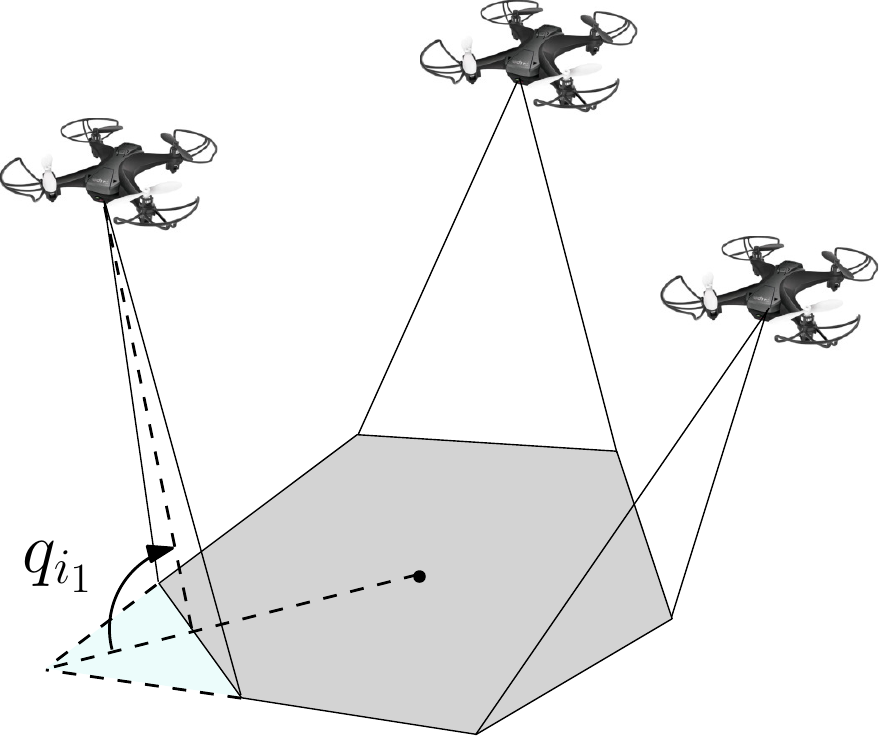}} \quad
	\subfloat[][\emph{$(N_{1c},N_{2c}, N)=(2,2,4)$}\label{fig:ex_multi_robot_224}] {\includegraphics[width=.22\textwidth]{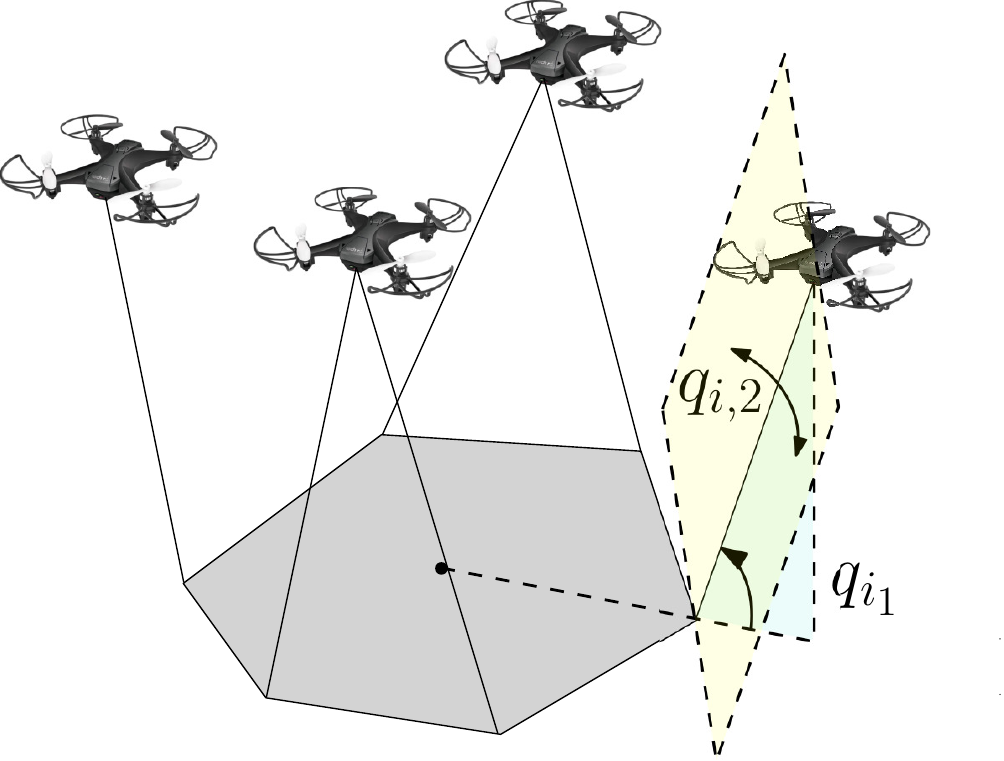}} \\
	\subfloat[][\emph{$(N_{1c},N_{2c}, N)=(4,1,5)$}\label{fig:ex_multi_robot_415}] {\includegraphics[width=.22\textwidth]{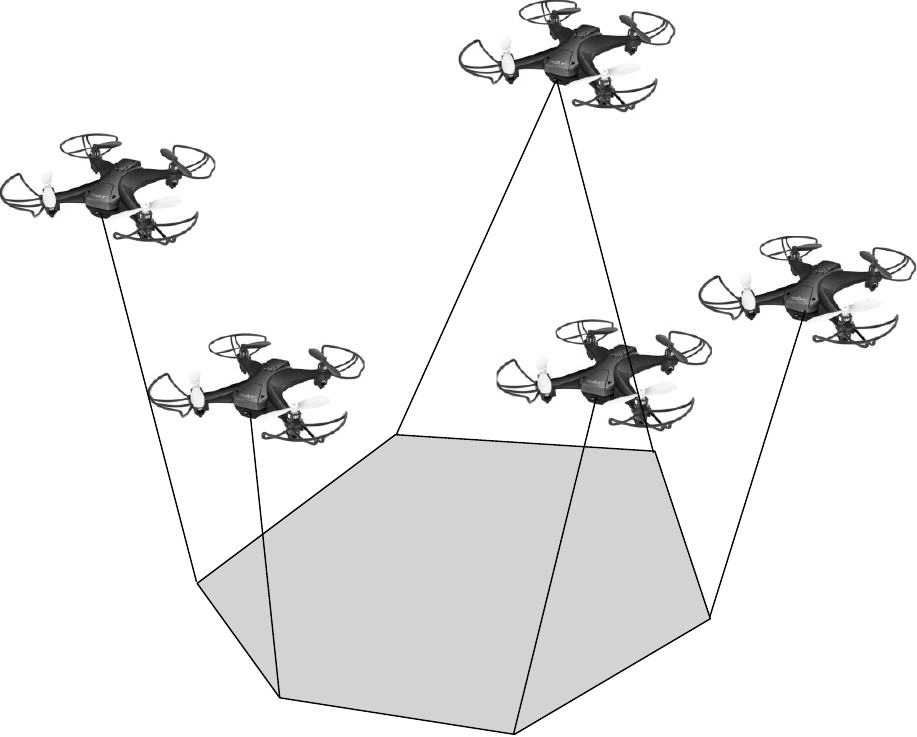}} 
	\quad
	\subfloat[][\emph{$(N_{1c},N_{2c}, N)=(6,0,6)$}\label{fig:ex_multi_robot_606}] {\includegraphics[width=.22\textwidth]{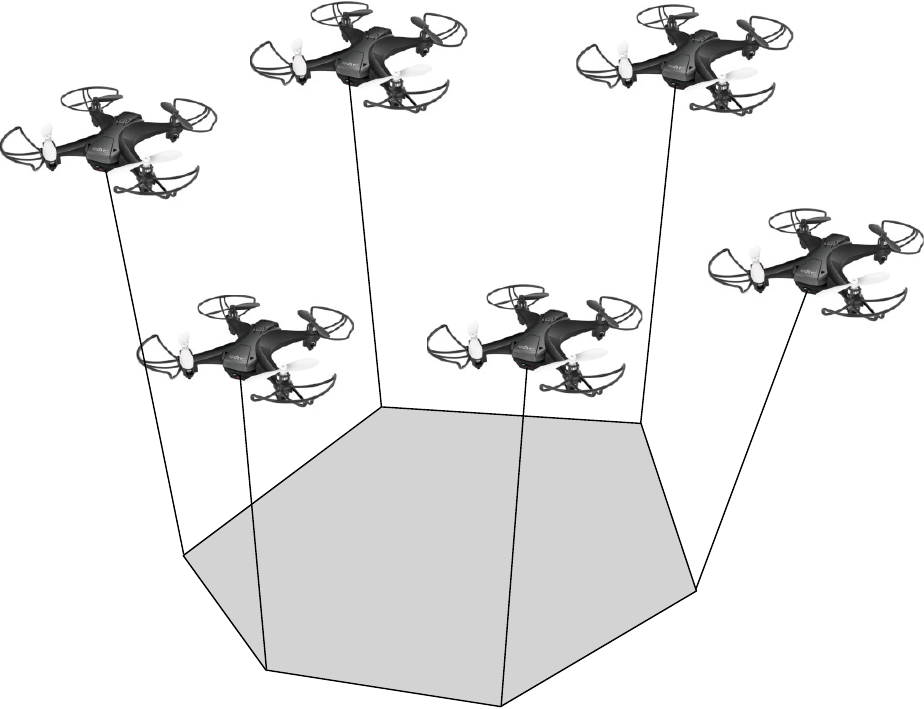}}
	\caption{Square systems where a common object is manipulated by a group of UAVs. All the possible combinations of $(N_{1c},N_{2c}, N)$ introduced in Sec.~\ref{subsec:examples} are represented.} 
	\label{fig:subfig}
\end{figure}

In the literature, we can find several examples of square systems. 
One is the multi-robot system depicted in Fig.~\ref{fig:ex_multi_robot_033} (called \textit{Fly-Crane}~\cite{2013-ManDevRosCor}), where three UAVs transport a common platform.
Each robot is linked to the load through two rigid cables.
The generalized coordinates are chosen as $\vect{q}=[q_1 \,\, q_2 \,\, q_3 \,\, \vect{q}_V^\top]^\top$. 
The robot positions $\vect{p}_i \, \in \mathbb{R}^{3}$ are collected in the vector $\vect{p}=[\vect{p}_1^\top \,\, \vect{p}_2^\top \,\, \vect{p}_3^\top]^\top$. 
Thus, the velocity vectors $\vect{\dot{q}}, \, \vect{\dot{p}}\in \mathbb{R}^{9}$ have the same dimension and $\vect{A_q} \, \in \mathbb{R}^{9\times 9}$ is a square matrix. 
However, this is a particular case of a larger class: different square systems can be obtained by simply changing the number of robots transporting the platform or the number of cables linking each robot to it. 
Notice that if one cable was used instead of two, as shown in Fig.~\ref{fig:ex_multi_robot_224}, then  $\vect{q}_i=[q_{i_1} \ q_{i_2}]^\top \in \mathbb{R}^2$ because each cable can move in two directions (assuming that movements about the cable axis are not allowed). 
In this case, the system is no more guaranteed to be square.
In particular, it holds $n=3N$ and $m=6+2N_{1c}+N_{2c}$ where $N_{1c}\geq 0$ is the number of robots linked to the load through one cable and $N_{2c}\geq 0$ indicates the number of robots linked through two cables. 
We have that $N=N_{1c}+N_{2c}$ and, for a square system $n=m$, it must be $3N=6+2N_{1c}+N_{2c}$. From these relations it turns out that $N_{2c} \leq 3$, $N_{1c} \leq 6$ and $N \leq 6$; in particular, the possible configurations  ($N_{1c}$,$N_{2c}$,$N$) are: (0,3,3) in Fig.~\ref{fig:ex_multi_robot_033}, (2,2,4) in Fig.~\ref{fig:ex_multi_robot_224}, (4,1,5) in Fig.~\ref{fig:ex_multi_robot_415}, (6,0,6) in Fig.~\ref{fig:ex_multi_robot_606}.


\begin{figure}[t]
	\centering
	\includegraphics[width=0.6\columnwidth]{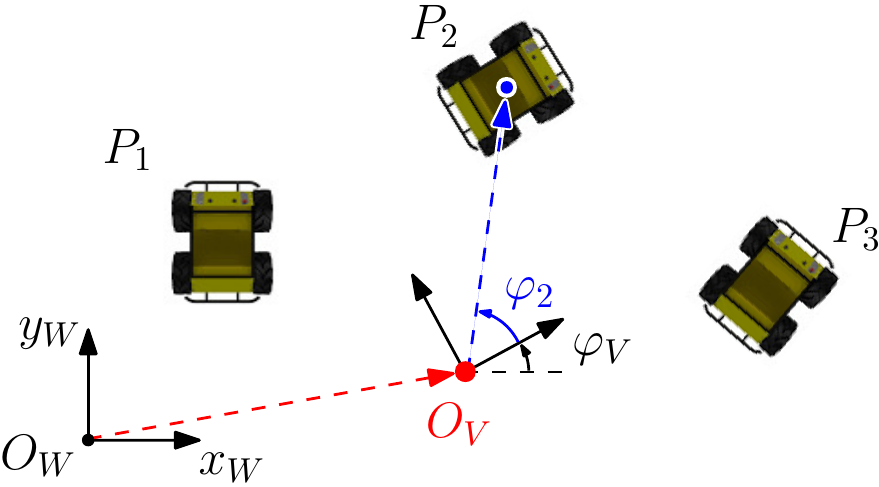}
	\caption{Representation of a multi-robot system where three robots must keep the same orientation $\varphi_i$ w.r.t. the reference frame centered in the point $O_c$ during the assigned task.}
	\label{fig:formation_ex}
\end{figure}
Other examples of square systems are found in the context of formation control problems \cite{doi:10.1177/0278364912462493}, as the one represented in Fig.~\ref{fig:formation_ex}.
Three ground robots move in a 2-dimensional space while respecting some constraints w.r.t. the frame $\mathcal{F}_V = \{O_V, \vect{x}_V,\vect{y}_V\}$ representing the pivot (e.g., the barycenter of the system).
The position $P_i$ of each robot, defined by the vector $\vect{p}_i = [x_i \;\, y_i]^\top$, is described as a function of: the pivot position $\vect{p}_V = [x_V \,\; y_V]^\top$, the distance $d_i$ from $P_i$ to $O_V$, the angle $\varphi_i$ between the line $O_V-P_i$ and the axis $\vect{x}_V$, and the angle $\varphi_V$ between $\vect{x}_V$ and $\vect{x}_W$ of the world reference frame $\mathcal{F}_W = \{O_W, \vect{x}_W,\vect{y}_W\}$.
It is trivial to express the position $P_i$ w.r.t. $\mathcal{F}_W$:
$$
\vect{p}_i 
=\vect{h}(\vect{q}_i,\vect{q}_V),
$$
where $\vect{q}_i=[d_i \ \varphi_i]^\top \in \mathbb{R}^2$, $\vect{q}_V=[x_V \ y_V \ \varphi_V]^\top \in \mathbb{R}^3$ and
$$
\vect{h}(\vect{q}_i,\vect{q}_V)  = 
\begin{bmatrix}
	x_V \\ y_V
\end{bmatrix} + 
\begin{bmatrix}
	\cos(\varphi_V) \ \ -\sin(\varphi_V) \\
	\sin(\varphi_V) \ \ \cos(\varphi_V)
\end{bmatrix}
\begin{bmatrix}
	d_i \, \cos(\varphi_i) \\
	d_i \, \sin(\varphi_i)
\end{bmatrix}.
$$
Let us define the vector of robot positions as $\vect{p}=[\vect{p}_1^\top \ \vect{p}_2^\top \ \vect{p}_3^\top]^\top  \in \mathbb{R}^{6 }$ and the vector of generalized coordinates as $\vect{q} = [\vect{q}_1^\top \ \vect{q}_2^\top \ \vect{q}_3^\top \ \vect{q}_V^\top]^\top   \in  \mathbb{R}^{9 }$. Hence, we can write $\vect{h(q)}\overset{\text{def}}{=} [\vect{h}(\vect{q}_1,\vect{q}_V)^\top \ \vect{h}(\vect{q}_2,\vect{q}_V)^\top \ \vect{h}(\vect{q}_3,\vect{q}_V)^\top]^\top$.
We assume that the robots must perform a task while keeping the angles $\varphi_i$ constant w.r.t. $\mathcal{F}_V$, hence the differential kinematic model is:
\begin{equation*}
	\vect{\dot{p}} =
	\vect{A_q}
	\vect{\dot{q}}',
\end{equation*}
where $\vect{\dot{q}}' =[\dot{\vect{q}}_V^\top \,\; \dot{d}_1 \,\; \dot{d}_2 \,\; \dot{d}_3 ]^\top$ and
\begin{align*}
	\vect{A_q} = 
	\left[\frac{\partial \vect{h}(\vect{q})}{\partial \vect{q}_V} \;\, \frac{\partial \vect{h}(\vect{q})}{\partial d_1} \;\, \frac{\partial \vect{h}(\vect{q})}{\partial d_2} \;\, \frac{\partial \vect{h}(\vect{q})}{\partial d_3} \right] \, \in \, \mathbb{R}^{6\times 6},
\end{align*}
is a square matrix, hence this system is square, too. 

The previous examples show that the considered class of square systems includes many relevant applications. 

\section{\revision{Control Architectures}}
\label{sec:prob-formulation}
\begin{figure}
	\centering
	\begin{minipage}{\linewidth}
		\centering
		\subfloat
		[\revision{Decentralized (no communication) control~\eqref{eq:decentralized}}.]
		{\qquad\quad\includegraphics[width=.6\linewidth]{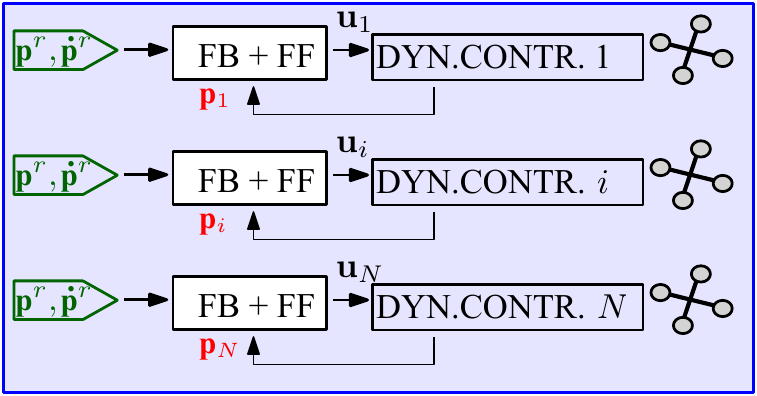}\qquad\quad
			\label{fig:scheme_global_local_planner:decentralized}}
	\end{minipage}%
	\\
	\vspace{5mm}
	\begin{minipage}{\linewidth}
		\centering
		\subfloat
		[\revision{Distributed communication control~\eqref{eqn:sparse}}.]
		{\includegraphics[width=.7\linewidth]{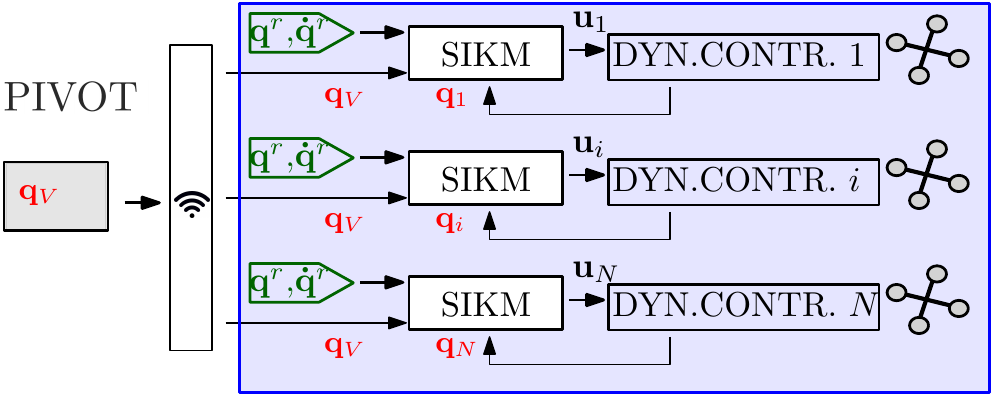}
			\label{fig:scheme_global_local_planner:sparse}}
	\end{minipage}%
	\\
	\vspace{5mm}
	\begin{minipage}{\linewidth}
		\centering
		\subfloat
		[\revision{Centralized communication control~\eqref{eqn:centralized}}.]
		{\includegraphics[width=.7\linewidth]{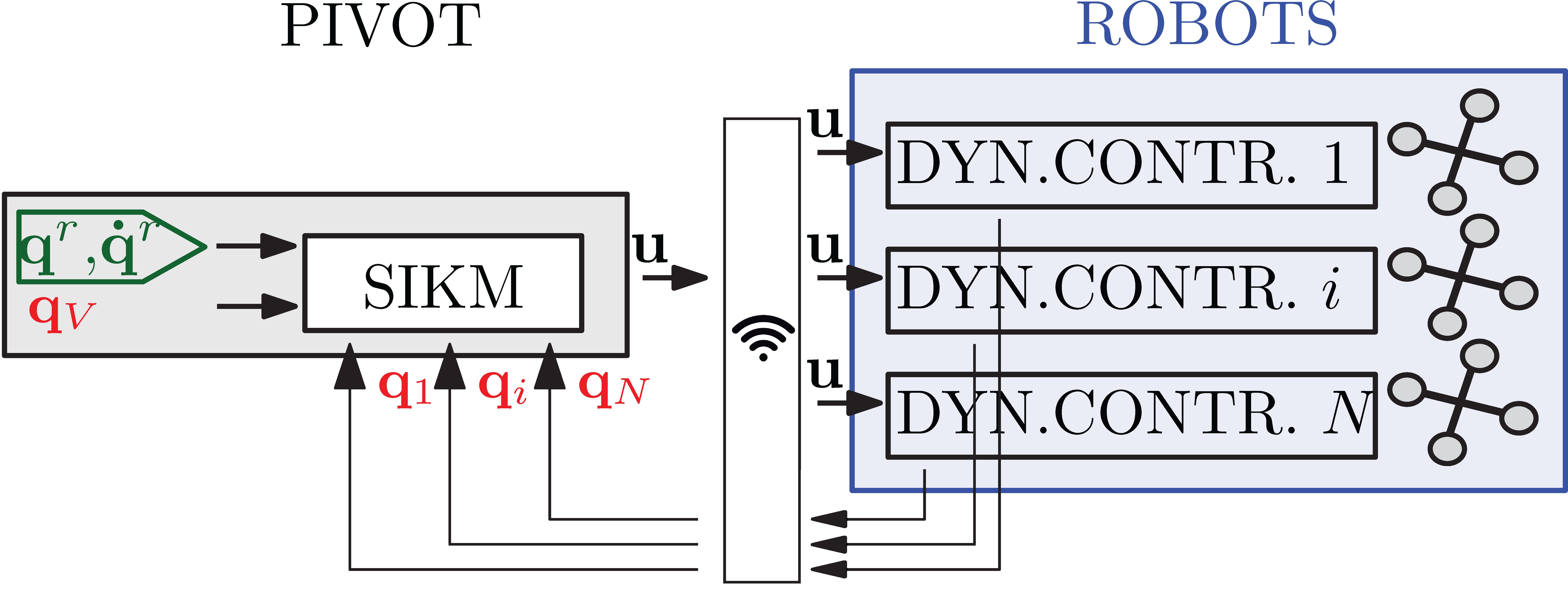}
			\label{fig:scheme_global_local_planner:central}}
	\end{minipage}
	\caption{\revision{Controller architectures for trajectory tracking}.
		The pivot is colored in gray, each robot (equipped with a dynamical controller which converts $\vect{u}$ to forces) in blue,
		sensor measurements in red,
		and the \revision{reference} trajectory in green.
		The wireless symbol refers to sampled communication.}
	\label{fig:scheme_global_local_planner}
\end{figure}
We consider a tracking problem where a multi-robot system is required to follow a sequence of desired configurations
assigned \textit{a priori}.
We assume that such reference trajectory is generated \revision{offline} by a \revision{high-level} planner that takes into account
goals such as obstacle avoidance, singular points, and energy minimization.
Also, we assume that robots are equipped with dynamical controllers sufficiently fast w.r.t.
the dynamics of the system,
such that (s.t.) their velocities are fully controllable,
\begin{align}
	\dot{\vect{p}}(t) = \vect{u}(t), \label{eq:controllability}
\end{align}
where $\mathbf{u}=[\mathbf{u}_1^\top\cdots \mathbf{u}_N^\top]^\top\in\mathbb{R}^{n}$.
\revision{We now enumerate possible control architectures to achieve trajectory tracking.}

\revision{\textbf{Decentralized (no communication) control.}
A commonly adopted architecture is the following \textit{fully decentralized} controller,
\begin{equation}\label{eq:decentralized}
	\mathbf{u}_i(t) = \kappa_i^\text{dec}\left(\vect{p}_i(t);\vect{p}_i^r(t),\vect{\dot{p}}_i^r(t)\right).
\end{equation}
In this case,
each robot needs only local position measurements $ \vect{p}_i(t) $ (no configuration variables $ \vect{q}(t) $ are needed)
to implement feedback and follow its reference trajectory $ (\vect{p}_i^r(t),\vect{\dot{p}}_i^r(t)) $,
namely,
controller~\eqref{eq:decentralized} is \textit{communication-less}
and the control design reduces to a distributed planning problem (Fig.~\ref{fig:scheme_global_local_planner:decentralized}). A typical example is \mbox{$\mathbf{u}_i(t) = -k(\vect{p}_i(t)-\vect{p}_i^r(t))+\vect{\dot{p}}_i^r(t)$}.
However,
lacking communication and coordination,
strategy~\eqref{eq:decentralized} is not robust against disturbances or robot failures/biases.}

\revision{\textbf{Distributed communication control.}
A distributed controller reads
\begin{equation}\label{eqn:sparse}
	\mathbf{u}_i(t) =\kappa_i^\text{distr} (\vect{q}_i(t),\vect{q}_V(\revision{hT});\vect{q}^r_i(t),\vect{q}^r_V(t),\vect{\dot{q}}^r_i(t),\vect{\dot{q}}^r_V(t)),
\end{equation}
\revision{where $ t \in [hT,(h+1)T) $, $ h\in\mathbb{N}  $.}
Specifically,
$\mathbf{u}_i(t) $ depends on the robot's own configuration $ \vect{q}_i(t) $,
which can be measured at all times,
and on the load configuration $ \vect{q}_V(hT) $,
which is transmitted via wireless and available at discrete time instants $ hT $,
$ T $ being the sampling time.
Also,
reference trajectories of robot $ (\vect{q}_i^r(t),\vect{\dot{q}}_i^r(t)) $ 
and load $ (\vect{q}_V^r(t),\vect{\dot{q}}_V^r(t)) $
are continuously available as they are computed offline and pre-stored on robots.
Hence,
controller~\eqref{eqn:sparse} implements a \textit{hybrid continuous-sampled} control
that includes both continuous-time and discrete-time signals,
inducing more challenging design (Fig.~\ref{fig:scheme_global_local_planner:sparse}).}

\revision{\textbf{Centralized control.}
Lastly,
the centralized controller
\begin{equation}\label{eqn:centralized}
	\mathbf{u}_i(t) =\kappa_i^\text{centr} (\vect{q}(hT);\vect{q}^r(t), \vect{\dot{q}}^r(t))
\end{equation}
depends on the full system configuration vector $\vect{q}(hT)$.
Figure~\ref{fig:scheme_global_local_planner:central} shows a possible implementation
where the controller,
located on the pivot,
receives all measurements $\vect{q}(t)$
and broadcasts the control input $\vect{u}(t)$ to the robots via wireless. 
In this case,
while performance is theoretically maximized,
all-to-all communication burden may cause issues
through limitations of wireless communication in terms of bandwidth and reliability.}

\subsection{\revision{Proposed SIKM Architectures}}\label{sec:proposed-controller}

\revision{
In this paper,
we focus on the two communication-based approaches.
While these are attractive by virtue of robustness properties,
their design gets challenging when shifting from continuous-time
(typically assumed in the literature)
to hybrid continuous- and discrete-time dynamics.
In particular,
the presence of sampled measurements in~\eqref{eqn:sparse}--\eqref{eqn:centralized}
makes both controller design and stability analysis nontrivial.}

\revision{
Indeed,
for continuous-time systems,
the controller~\cite{2009-SicSciVilOri}
\begin{equation}\label{eq:FF-continuous}
	\vect{u}(t) = \underbrace{-k\vect{A}_{\vect{q}(t)}(\vect{q}(t)-\vect{q}^r)}_{:=\vect{u}_{k}(t)}
				+ \underbrace{\vect{A}_{\vect{q}(t)} \vect{\dot{q}}^r(t)}_{:=\vect{u}_{\text{ff}}(t)}
\end{equation}
drives the system configuration $\vect{q}(t)$ to the desired reference trajectory
$\vect{q}^r(t)$ exponentially fast, 
whereby
the tracking error $\vect{e}(t)\overset{\text{def}}{=}\vect{q}(t)-\vect{q}^r(t)$
obeys dynamics
\begin{equation*} 
	\vect{\dot{e}}(t)=\vect{\dot{q}}(t)-\vect{\dot{q}}^r(t) = -k\big(\vect{q}(t)-\vect{q}^r(t)\big)=-k \vect{e}(t).
\end{equation*}
Moreover,
in light of~\eqref{eq:jacobian},
the centralized controller~\eqref{eq:FF-continuous} can be implemented with a distributed architecture,
\begin{align}\label{eq:decentr_cont_time}
	\begin{split}
		\vect{u}_i(t) = 
		&-k \vect{A}^{(i)}_{\vect{q}_i(t)} \left( \vect{q}_i(t)\!-\!\vect{q}_i^r(t)\right) +\vect{A}^{(i)}_{\vect{q}_i(t)} \vect{\dot{q}}_i^r(t)\\
		&-k\vect{A}^{(i)}_{\vect{q}_V(t)} \left( \vect{q}_V(t)\!-\!\vect{q}_V^r(t)\right) + \vect{A}^{(i)}_{\vect{q}_V(t)} \vect{\dot{q}}_V^r(t).
	\end{split}
\end{align}
}

\revision{
When measurements are sampled,
suitably modifying~\eqref{eq:FF-continuous} is nontrivial.
In~\cite{8743454},
the authors proposed the following feedback controller for point-stabilization,
\begin{equation}\label{eq:u_k}
\vect{u}(t)=\vect{u}_{k}(hT)=-k\vect{A}_{\vect{q}(hT)}(\vect{q}(hT)-\vect{q}^r),
\end{equation}
for $t \in \left[hT, \,(h+1)T \right)$.
A common strategy for trajectory tracking  in robotic applications is sampling~\eqref{eq:u_k},
\begin{equation}\label{eq:u_k_hT}
	\vect{u}(t)=\vect{u}_{k}(hT)=-k\vect{A}_{\vect{q}(hT)}(\vect{q}(hT)-\vect{q}^r(hT)),
\end{equation}
which however cannot guarantee perfect tracking under nominal conditions
because it lacks feedforward correction
(see~Appendix~\ref{app:point-stabilization} for details).
}

\revision{A naive attempt to improve~\eqref{eq:u_k_hT} is sampling~\eqref{eq:FF-continuous},
\begin{align}\label{eq:u_sampling_contr_law}
	\begin{split}
		\vect{u}(hT+\tau) =
		&-k\vect{A}_{\vect{q}(hT)} (\vect{q}(hT)-\vect{q}^r(hT)) \\
		&+\vect{A}_{\vect{q}(hT)}\vect{\dot{q}}^r(hT+\tau),\quad \tau \in[0,T),
	\end{split}
\end{align}
which results in
the following error dynamics,
\begin{align*}
	&\vect{\dot{e}}(hT+\tau)
	=-k \vect{A}^{-1}_{\vect{e}(hT+\tau)+\vect{q}^r(hT+\tau)} \vect{A}_{\vect{e}(hT)+\vect{q}^r(hT)}\vect{e}(hT) +\\
	&+ \big(\vect{A}^{-1}_{\vect{e}(hT+\tau)+\vect{q}^r(hT+\tau)} \vect{A}_{\vect{e}(hT)+\vect{q}^r(hT)}-\vect{I}\big)\vect{\dot{q}}^r(hT+\tau).
\end{align*}
It is indeed easy to show that,
under controller~\eqref{eq:u_sampling_contr_law},
reference $ \vect{q}^r(t) $ is not an equilibrium trajectory,
making also the latter attempt not suitable for trajectory tracking.}

\begin{prop}\label{prop:no_eq_sampled}	
	Under control strategy~\eqref{eq:u_sampling_contr_law}, 
	the reference trajectory is not an equilibrium trajectory, i.e.,
	$$ \vect{q}(t)=	\vect{q}^r(t) \nRightarrow \vect{\dot e}(t)=0 , \forall t\geq 0.$$
\end{prop}	
\begin{pf}
	If $ \vect{q}(t)=\vect{q}^r(t)$, then $ \vect{e}(t)=0 $ and
	$$\vect{\dot{e}}(hT+\tau) = \big(\vect{A}^{-1}_{\vect{q}^r(hT+\tau)} \vect{A}_{\vect{q}^r(hT)}-\vect{I}\big)\vect{\dot{q}}^r(hT+\tau),$$
	which is nonzero for all $h\in \mathbb{N}$ and $\tau\in (0,T)$ 
	unless $\vect{q}^r(t)\equiv\vect{q}^r$, i.e., the reference trajectory is constant.\qed
\end{pf}

\revision{Finally,
we modify controller~\eqref{eq:u_sampling_contr_law} as follows,
\begin{align} \label{eq:new_control_law}
	\begin{split}
		\vect{u}(hT+\tau) = 
		&-k\vect{A}_{\vect{q}(hT)} (\vect{q}(hT)-\vect{q}^r(hT)) \\
		&+\vect{A}_{\vect{q}^r(hT+\tau)}\vect{\dot{q}}^r(hT+\tau),\quad \tau \in[0,T).
	\end{split}
\end{align}
Differently from \eqref{eq:u_sampling_contr_law},
the Jacobian of the feedforward term in~\eqref{eq:new_control_law}
is computed at $\vect{q}^r(hT+\tau)$ instead of $\vect{q}(hT)$.
Accordingly,
the error dynamics become
\begin{align} \label{eqn:error_flow}
	\!\!\!\!\!\!	\vect{\dot{e}}(hT\!+\!\tau) &=-k \vect{A}^{-1}_{\vect{e}(hT\!+\!\tau)\!+\!\vect{q}^r(hT\!+\!\tau)} \vect{A}_{\vect{e}(hT)\!+\!\vect{q}^r(hT)}\vect{e}(hT) +\nonumber\\
	&+\! \!\big(\!\vect{A}^{-1}_{\vect{e}(hT+\tau)\!+\!\vect{q}^r(hT\!+\!\tau)} \vect{A}_{\vect{q}^r(hT\!+\!\tau)}\!\!-\!\vect{I}\big)\vect{\dot{q}}^r(hT\!\!+\!\tau)\nonumber \\
	&\overset{\mathrm{def}}{=} \vect{f}(\vect{e}(hT\!+\!\tau), \vect{q}^r(hT\!+\!\tau), \vect{\dot{q}}^r(hT\!+\!\tau)),
\end{align}
whereby the reference can be shown to be an equilibrium trajectory.
\begin{prop}\label{prop:equilib_traj}
	Under control strategy~\eqref{eq:new_control_law} the reference trajectory $\vect{q}^r(t)$ is an equilibrium trajectory, i.e.,
	$$ \vect{q}(t)=	\vect{q}^r(t) \Rightarrow \vect{ \dot e}(t)=0 , \forall t\geq 0.$$
\end{prop}
\begin{pf}
	If $\vect{q}(t)=\vect{q}^r(t)$, then $\vect{e}(t)\equiv0$ and
	\begin{equation*}
		\vect{\dot{e}}(hT\!+\!\tau) =
		\big({\!\vect{A}^{-1}_{\!\vect{q}^r(hT\!+\!\tau)} \vect{A}_{\vect{q}^r(hT\!+\!\tau)}\!\!-\!\vect{I}}\big)\vect{\dot{q}}^r(hT\!\!+\!\tau) \equiv0.\hspace{-.6cm}\qed
	\end{equation*}
\end{pf}
Also,
by virtue of the assumed structure~\eqref{eq:jacobian} of the Jacobian,
controller~\eqref{eq:new_control_law} can be decoupled
and is thus amenable of a distributed communication implementation of the form~\eqref{eqn:sparse},
\begin{align}\label{eq:decentr_disc_time}
	\begin{split}
		\hspace{-0.3cm}\vect{u}_i(hT+\tau)= 
		&-k \vect{A}^{(i)}_{\vect{q}_i(hT)} \left( \vect{q}_i(hT)\!-\!\vect{q}_i^r(hT)\right) \\
		&-k\vect{A}^{(i)}_{\vect{q}_V(hT)} \left( \vect{q}_V(hT)\!-\!\vect{q}_V^r(hT)\right)\\
		&+\vect{A}^{(i)}_{\vect{q}_i^r(hT+\tau)} \vect{\dot{q}}_i^r(hT+\tau) \\
		&+\vect{A}^{(i)}_{\vect{q}_V^r(hT+\tau)} \vect{\dot{q}}_V^r(hT+\tau).
	\end{split}
\end{align}
In the following,
we analyze system dynamics under controller~\eqref{eq:new_control_law}
to evaluate stability and performance.
In view of~\eqref{eq:decentr_disc_time},
all results straightly carry over to a distributed architecture.
Furthermore,
in simulation we will test
the centralized controller where the feedback gain $ k $ is adaptive:
this cannot be decoupled,
because computing $ k $ online requires the full system configuration.
}

\section{\revision{Stability Limitations of SIKM}}\label{sec:stability-limitations-SIKM}

\revision{In this section we prove a negative result:
perfect nominal tracking cannot be ensured for some values of $ (k,T) $.
In particular,
the gain $k$ needs to be sufficiently large \textit{and} the sampling time $ T $ sufficiently small to avoid instability.
To this aim,
we state some preliminary assumptions.}

\begin{ass}\label{ass:squareInvertibleSystem}
	The following relations hold.
	\begin{enumerate}[label=\roman*)]
		\item\label{ass:squareInvertibleSystem:1} The reference trajectory $\vect{q}^r(t)\in \mathcal{Q}$ is twice continuously differentiable and $\mathcal{Q}$ is a compact set.
		Moreover, velocities and accelerations are uniformly bounded, i.e.,  $\|\dot{\vect{q}}^r(t)\|\le v_{\textit{max}}$ and $\|\ddot{\vect{q}}^r(t)\|\le a_{\textit{max}}$.
		\item\label{ass:squareInvertibleSystem:3} \revision{$\vect{A}_{\vect{q}(0)}$ is twice continuously differentiable and invertible.}
		\item\label{ass:squareInvertibleSystem:2} There exists $d>0$ such that,
		for any \revision{configuration $\vect{q}$ at distance smaller than $d$ from the reference trajectory},
		i.e.,
		$\exists t : \|\vect{q}-\vect{q}^r(t)\| < d$,
		$\vect{A}_\vect{q}$ is twice continuously differentiable and invertible.
	\end{enumerate}
\end{ass}
Assumption~\ref{ass:squareInvertibleSystem}-\ref{ass:squareInvertibleSystem:1} are
smoothness properties of the reference trajectory needed to derive error bounds.
\revision{Assumption~\ref{ass:squareInvertibleSystem}-\ref{ass:squareInvertibleSystem:3} is required to avoid that the initial condition is a singular point.
Assumption~\ref{ass:squareInvertibleSystem}-\ref{ass:squareInvertibleSystem:2} further guarantees that all configurations $ \vect{q}^r(t) $ belonging to
the reference trajectory are distant enough from singular points,
ensuring robustness} to,
e.g.,
external disturbances,
and can be accommodated through an offline high-level planner.
We will later show that our proposed control strategies,
under such assumptions, 
also guarantees that the \revision{actual} trajectory is always bounded away from \revision{singular points}.

The error flow $\vect{f}(\cdot,\cdot,\cdot)$ in~\eqref{eq:decentr_disc_time} is discontinuous 
because the feedback term depends on $\vect{e}(hT)$ and resets at every sampling instant $t=hT$.
Thus, 
existence of a global solution based on standard Lipschitz continuity cannot be invoked, 
in general.
However, 
in view of Assumption \ref{ass:squareInvertibleSystem}, 
the flow is Lipschitz continuous for $\tau\in[0,T)$.
Hence, 
if we can show that a solution $\vect{e}(hT+\tau)$ exists for any $\tau\in[0,T)$,
and that the limit $\lim_{\tau\to T} \vect{e}(hT+\tau)$ exists finite starting from any $\vect{e}(hT)$ satisfying Assumption \ref{ass:squareInvertibleSystem},
then global existence is guaranteed by patching together those intervals.
As so, 
we study the flow $\vect{f}(\vect{e}(t), \vect{q}^r(t), \vect{\dot{q}}^r(t); \vect{e}(hT))$ for $t=hT+\tau, \, \tau\in[0,T)$, where
\revision{we make the dependence on the ``initial condition'' $\vect{e}(hT)$ explicit.}
To prove asymptotic stability, we will use the following notion of contraction.
\begin{defn}
	\label{def:rho_monot}
	Given fixed $T$, $k$, and under Assumption \ref{ass:squareInvertibleSystem}, the flow $ \vect{f}(\vect{e}(t), \vect{q}^r(t), \vect{\dot{q}}^r(t); \vect{e}(hT))$ is \emph{$\rho$-monotonically contractive} if, for any $\|\vect{e}(hT)\|<d$, it holds
	\begin{enumerate}
		\item $ \| e(hT+\tau)\|\leq  \|e(hT)\|, \ \ \tau \in[0,T) $,
		\item $ \lim_{\tau\to T}\| e(hT+\tau)\|\leq \rho \|e(hT)\|, \ \rho\in[0,1) $.
	\end{enumerate}
\end{defn}
Given Definition~\ref{def:rho_monot}, the following lemma easily follows.
\begin{lem}\label{lem:contractive_trajectory}
	If the error flow defined in~\eqref{eqn:error_flow} is $\rho$-monotonically contractive and under Assumption~\ref{ass:squareInvertibleSystem}, then 
	$$\|\vect{q}(t)-\vect{q}^r(t)\|\leq \rho^{\frac{t}{T}-1} \|\vect{q}(0)-\vect{q}^r(0)\|, \ \ t\geq 0.$$
\end{lem}
\revision{Lemma~\ref{lem:contractive_trajectory} ensures exponential convergence to the reference trajectory.
The property of $\rho$-monotonically contractiveness might appear rather strong since it must hold for any segment of the trajectory of length $T$, 
however,
it is necessary to guarantee that the system avoids singularity configurations at all times.
The next result encodes necessary conditions for stability.}
\begin{prop}\label{prop:negative_property}
	Under Assumption \ref{ass:squareInvertibleSystem}, there exists $k_{min}>0$ such that, if one of the following conditions is satisfied,
	\begin{enumerate}
		\item $k<k_{min}$, 
		\item $k>\frac{2}{T}$,
	\end{enumerate}
	the error flow~\eqref{eqn:error_flow} is {not} $\rho$-monotonically contractive.
\end{prop}
Intuitively, if the gain $ k $ is too small,
there exist reference trajectories such that the feedforward term ``pushes" the system too much without being suitably balanced by the feedback term.
{Conversely, if $ k $ is too large, the feedback causes the trajectory to overshoot, possibly amplifying the error of the initial condition.}
\begin{pf}
	\emph{1)} We will prove the first condition showing that the error norm initially increases for some initial conditions $\vect{e}(hT)$ and choice of reference trajectory $\vect{q}^r(t)$. 
	Consider the following Lyapunov function
	\begin{equation}\label{eqn:lyap}
		V(\tau)=\frac{1}{2}\|\vect{e}(hT+\tau)\|^2,
	\end{equation}
	and its time derivative at $\tau=0$ (cf.~\eqref{eqn:error_flow}):
	\begin{align*}
		\dot{V}(0)&= &&\vect{e}(hT)^\top  \dot{\vect{e}}(hT) \\
		&= &&-k\|\vect{e}(hT)\|^2 +\\
		&  &&\underbrace{\vect{e}^\top(hT)(\vect{A}^{-1}_{\vect{e}(hT)+\vect{q}^r(hT)}\vect{A}_{\vect{q}^r(hT)}-\vect{I})\dot{\vect{q}}^r(hT)}_{\overset{\text{def}}{=}\phi(\vect{e}(hT),\vect{q}^r(hT),\dot{\vect{q}}^r(hT))}. 
	\end{align*}
	\revision{Let us define
	\begin{align*}
	     c_{max}&\overset{\text{def}}{=}&& \max \; c \\
		 & \mbox{ s.t.}  && \|\vect{e}(hT)\|\leq d, \ \vect{q}^r(hT)\in\mathcal{Q}, \\
		 & &&\phi(\vect{e}(hT),\vect{q}^r(hT),\dot{\vect{q}}^r(hT))\geq c v_{\textit{max}} \|\vect{e}(hT)\|^2 .
  	\end{align*}
	The parameter $c_{max}$ is surely strictly positive because 
	(i) all arguments of $\phi(\cdot)$ are defined on a compact set
	(see also Assumption 1-i)) and 
	(ii) $\phi(\cdot)$ is continuously differentiable. 
	The only case for which $c_{max}=0$ is when $\vect{A}_{\vect{q}}$ is a constant matrix for all $\vect{q}$. 
	This scenario is not admissible since it would imply $\dot{\vect{p}}=\vect{A}\dot{\vect{q}}$,
	which is not the case for the problem at hand. 
    }
    This implies that there exist tuples $ (\vect{e}(hT),\vect{q}^r(hT),\dot{\vect{q}}^r(hT)) $ such that
	$$ \dot{V}(0) \ge -(k-c_{max} v_{\textit{max}})\|\vect{e}(hT)\|^2$$
	Let $k_{min}\overset{\text{def}}{=}c_{max} v_{\textit{max}} $, then $\dot{V}(0)>0$ for any $k<k_{min}$, hence there exists $0<\bar \tau<T$ s.t. $\|\vect{e}(hT+\tau)\|>\|\vect{e}(hT)\| \, \forall \tau \in (0,\bar\tau)$ and~\eqref{eqn:error_flow}  it is not $\rho$-monotonically contractive.
	
	\emph{2)} In order to prove the second condition we first choose $\vect{q}^r(t)\equiv\vect{q}^r, \, t \geq 0$, which satisfies 
	Assumption~\ref{ass:squareInvertibleSystem}.
	As so, the error dynamics reduce to  
	$$ \vect{\dot e}(hT+\tau)= -k \underbrace{\vect{A}^{-1}_{\vect{e}(hT+\tau)+\vect{q}^r} \vect{A}_{\vect{e}(hT)+\vect{q}^r}\vect{e}(hT)}_{\overset{\mathrm{def}}{=} \vect{g}(\vect{e}(hT+\tau))}.$$
	The error trajectory can be written in the form 
	\begin{equation*}
		\vect{e}(hT+\tau) = \vect{e}(hT)+k\int_{hT}^{hT+\tau}{\vect{g}(\vect{e}(hT+\tau')) d\tau'}.
	\end{equation*}
	By using Taylor's theorem for multivariate functions with integral form of the remainder, it becomes
	\begin{align}
		&\vect{e}(hT+\tau) = \vect{e}(hT) +k \tau\, \vect{g}(\vect{e}(hT)) + \nonumber \\
		& +k^2\tau^2 \int_{0}^{1}(1-\varepsilon) \frac{\partial\vect{g}(\vect{e}(hT+\epsilon \tau))}{\partial \vect{e}} \ \vect{g}(\vect{{e}}(hT+\varepsilon\tau))d \varepsilon \nonumber \\
		&= (1-k\tau)\vect{e}(hT)+ k^2\tau^2 \vect{r}(\vect{e}(hT+\tau);\vect{e}(hT)),
		\label{eq:sol_taylor1}
	\end{align}
	where $\vect{r}(\cdot)$ is the second-order reminder where we made explicit the dependence on $\vect{e}(hT)$. Under Assumption \ref{ass:squareInvertibleSystem}, the functions $\vect{g}$ and $\frac{\partial \vect{g}}{\partial \vect{e}}$ are continuously differentiable. Moreover their arguments are defined in a compact set and have the additional properties that  $\vect{g}(\vect{e}(hT+\tau))=0, \frac{\partial \vect{g}(hT+\tau)}{\partial \vect{e}}=\vect{0}, \   \forall \tau \in[0,T]$ if $\vect{e}(hT)=\vect{0}$. 
	Therefore, applying Lemma \ref{lem:up_bound_2_fcn}, there must exist $\delta>0$ such that
	$$ \|\vect{r}(\vect{e}(hT+\tau);\vect{e}(hT))\|\leq \delta \|\vect{e}(hT)\|^2, \ \ \forall \tau \in[0,T),\forall \vect{e}(hT).$$
	We now use the reverse triangle inequality and the previous inequality to get:
	\begin{align*}
		&\|\vect{e}(hT+\tau)\| \geq \\
		& \geq\left| \|(1-k\tau)\vect{e}(hT)\| - k^2\tau^2 \|\vect{r}(\vect{e}(hT+\tau);\vect{e}(hT))\| \right| \\
		&\geq \max\{ 0\, , \, |1-k\tau|-k^2\tau^2\delta\|\vect{e}(hT)\| \}\|\vect{e}(hT)\|.
	\end{align*}
	If $kT>2$, there exist $\epsilon>0$ and $\bar \tau\in(0,T)$ s.t. $k\bar\tau = 2+\epsilon$.
	Then, we can choose $\vect{e}(hT)$ s.t. $\|\vect{e}(hT)\| = \frac{\epsilon}{2k\bar{\tau}^2\delta}$, hence
	$$  \|\vect{e}(hT+\bar \tau)\| \geq \left(1+\epsilon -\frac{\epsilon}{2}\right)\|\vect{e}(hT)\| \geq \|\vect{e}(hT)\|.\qed$$
\end{pf}

\begin{rem}[Stability limitations on $T$ and $k$]
	Proposition~\ref{prop:negative_property} states that
	there are choices of $T$ and $k$ for which the error flow~\eqref{eqn:error_flow} is not $\rho$-monotonically contractive,
\revision{implying that the robots may not track the reference trajectory. 
	In particular,
	there exists a feedback gain $k_{min}$ below which
	the system cannot be $\rho$-monotonically contractive.
	Also, 
	there exists a sampling time $T_{max}=\frac{2}{k_{min}}$ above which the same instability consideration applies, 
	as graphically shown in Fig.~\ref{fig:tauS_idea}. 
	This is in stark contrast with the result for position stabilization in~\cite{8743454},
	where (doubly) exponential stability could be achieved without sharp limitation on sampling time $T$ or control gain $k$.
	This fundamental difference is mainly due to the novel feedforward term considered in this work.}
\end{rem}

The previous negative results entails necessary conditions on $ k $ and $ T $ for stability.
Conversely,
we next prove that there exist pairs of $T$ and $k$ for which $\rho$-monotonically contractiveness holds.

\begin{prop}
	\label{prop:rho_monotonically}
	Under Assumption \ref{ass:squareInvertibleSystem}, there exists $k_2>0$ such that for any $k>k_2$ there exists $T_c(k)$ such that for all $T< T_c(k)$ the error flow defined in~\eqref{eqn:error_flow} is  $\rho$-monotonically contractive.
\end{prop}
\begin{pf}
	Consider the Lyapunov function~\eqref{eqn:lyap} and its time derivative at $\tau=0$:
	\begin{multline*}
		\dot{V}(0)= -k\|\vect{e}(hT)\|^2+\\ 
		+\vect{e}^\top(hT)(\vect{A}^{-1}_{\vect{e}(hT)+\vect{q}^r(hT)}\vect{A}_{\vect{q}^r(hT)}-\vect{I})\dot{\vect{q}}^r(hT).
	\end{multline*}
	Note that $\vect{A}^{-1}_{\vect{e}(hT)\!+\!\vect{q}^r(hT)}\vect{A}_{\vect{q}^r(hT)}\!-\!\vect{I}=\vect{0} \ \forall \vect{q}^r(hT)$ if $\vect{e}(hT)=0$.
	Moreover, 
	being $\vect{e}(hT)$ and $\vect{q}^r(hT)$ in a compact set according to Assumption \ref{ass:squareInvertibleSystem},
	by virtue of Lemma~\ref{lem:up_bound_2_vars} it holds 
	$\| \vect{A}^{-1}_{\vect{e}(hT)\!+\!\vect{q}^r(hT)}\vect{A}_{\vect{q}^r(hT)}\!-\!\vect{I}\| \leq \delta \| \vect{e}(hT) \|$ for some $\delta>0$, 
	therefore
	\begin{align*}
		\dot{V}(0) &\leq -k\|\vect{e}(hT)\|^2 + \delta \|\vect{e}(hT)\|^2 v_{max}\\
		&=-(k-\delta v_{max})\|\vect{e}(hT)\|^2 = -2(k-\delta v_{max})V(0).
	\end{align*}
	Let $k_2\overset{\text{def}}{=} \delta v_{max}$, then for each $k>k_2$, there exist $\xi \in(0,k-\delta v_{max})$ and  $T_c(k)$ s.t.
	\begin{align*} 
		&V(\tau) \leq e^{-2\xi \tau}V(0), \tau\in[0,T_c(k) ) \\
		&\Longrightarrow \|\vect{e}(hT+\tau)\| \leq e^{-\xi\tau} \|\vect{e}(hT)\|, \ \tau\in[0,T_c(k)).
	\end{align*}
	As a consequence, by defining $\rho\overset{\text{def}}{=} e^{-T\xi}<1$ for any $T<T_c(k)$, it holds
	\begin{align*} 
		& \|\vect{e}(hT+\tau)\| \leq  \rho\|\vect{e}(hT)\|, \ \tau\in[0,T)\\
		&\lim_{\tau\to T}\|\vect{e}(hT+\tau)\| \leq \rho \|\vect{e}(hT)\|,
	\end{align*}
	i.e., the system is $\rho$-monotonically contractive.\qed
\end{pf}

By combining this result with the previous one,
it follows that $\lim_{k\to \infty} T_c(k) = 0 $ and that $T_c(k) < T_{max} \, \forall k>k_2 $.
From Proposition~\ref{prop:negative_property}, 
we know that~\eqref{eqn:error_flow} is not $\rho$-monotonically contractive for values of $(k,T)$ in the red area of Fig.~\ref{fig:tauS_idea}.
Conversely, Proposition~\ref{prop:rho_monotonically} 
asserts that there exist a non-empty area within the parameter space,
represented by the white area below the dashed red curve in Fig.~\ref{fig:tauS_idea},
such that~\eqref{eqn:error_flow} is $\rho$-monotonically contractive.

\revision{In the next section,
	we explicitly compute a bounded area in the $ (k,T) $-space that guarantees $\rho$-monotonic contractiveness of~\eqref{eqn:error_flow},
	and hence perfect asymptotic tracking under nominal conditions.}

\section{Stability and Convergence Rate for SIKM}
\label{sec:stab_and_conv_rate}

All results in the previous section are of existential type, i.e.,
they give no indication about computing the pairs $k$ and $T$ for which the system is $\rho$-monotonically contractive,
nor the convergence rate $\rho$. 
In this section, we find explicit bounds on the ``stability'' set (green area in Fig.~\ref{fig:tauS_idea}) where the error norm decreases.
Also, we characterize the convergence rate and find the pairs of gain and sampling time that induce the fastest convergence.
\begin{rem}[Stability]
	When mentioning stability, we imply that the flow \eqref{eqn:error_flow} is $\rho$-monotonically contractive.
\end{rem}


\begin{prop}\label{prop:z_intro}
	For system \eqref{eqn:error_flow},
	it is possible to upper bound the decrease rate of the error norm $ \|\vect{e}(\cdot)\| $,
	\begin{align}
	\|\vect{e}(hT+\tau)\| \leq z\big(k,\tau ; \mu, \alpha, \gamma_1, \gamma_2\big)\cdot \|\vect{e}(hT)\|, 
	\label{eq:z_def}
	\end{align}
	for $h\in\mathbb{N}$, $\tau \in (0, \tau_s(k))$, and
	\begin{align*}
	z\big(k,\tau ; \mu, \alpha, \gamma_1, \gamma_2\big)\!=\!|1-k\tau|\!+\!\tau \alpha\!+\!\tau^2 (k^2 \mu + k \gamma_1+\gamma_2),
	\end{align*}
	where $\alpha, \mu, \gamma_1,\gamma_2 $ are positive constants.
\end{prop}
\begin{pf}[Sketch]
	In virtue of the Lipschitz properties of the error flow in the interval $ (hT,hT+T) $,
	Taylor's theorem with integral form of the remainder for multivariate functions is used to decompose the error $ \vect{e}(hT+\tau) $
	as a sum of terms which can be upper bounded individually.
	Notice that $ \alpha $, $ \mu $, $ \gamma_1 $ and $ \gamma_2 $ are still unknown at this stage.
\end{pf}
\begin{pf}
	The flow defined by~\eqref{eqn:error_flow} can be written as 
	\begin{equation*}
	\vect{e}(hT+\tau) = \vect{e}(hT)+\int_{hT}^{hT+\tau}{\vect{f}(\vect{e}(\cdot), \vect{q}^r(\cdot), \vect{\dot{q}}^r(\cdot))\bigg |_{hT+\tau'}  d\tau'}.
	\end{equation*}
	By using Taylor's theorem for multivariate functions with integral form of the remainder, we get
	\begin{align}
	&\vect{e}(hT+\tau) = \vect{e}(hT) +\tau \,\vect{f}(\vect{e}(hT), \vect{q}^r(hT), \vect{\dot{q}}^r(hT))+ \nonumber \\
	& +\tau^2 \int_{0}^{1}(1-\varepsilon)\cdot\bigg[\frac{\partial\vect{f}(\vect{e}(\cdot), \vect{q}^r(\cdot), \vect{\dot{q}}^r(\cdot))}{\partial\vect{e}(\cdot)} \ \vect{f}(\vect{e}(\cdot), \vect{q}^r(\cdot), \vect{\dot{q}}^r(\cdot)) +\nonumber \\
	&+ \frac{\partial\vect{f}(\vect{e}(\cdot), \vect{q}^r(\cdot), \vect{\dot{q}}^r(\cdot))}{\partial \vect{q}^r(\cdot)} \ \vect{\dot{q}}^r(\cdot) + \nonumber\\
	&+ \frac{\partial\vect{f}(\vect{e}(\cdot), \vect{q}^r(\cdot), \vect{\dot{q}}^r(\cdot))}{\partial \vect{\dot{q}}^r(\cdot)} \ \vect{\ddot{q}}^r(\cdot)\bigg]_{hT+\varepsilon\tau}d \varepsilon.
	\label{eq:sol_taylor}
	\end{align}
	The above expression allows to find an upper bound of $\|\vect{e}(hT+\tau)\|, \ \tau \in (0, \tau_s(k))$, by acting on the single terms of \eqref{eq:sol_taylor}.
	Hence, we can compute a more precise estimate of the convergence rate, i.e., how quickly $\|\vect{e}(hT+\tau)\|$, with $ \tau \in (0, \tau_s(k))$, decreases w.r.t. $\|\vect{e}(hT)\|$.
	Moreover, the following computations will be useful to find values $(k,T)$ s.t. the system is stable and the tracking error converges to zero.
	First, we observe that
	\begin{multline}\label{eq:f_hT_property}
	\vect{f}(\vect{e}(hT), \vect{q}^r(hT), \vect{\dot{q}}^r(hT))= -k\vect{e}(hT)+\\
	+\big(\vect{A}^{-1}_{\vect{e}(hT)+\vect{q}^r(hT)} \vect{A}_{\vect{q}^r(hT)}-\vect{I}\big) \ \vect{\dot{q}}^r(hT).
	\end{multline}
	By using Lemma~\ref{lem:up_bound_2_vars} on the second addend, it holds
	$$\|\big(\vect{A}^{-1}_{\vect{e}(hT)+\vect{q}^r(hT)} \vect{A}_{\vect{q}^r(hT)}-\vect{I}\big) \ \vect{\dot{q}}^r(hT)\|\leq \alpha\|\vect{e}(hT)\|,$$
	where $\alpha\overset{\text{def}}{=} a   v_{max}$ for some $ a>0 $ that depends on the system dynamics. Moreover, from \eqref{eqn:error_flow}, 
	\begin{align}
	&\|\vect{f}(\vect{e}(hT+\tau), \vect{q}^r(hT+\tau), \vect{\dot{q}}^r(hT+\tau))\|  \leq \nonumber\\
	&\leq \|-k \vect{A}^{-1}_{\vect{e}(hT+\tau)+\vect{q}^r(hT+\tau)} \vect{A}_{\vect{e}(hT)+\vect{q}^r(hT)}\vect{e}(hT)\| + \nonumber\\
	& + \| \big(\vect{A}^{-1}_{\vect{e}(hT+\tau)+\vect{q}^r(hT+\tau)}  \vect{A}_{\vect{q}^r(hT+\tau)}-\vect{I}\big)\vect{\dot{q}}^r(hT+\tau)\|. \label{eq:f_hT_plus_tau_leq}
	\end{align}
	Both addends in \eqref{eq:f_hT_plus_tau_leq} are continuously differentiable functions on a compact set.
	In addition, they are equal to zero if $\vect{e}(hT)=\vect{0}$ and $\vect{e}(hT+\tau)=\vect{0}$, respectively.
	Then, we can apply Lemmas~\ref{lem:up_bound}-\ref{lem:up_bound_2_vars} to get
	$
	\|k \vect{A}^{-1}_{\vect{e}(hT+\tau)+\vect{q}^r(hT+\tau)} \vect{A}_{\vect{e}(hT)+\vect{q}^r(hT)}\vect{e}(hT)\|\leq kb\|\vect{e}(hT)\|, \ b>0,
	$ and $
	\| \big(\vect{A}^{-1}_{\vect{e}(hT+\tau)+\vect{q}^r(hT+\tau)}  \vect{A}_{\vect{q}^r(hT+\tau)}-\vect{I}\big)\vect{\dot{q}}^r(hT+\tau)\| \leq c\|\vect{e}(hT)\| v_{max}, \ c>0
	$,
	where we used Proposition~\ref{prop:rho_monotonically} and Definition~\ref{def:rho_monot}
	and constants $ b,c $ depend on the system dynamics.
	Thus, we rewrite \eqref{eq:f_hT_plus_tau_leq} as
	\begin{align}\label{eq:f_leq_bc}
	\hspace{-3.4mm}\|\vect{f}(\vect{e}(\cdot), \vect{q}^r(\cdot), \vect{\dot{q}}^r(\cdot)\!)\big|_{hT+\tau}\|
	\!\leq\! (k  b\! +\! c  v_{max})\|\vect{e}(hT)\|.
	\end{align}
	Also, being $\vect{f}(\vect{e}(\cdot), \vect{q}^r(\cdot), \vect{\dot{q}}^r(\cdot))$ differentiable w.r.t. all its variables on the compact set $\mathcal{B}_{r}(\vect{e}(\cdot))$, 
	 we can derive the following bounds based on \eqref{eq:f_hT_property} :
	\begin{align}
	&\bigg \| \frac{\partial\vect{f}(\vect{e}(\cdot), \vect{q}^r(\cdot), \vect{\dot{q}}^r(\cdot))}{\partial \vect{e}}\bigg|_{hT+\tau} \bigg\|\leq (k+gv_{max}) \|\vect{e}(hT)\| \nonumber\\
	&\bigg \| \frac{\partial\vect{f}(\vect{e}(\cdot), \vect{q}^r(\cdot), \vect{\dot{q}}^r(\cdot))}{ \partial \vect{q}^r}\bigg|_{hT+\tau}\bigg\|\leq m v_{max} \|\vect{e}(hT)\| \nonumber\\
	&\bigg \| \frac{\partial\vect{f}(\vect{e}(\cdot), \vect{q}^r(\cdot), \vect{\dot{q}}^r(\cdot))}{ \partial \vect{\dot{q}}^r}\bigg|_{hT+\tau}\bigg\|\leq kn \|\vect{e}(hT)\| ,
	\label{eq:df_less_than_dghmn}
	\end{align}
	where $g,m,n>0$ also depend on the system dynamics.
	In virtue of \eqref{eq:f_hT_property}, \eqref{eq:f_leq_bc} and \eqref{eq:df_less_than_dghmn},
	we can derive the overall bound defined in \eqref{eq:z_def}:
	\begin{align*}
	\|\vect{e}(hT\!+\!\tau)\|&\!\leq\!\big(|1\!-\!k\tau|\!+\!\alpha\tau\! +\! \tau^2(k^2\mu\!+\!k\gamma_1\!+\!\gamma_2)\big)\|\vect{e}(hT)\|
	\end{align*}
	where, given $a>0$ defined in the proof of Proposition~\ref{prop:negative_property},
	\begin{align*}
	\alpha&\overset{\text{def}}{=} a \,  v_{max} \\
	\mu&\overset{\text{def}}{=}b\|\vect{e}(hT)\|\\ 
	\gamma_1&\overset{\text{def}}{=}(c+gb)v_{max} \|\vect{e}(hT)\|+n \, a_{max} \\
	\gamma_2&\overset{\text{def}}{=} g\, c\, v_{max}\|\vect{e}(hT)\|+m \, v_{max}.\qed
	\end{align*}
	
\end{pf}

Proposition~\ref{prop:z_intro} 
provides an expression for the convergence rate as a function
of the parameters $\mu,\alpha,\gamma_1,\gamma_2$. 
In the next section, we provide a procedure to estimate them.

\begin{rem}[Constant reference]
	Consider \eqref{eq:z_def} and suppose a constant reference is assigned, i.e., $\frac{d^{(n)} \, \vect{q}^r(t)}{d \, t^{(n)}}\equiv \vect{0}, n\geq1$. Then $\alpha \!= \!\gamma_1\! =\! \gamma_2\! = \!0$ and $z(k,\tau; \mu)$ coincides with the function $g(\tau;\mu)\big|_{\tau=k\tau}$ found in \cite{8743454}.
\end{rem}

\subsection{Estimating Parameters $\mu,\alpha,\gamma_1,\gamma_2$}
\label{sec:param_estim}

We would like to find an estimate of the function $z(k,\tau;\mu,\alpha,\gamma_1,\gamma_2)$ that bounds the convergence rate of the tracking error,
and choose the values $(k,T)$ which yield the fastest convergence.
To this aim, we provide a numerical procedure to estimate $\theta\overset{\text{def}}{=}(\mu,\alpha,\gamma_1,\gamma_2)$. 
Proposition~\ref{prop:z_intro} implies that the following set is nonempty,
\begin{multline*}
	\Theta \overset{\text{def}}{=}\{ \theta >0  \, | \, \|\vect{e}(hT+\tau)\| \leq z\big(k,\tau ; \theta \big)\|\vect{e}(hT)\| 
	, \\ \forall \vect{q}^r(\cdot), \vect{q}(0) \},
\end{multline*}
where the inequality is componentwise.
Ideally, we would like to pick the smallest possible values for the parameters in $\theta$
in order to get the largest set of pairs $(k,T)$ that induce stability.
One possibility is choosing
$$ \vartheta \overset{\text{def}}{=} \argmin_{\theta\in \Theta} \|\theta\|.$$
Such $\vartheta$ surely exists because $\vect{q}^r(\cdot)$ and $\vect{q}(0)$ belong to a compact set.
However, it cannot be computed numerically because one would need to check all pairs $ (k,T) $ and points $\vect{q}^r(t)$ and  $\vect{q}(0)$.
We propose a strategy to estimate $\vartheta$ by sampling $\vect{q}^r(t)$ and $\vect{q}(0)$ from their domains for different values of $ (k,T) $
and run simulations to get a set of samples $\{( k_i,T_i,\vect{e}_{h+1}^i,\vect{e}_h^i)\}_{i=1}^S$, $ S $ being the number of samples.
Let us define the following quantities,
\begin{align*}
y_i&\overset{\text{def}}{=}\|\vect{e}^i_{h+1}\|, \\
s_i^\top&\overset{\text{def}}{=}[T_i^2\  k_i^2 \ T_i \  T_i^2 k_i \ T_i^2] \|\vect{e}^i_h\|\\
b_i &\overset{\text{def}}{=} |1-k_iT_i| \, \|\vect{e}^i_h\|,
\end{align*}	 
then the 
inequality $
\|\vect{e}^i_{h+1}\| \leq z\big(k_i,T_i ; \theta \big)\|\vect{e}^i_h\|
$
can be written as $ y_i \leq s_i^\top \theta + b_i$.
Based on such sampled trajectories,
we solve the following quadratic programming:
\begin{equation}\label{NLP}
\begin{aligned}
\widehat \theta_S\overset{\text{def}}{=} &\,\argmin_{\theta}  && \|\theta\|^2 \\
&\ \, \text{s. t.}    && \theta\geq 0,\\
& 					&& y_i \leq s_i^\top \theta + b_i,\ \ i=1,\ldots,S.
\end{aligned}
\end{equation}
Unfortunately, it is possible that $\widehat \theta_S \notin \Theta$ since we are checking the inequality $\|\vect{e}(hT+\tau)\| \leq z\big(k,\tau ; \theta \big)\|\vect{e}(hT)\|$ over a finite number of points.
We expect that $\lim_{S\to \infty} \widehat \theta_S = \vartheta$ if
\revision{the sampling procedure covers domains of $\vect{q}^r(\cdot)$ and  $\vect{q}(0)$ widely enough}.
However, 
\revision{formally} proving this claim is nontrivial.
There might be alternative numerical strategies to compute better estimates of $\vartheta$,
or other parameter choices in the set $\Theta$.
Such comparison goes beyond the scope of this work, 
however, 
we will show effectiveness of our proposed strategy through simulations.

\subsection{Analysis of the Function $z\big(k,\tau ; \mu, \alpha, \gamma_1, \gamma_2\big)$}\label{sec:optimal-param}
\revision{In this section, 
	we analyze the function $z\left(k,\tau\right)$ in order to compute
	the bound $ \tau_s(k) $ of the stability region (green area in Fig.~\ref{fig:tauS_idea}),
	defined as\footnote{
		In the following, 
		we use the shorthand notation $z\big(k,\tau\big)\overset{\text{def}}{=}z\big(k,\tau ; \mu, \alpha, \gamma_1, \gamma_2\big)$ 
		for the sake of readability.}
	\begin{equation}\label{eq:stability-region}
		\mathcal{U}\overset{\text{def}}{=}\{(k,\tau) : z(k,\tau)<1\},
	\end{equation}
	and the convergence rate for each point in such region.}

We also compute the optimal controller gain $k_o(T)$ for a fixed sampling time $T$
and the optimal sampling time $T_s(k)$ for a fixed gain $ k $ to achieve the fastest convergence rate.
Notice that $z\big(k,\tau; \mu, \alpha, \gamma_1, \gamma_2\big)$ can be written as\footnote{In the following, we use the shorthand notation $z\big(k,\tau\big)\overset{\text{def}}{=}z\big(k,\tau ; \mu, \alpha, \gamma_1, \gamma_2\big)$ for the sake of readability.}:
\[z\big(k,\tau \big)=\left\{\begin{array}{lr}
	z^-(k,\tau) & \text{ if } k\tau<1 \\
	z^+(k,\tau) & \text{ if } k\tau>1 
\end{array}\right.,\]
where
\begin{align*}
	z^-(k,\tau) &= 1+\tau(\alpha-k)+\tau^2(k^2\mu+k\gamma_1+\gamma_2) \\
	z^+(k,\tau) & =-1+\tau(\alpha+k)+\tau^2(k^2\mu+k\gamma_1+\gamma_2). 
\end{align*}
\begin{prop}\label{prop:tau_s}
	The guaranteed stability set of $z(k,\tau)$ is 
	$$\mathcal{U}=\{\alpha<k<+\infty,0<\tau<\tau_s(k)\},$$
	where $\alpha=k_{min}$ was defined in Proposition~\ref{prop:z_intro} and
	\[{
			\tau_s(k)=\left\{\begin{array}{lll}
				\tau_{s_1}(k) & &\text{ if } \mu>1 \\
				\tau_{s_1}(k) & \text{ for }\alpha < k < \bar{k} &\text{ if } \mu<1 \\
				\tau_{s_2}(k) & \text{ for } k > \bar{k} & \text{ if } \mu<1\\
			\end{array}\right.
	},
	\]
	where $\tau_{s_1}\overset{\text{def}}{=}\frac{k-\alpha}{k^2\mu+k\gamma_1+\gamma_2} $,
	$\bar{k}\overset{\text{def}}{=}\frac{\alpha+\gamma_1+ \sqrt{(\alpha+\gamma_1)^2+4\gamma_2(1-\mu)}}{2(1-\mu)} $, $\tau_{s_2}\overset{\text{def}}{=}\frac{-(\alpha+k)+\sqrt{(\alpha+k)^2+8(k^2\mu+k\gamma_1+\gamma_2)}}{2(k^2\mu+k\gamma_1+\gamma_2)}$.
\end{prop}

With fixed $k$, we can define the sampling time and convergence rate as functions of $k$: 
\begin{gather}
	\!\!\!\tau_o(k)\overset{\text{def}}{=}\argmin_{\tau} z(k,\tau)\label{eq:optimal_time}\\
	\rho_o(k)\overset{\text{def}}{=}z(k,\tau_o(k)), \label{def:rho}
\end{gather}
where $\tau_o$ is the time when $\|\vect{e}(hT+\tau)\|, \ 0\leq \tau \leq T,$ is closest to the origin, hence it corresponds to the fastest convergence rate of the error $\rho_o$.
\begin{prop}\label{prop:tau_o_k_fixed}
	With fixed $k$, the optimal sampling time $\tau_o(k)$ is given by
	\[{
			\tau_o(k)=\left\{\begin{array}{lll}
				\tau_{o_1}(k) & &\text{if } \mu>\frac{1}{2}\\
				\tau_{o_1}(k) &\text{ for }\alpha < k < \bar{\bar{k}} &\text{if } \mu<\frac{1}{2}\\
				\frac{1}{k} &\text{for }k > \bar{\bar{k}} &\text{if } \mu<\frac{1}{2}\\
			\end{array}\right.
	},
	\]
	$\tau_{o_1}(k)\overset{\text{def}}{=}\frac{k-\alpha}{2(k^2\mu+k\gamma_1+\gamma_2)}$,
	$\bar{\bar{k}}\overset{\text{def}}{=}\frac{\alpha+2\gamma_1+ \sqrt{(\alpha+2\gamma_1)^2+8\gamma_2(1-2\mu)}}{2(1-2\mu)}$.
	The corresponding convergence rate $\rho_o(k)$ is
	\[{
			\rho_o(k)=\left\{\begin{array}{llr}
				\rho_{k_1}(k) & & \text{ if } \mu>\frac{1}{2} \\
				\rho_{k_1}(k) &\text{for } \alpha<k < \bar{\bar{k}} &\text{if } \mu<\frac{1}{2}\\
				\rho_{k_2}(k) &\text{for }k > \bar{\bar{k}} &\text{if } \mu<\frac{1}{2}\\
			\end{array}\right.
	},
	\]
	where $\rho_{k_1}(k)\overset{\text{def}}{=}1-\frac{(\alpha-k)^2}{4(k^2\mu+k\gamma_1+\gamma_2)}$ and $\rho_{k_2}(k)\overset{\text{def}}{=}\mu + \frac{1}{k}(\alpha+\gamma_1) + \frac{\gamma_2}{k^2}$.
\end{prop}

The same quantities can be found as functions of $\tau$:
\begin{gather}
	k_o(\tau)\overset{\text{def}}{=}\argmin_{k} z(k,\tau)\label{eq:optimal_k}\\
	\rho_o(\tau)\overset{\text{def}}{=}z(k_o(\tau),\tau). \label{def:rho_tau_fixed}
\end{gather}
\begin{prop}\label{prop:k_o_tau_fixed}
	With fixed $\tau$, the optimal gain $k_o(\tau)$ is given by
	\[{
			k_o(\tau)=\left\{\begin{array}{llr}
				\frac{1-\tau \gamma_1}{2\tau \mu} & \text{for } 0 < \tau < \tau_{M_\tau} \ \ &\text{if } \mu>\frac{1}{2} \\
				\frac{1-\tau \gamma_1}{2\tau \mu} &\text{for } \tau_{m_k} < \tau < \tau_{M_\tau} &\text{if }  \mu<\frac{1}{2}\\
				\frac{1}{\tau} &\text{for }0<\tau<\tau_{m_k} &\text{if }  \mu<\frac{1}{2}\\
			\end{array}\right. 
	},
	\]
	where $\tau_{m_k}\overset{\text{def}}{=}\frac{1-2\mu}{\gamma_1}$,
	$\tau_{M_\tau}\overset{\text{def}}{=}\min\{\frac{1}{\gamma_1},\tau^-_{v_2}\}$, and
	$\tau_{v_{2}}^-=\frac{-(\gamma_1+\alpha \mu)+ \sqrt{(\gamma_1+\alpha \mu)^2+(-\gamma_1^2+4\gamma_2\mu)}}{-\gamma_1^2+4\gamma_2\mu}$.
	The corresponding convergence rate $\rho_o(\tau)$ is:
	\[{
			\rho_o(\tau)=\left\{\begin{array}{lll}
				\rho_{\tau_1}(\tau) &\text{for } 0 < \tau <\tau_{M_\tau} & \text{if } \mu>\frac{1}{2} \\
				\rho_{\tau_1}(\tau) &\text{for } \tau_{m_k} < \tau < \tau_{M_\tau} & \text{if } \mu<\frac{1}{2}\\
				\rho_{\tau_2}(\tau) &\text{for }0<\tau<\tau_{m_k} & \text{if } \mu<\frac{1}{2} \\
			\end{array}\right.
	},
	\]
	where $\rho_{\tau_1}(\tau)\overset{\text{def}}{=}\frac{(-\gamma_1^2+4\gamma_2 \mu)\tau^2+2(\gamma_1+2\alpha \mu)\tau+4\mu-1}{4 \mu}$ and  $\rho_{\tau_2}(\tau)\overset{\text{def}}{=}\gamma_2\tau^2+(\alpha+\gamma_1)\tau+\mu$.
\end{prop}

The proofs of Propositions~\ref{prop:tau_s}--\ref{prop:k_o_tau_fixed} are provided in Appendix~\ref{app:proof_props}.

\begin{figure*}
	\centering
		\subfloat[][\label{fig:plot_complete_mu04}]
		{\includegraphics[width=0.33\textwidth]{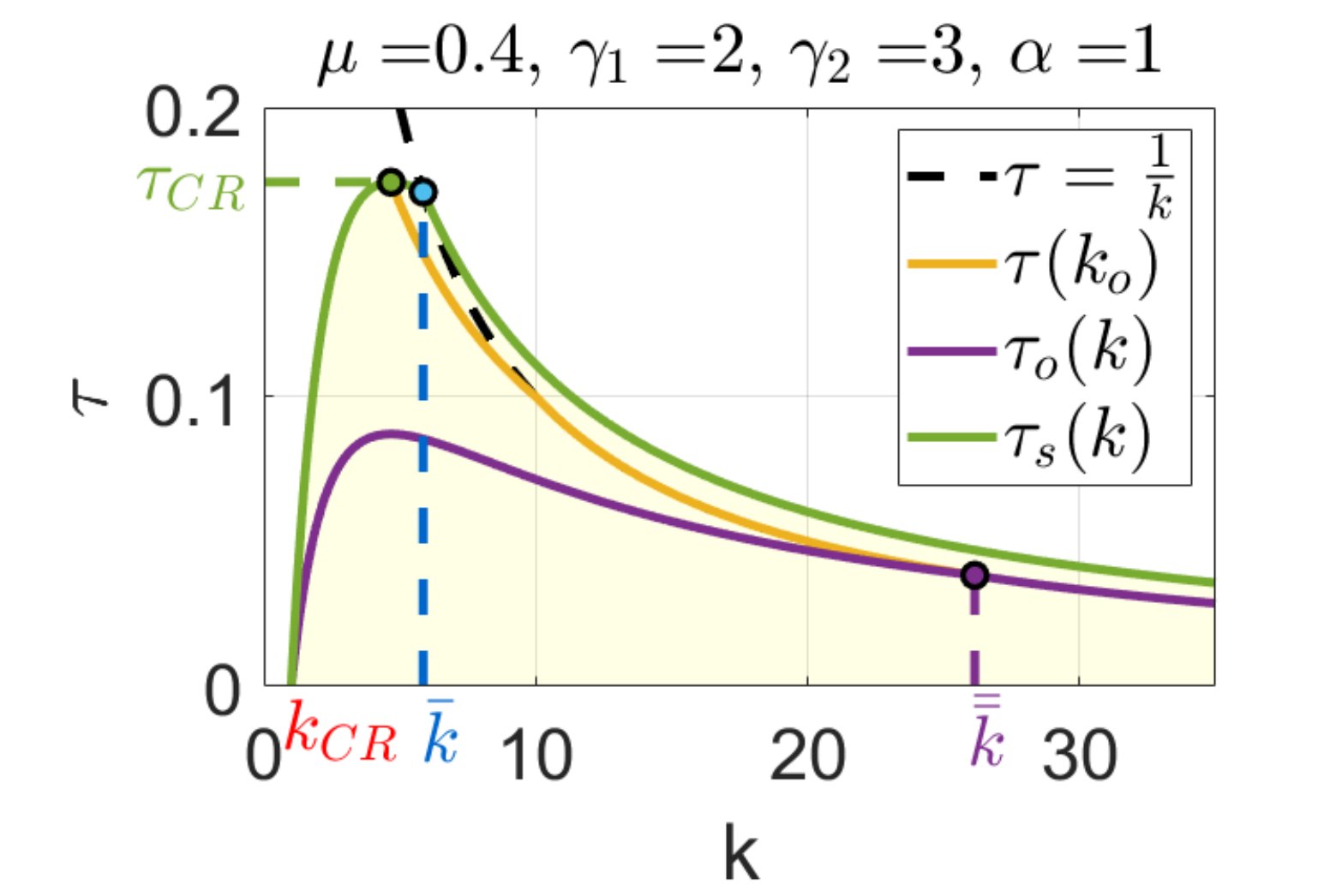}} \hspace{-0.5cm}
		\subfloat[][\label{fig:plot_complete_mu07}]
		{\includegraphics[width=0.33\textwidth]{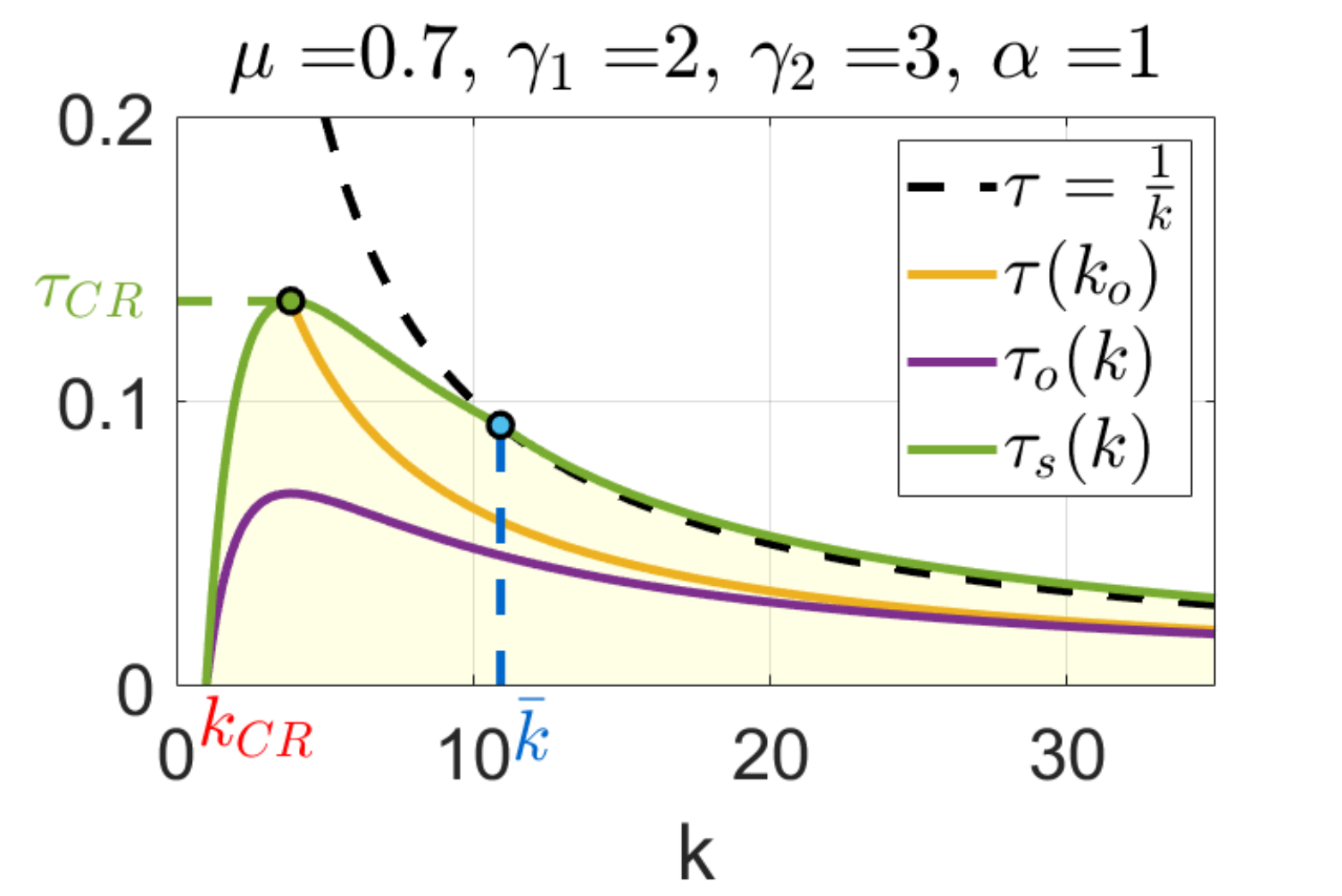}} \hspace{-0.5cm}
		\subfloat[][\label{fig:plot_complete_mu5}]
		{\includegraphics[width=0.33\textwidth]{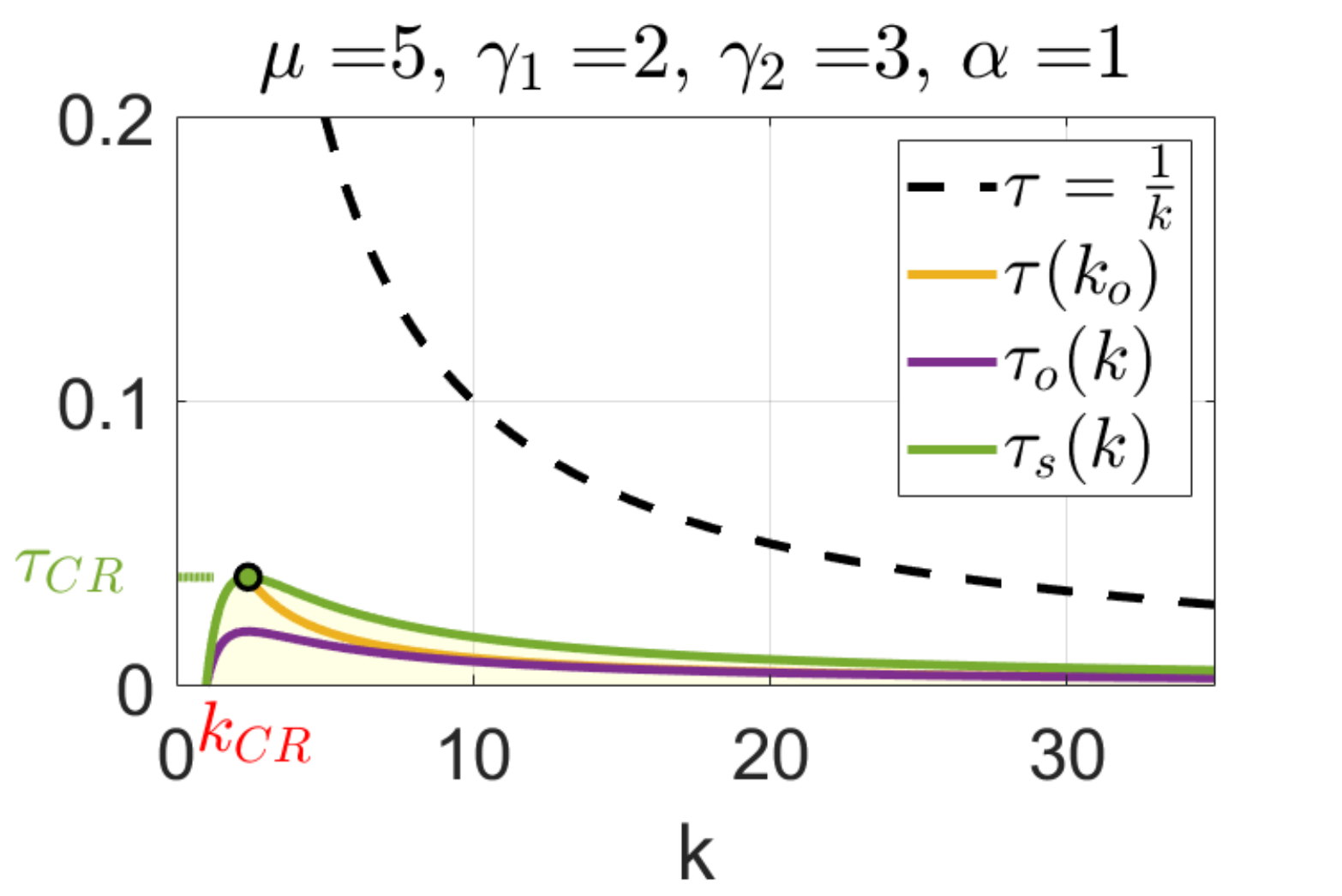}}
	\caption{Representation of the quantities $\tau_s(k)$, $\tau_o(k)$, $\tau(k_o)$ defined in Propositions~\ref{prop:tau_s}--\ref{prop:k_o_tau_fixed}.}
	\label{fig:plot_complete}
\end{figure*}

\begin{figure}[t]
	\centering
	\includegraphics[width=0.8\columnwidth]{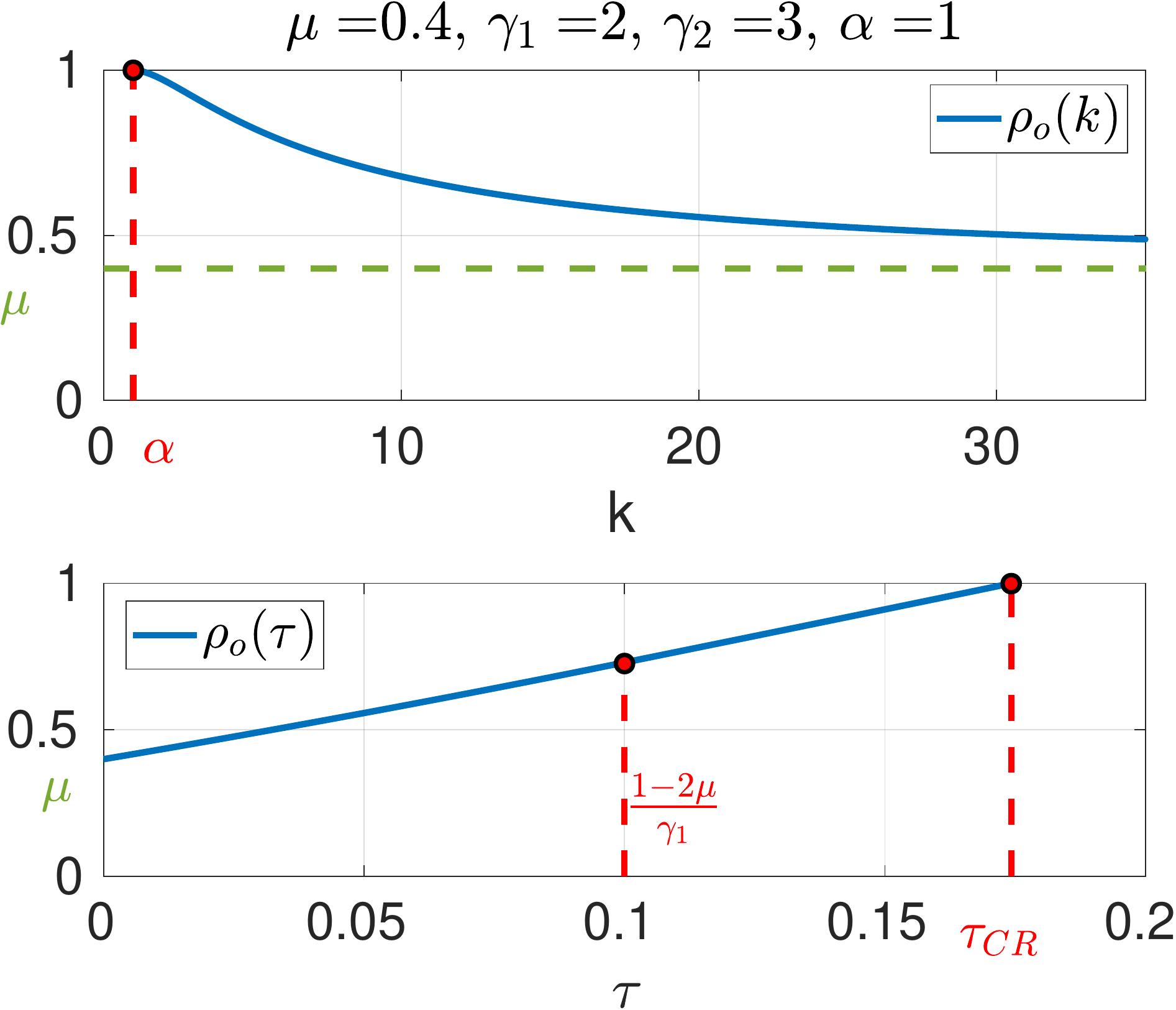}
	\caption{Estimated convergence rate $ \rho_0 $ as a function of the gain $k$ and of the sampling time $\tau$.}
	\label{fig:rho_mu04}
\end{figure}

Figure~\ref{fig:plot_complete} depicts the quantities defined above:
{in particular, $\tau(k_o)$ is obtained by 
inverting $k_o(\tau)$, being this monotonic decreasing,
and represents the sampling time for which $ k_o $ is the optimal gain.}
We chose three cases corresponding to $\mu<\nicefrac{1}{2}$ (Fig.~\ref{fig:plot_complete_mu04}),
$\nicefrac{1}{2}<\mu<1$ (Fig.~\ref{fig:plot_complete_mu07}),
and $\mu>1$ (Fig.~\ref{fig:plot_complete_mu5}), in order to span the cases given in the previous propositions.
Note that $\tau=\nicefrac{1}{k}$ represents the value s.t. $z^-(k,\tau)=z^+(k,\tau)$. 
Figure~\ref{fig:rho_mu04} represents the convergence rate $\rho_o$ as a function of $k$ (top plot) and $\tau$ (bottom plot).
The convergence rate is always smaller than one, hence
$\|\vect{e}(hT+\tau)\|\leq \|\vect{e}(hT)\|, \ \tau \in [0,T]$.

\begin{rem}[Sampling time and stabilizability]
	\revision{Proposition~\ref{prop:negative_property} shows that
		there exists a maximum sampling time $ T_\text{max} $ s.t. no feedback gain $ k $ can ensure stability if $ T>T_\text{max} $.
		Conversely,
		Proposition~\ref{prop:tau_s} implies that there exists a threshold $ \tau_{CR} $,
		which can be analytically found by setting $ \frac{d\tau_{s_1}(k)}{dk} = 0 $,
		s.t. a stabilizing gain always exists if $ T < \tau_{CR} $
		(see Fig.~\ref{fig:tauS_idea}).
		Such values may help to evaluate the communication hardware to be used.}
\end{rem}

\section{Simulation Results}\label{sec:simulations}

\begin{table}[b]
	\centering
	\setlength\belowcaptionskip{-0.4cm}
	\begin{tabular}{ l | c | c}
		Control Strategy & $\vect{u}_k(hT)$& $\vect{u}_{\mathrm{ff}}(hT+\tau)$\\
		\hline	
		SIKM-D (Eq.~\eqref{eq:decentr_disc_time}) & $-k_{\mathrm{off}}\vect{A}_{\vect{q}_h}(\vect{q}_h-\vect{q}^r_h)$ & $\vect{A}_{\vect{q}^r_{hT+\tau}}\vect{\dot{q}}^r_{hT+\tau}$ \\ PS (Eq.~\eqref{eq:u_k_hT}) & $-k_{\mathrm{on}}\vect{A}_{\vect{q}_h}(\vect{q}_h-\vect{q}^r_h)$ & $\vect{0}$  \\
		FF (Eq.~\eqref{eq:u_sampling_contr_law}) & $-k_{\mathrm{on}}\vect{A}_{\vect{q}_h}(\vect{q}_h-\vect{q}^r_h)$ & $\vect{A}_{\vect{q}_{hT}}\vect{\dot{q}}^r_{hT+\tau}$ \\
		SIKM-C (Eq.~\eqref{eq:new_control_law}) & $-k_{\mathrm{on}}\vect{A}_{\vect{q}_h}(\vect{q}_h-\vect{q}^r_h)$ & $\vect{A}_{\vect{q}^r_{hT+\tau}}\vect{\dot{q}}^r_{hT+\tau}$
	\end{tabular}
	\caption{Description of the four strategies used in simulation.}
	\label{tab:sim_cases}
\end{table}

In this section, we implement and compare four techniques for trajectory tracking, which are summarized in Table~\ref{tab:sim_cases}.
\revision{The first one, 
	named SIKM-D,
	is the distributed controller~\eqref{eq:decentr_disc_time} proposed in this work,
	where the gain $ k = k_\text{off} $ is computed offline following the procedure in Section~\ref{sec:stab_and_conv_rate}. 
	The last three, 
	referred to as PS, 
	FF, 
	and SIKM-C, 
	respectively, 
	are inspired by the online control strategy proposed in~\cite{8743454}
	and reported in Appendix~\ref{app:auxiliary_system}.
	In particular, 
	they all adopt a centralized communication architecture and differ only in the feedforward term design.
	Specifically, 
	at the $ h $-th step, 
	$ k_\text{on} $ is the solution to the following optimization problem with initial condition $ \vect{e}_h=\vect{e}(hT) $,
\begin{equation}\label{eq:k_online_design}
	\begin{aligned}
		\begin{split}
			k_\text{on}(hT) = &\;\argmin_k 	\!\!\!\!		&&\|\vect{e}(hT+T)\|\\
						  			  &\ \,\text{s. t.} && \vect{u}(\tau)= \vect{u}_k(hT)+\vect{u}_{\mathrm{ff}}(hT+\tau),
		\end{split}
	\end{aligned}
\end{equation}
where $\tau\in[hT,hT+T)$.
The PS technique corresponds to a simple point-stabilization with no feedforward correction, 
therefore it is expected to always lag behind the desired trajectory. 
The FF technique implements the naive controller~\eqref{eq:u_sampling_contr_law}
obtained by discretizing the standard continuous-time feedforward term which evaluates the Jacobian at $\vect{q}_{hT} $, 
that was shown insufficient for asymptotic tracking.
Even this strategy is expected to perform worse than SIKM-D despite the potential benefit coming from the online design of the feedback gain. 
Finally, 
SIKM-C has the same control structure as SIKM-D, 
but it re-computes the optimal feedback gain $k_{on}$ at each sampling time. 
This strategy should provide the best possible performance as opposed to SIKM-D,
which computes its gain offline solving a worst-case minmax problem as shown in Section~\ref{sec:stab_and_conv_rate}.
However,
SIKM-C requires solving an optimization problem and receiving the state of all robots at each sampling time.
}

\begin{rem}[\revision{Distributed vs. centralized control}]
	\revision{In contrast with the centralized scheme,
	a distributed communication control architecture avoids limitations due to all-to-all wireless communication,
	such as larger latency or packet loss, and 
	is more robust, cheaper and easier to maintain since it does not depend on the number of robots.  Specifically,
	the distributed communication controller requires \linebreak only one broadcast communication packet from the load/pivot to the robots, 
	while the centralized communication requires in addition to the broadcast packet also the
	(possibly synchronized) transmission of $N$ packets from robots to load.}
\end{rem}


\begin{figure}
	\centering
	\includegraphics[width=\columnwidth]{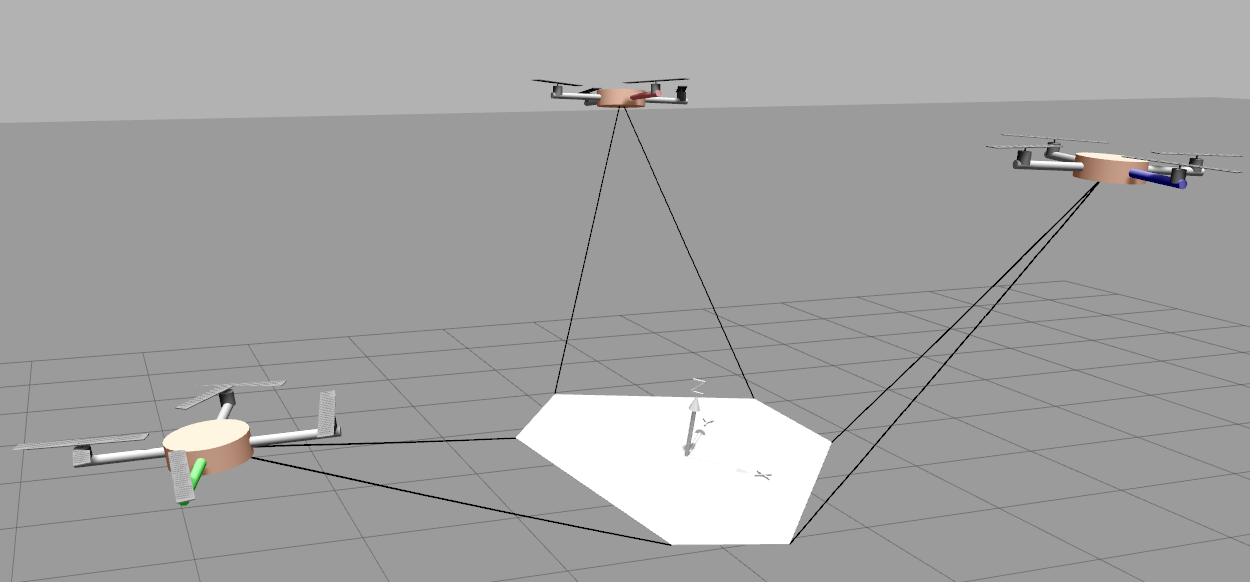}
	\caption{The Fly-Crane system in the Gazebo simulator.}
	\label{fig:sim}
\end{figure}

We tested the four techniques on the \textit{Fly-Crane} system \cite{75},
whose simulated environment is depicted in Fig.~\ref{fig:sim}.
The dynamical model of the system has been developed in a physics-based simulator (\textit{Gazebo}), 
simulating cables dynamics as well.
\revision{Simulations have been performed with \textit{software in the loop},
including measurement noise and communication latency.}
Communication across the system,
planning,
and sensing are implemented using the middle-ware \textit{Pocolibs} and the software framework \textit{genoM},
enabling realistic simulations.
\revision{The simulator of \textit{Fly-Crane} has been used as preliminary validation step 
	for several experiments,
	such as the ones in~\cite{75,jimenez2022precise},
	proving excellent adherence to the real testbed.
	Technical details about the simulation software and realistic numerical experiments can be found in~\cite{9213842}.}

\begin{figure}[t]
	\centering
	\includegraphics[width=0.45\textwidth]{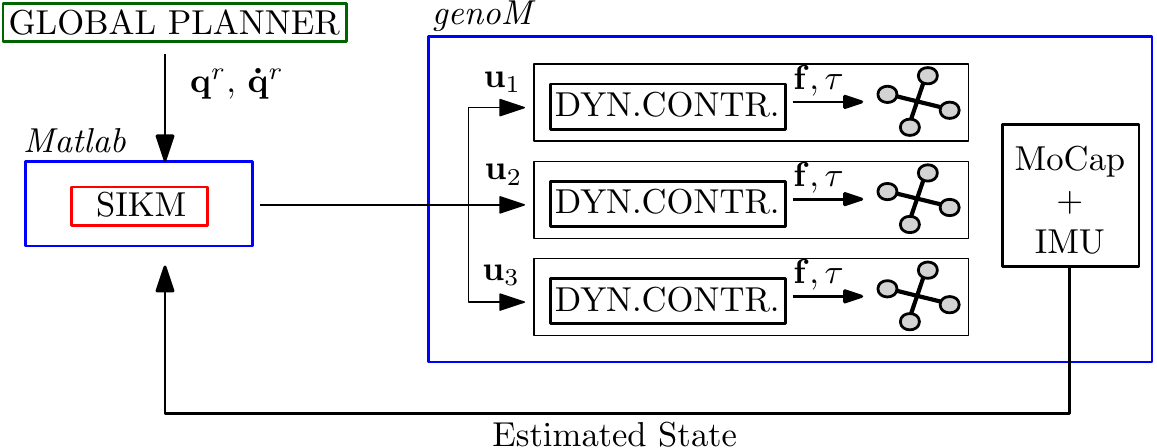}
	\caption{Architecture used to perform simulations: a global planner generates the desired trajectory $\vect{q}^r, \vect{\dot{q}}^r$ and sends it to the local planner which generates the desired robot velocities. The blue rectangle on the right represents a realistic environment where the robotic system is simulated.}
	\label{fig:scheme_global_local_planner_with_ROS}
\end{figure}

The used system architecture is represented in Fig.~\ref{fig:scheme_global_local_planner_with_ROS}:
a global planner generates offline the desired trajectory $(\vect{q}^r, \vect{\dot{q}}^r)$ and this information is made available to the SIKM controller, implemented in Matlab-Simulink.
The latter generates the desired robot velocities $\vect{u}_i$ which are sent to the robots every $T$ seconds via wireless.
Then, 
the low-level dynamical controller (geometric position controller~\cite{2010-LeeLeoMcc}) of each robot 
converts these velocities into thrust and torque for the quadrotors. 
An unscented Kalman filter, 
running at 1 [kHz], 
fuses the Motion Capture (MoCap) system measurements (at 120 [Hz]) with the IMU measurements (at 1 [kHz]). 
The estimated system state is then sent to the SIKM controller.

\begin{figure}
	\centering
	\includegraphics[width=.45\textwidth]{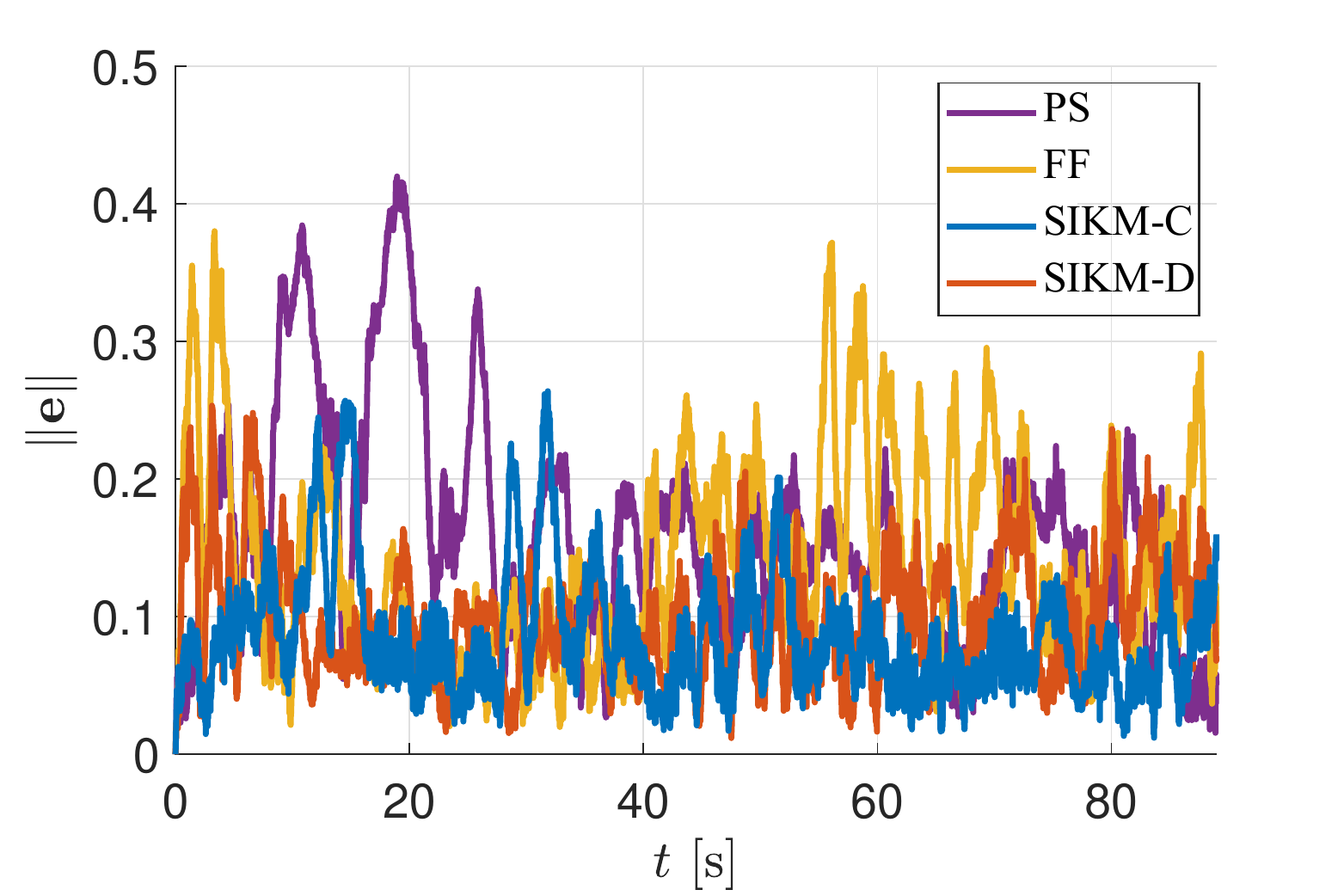}
	\caption{Tracking error norm $\|\vect{e}(t)\| = \|\vect{q}(t)-\vect{q}^r(t)\|$ obtained with the tracking strategies described in Table~\ref{tab:sim_cases} ($T=1.5$[s]).}
	\label{fig:err_norm_T_1_5s_slowTraj}
\end{figure}

\begin{figure*}[t]
	\centering
	\includegraphics[width=.83\textwidth,trim={3cm .5cm 3cm 1cm},clip]{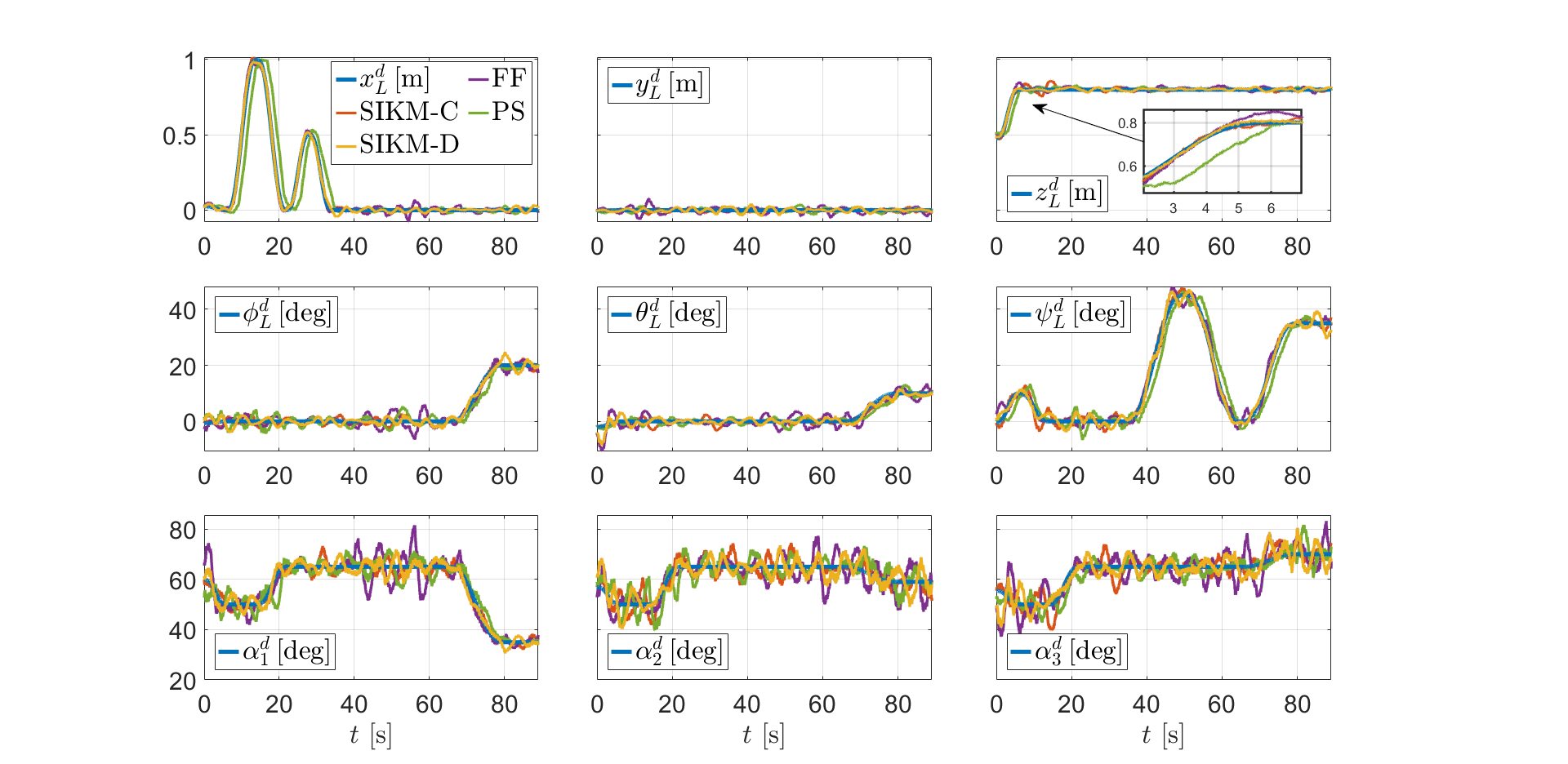}
	\caption{Comparison of the variables $\vect{q}(t)$ in four different simulations, where the tracking strategies described in Table~\ref{tab:sim_cases} are used. The sampling time is $T=1.5$[s]. The first two rows represent respectively the position $x_l,y_L,z_L$ and orientation $\phi_L,\theta_L, \psi_L$ (roll, pitch and yaw) of the load. On the bottom, row the angles $\alpha_i\overset{\text{def}}{=}q_i, \, i=1,2,3$ between cables and load are depicted. }
	\label{fig:q_T_1_5s_slowTraj}
\end{figure*}

We first compare the tracking error norm for the four control strategies when $T=1.5$ [s] 
(see Figs.~\ref{fig:err_norm_T_1_5s_slowTraj}--\ref{fig:q_T_1_5s_slowTraj}).
Note that the tracking error does not converge to zero because of non-idealities implemented in simulation, 
such as sensor noise.
The desired trajectory was generated in order to stress all components of $\vect{q}$, except for $y$-translation because of the system symmetry.
From Fig.~\ref{fig:q_T_1_5s_slowTraj}, one can see the benefits of the feedforward term w.r.t. to point-stabilization (PS), 
which is slower in tracking the desired trajectory
as emphasized in the zoom plot of the component $z_L$.
This is because the desired velocity $\vect{u}$ is updated only when a new measurement arrives. 
As for the three feedback-feedforward techniques, 
SIKM-C and SIKM-D exhibit the best performance.
This is highlighted in the zoomed plot of $z_L$, 
where FF causes an overshoot.

\begin{figure}[t]
	\centering
	\includegraphics[width=0.4\textwidth]{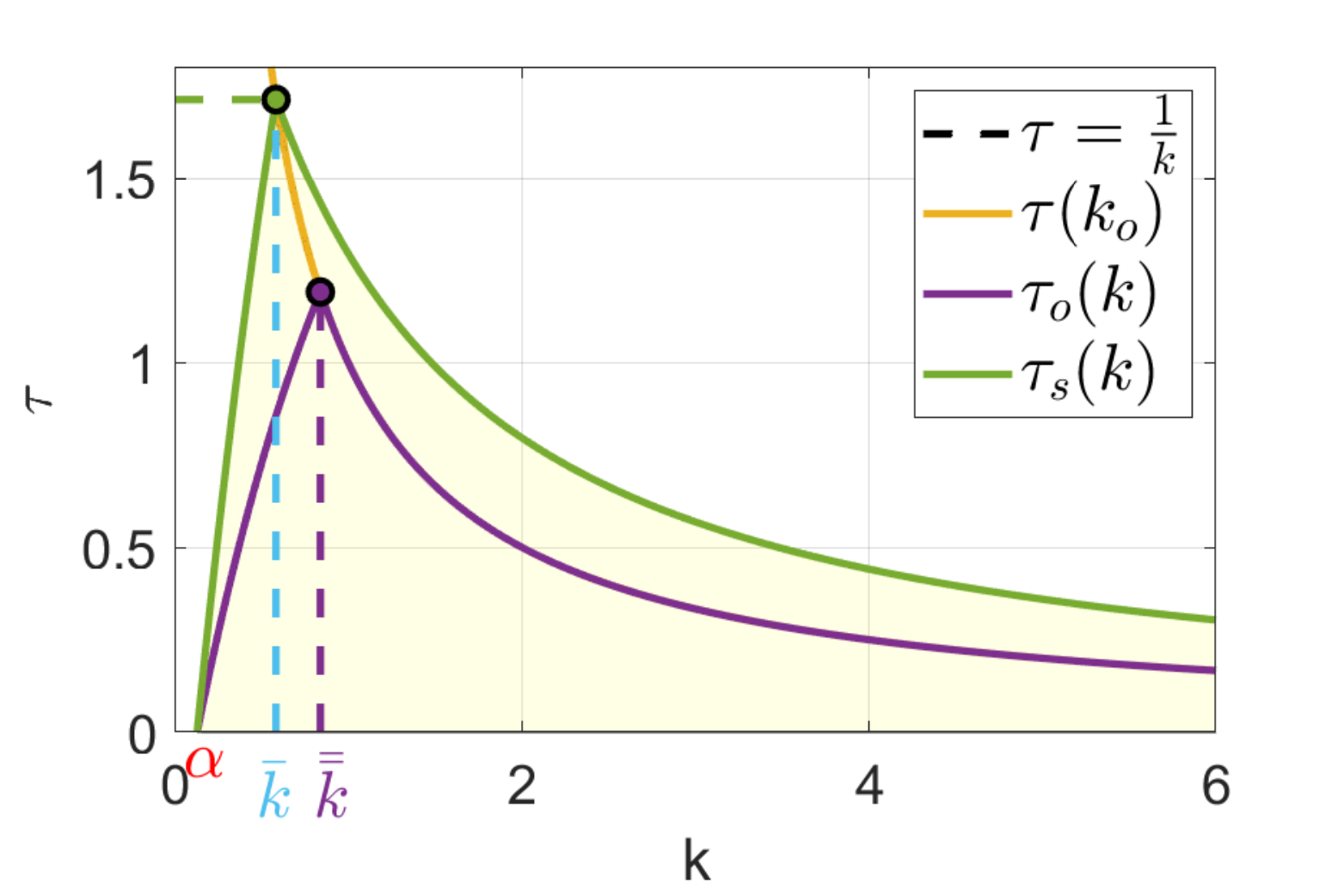}
	\caption{Stability region, optimal gain $k_o$, and sampling time $\tau_o$ obtained by estimating the parameters $\mu,\alpha,\gamma_1,\gamma_2$ along the trajectory depicted in Fig.~\ref{fig:q_T_1_5s_slowTraj} for different couples $(k,T)$.}
	\label{fig:stab_region_sim}
\end{figure}

More interesting and general is the comparison of the four strategies with different sampling times. 
To compute the feedback gain $ k_{\text{off}} $ used in the SIKM-D,
we first estimated the parameters $[\mu,\alpha,\gamma_1,\gamma_2]=[0.02, \, 0.13, \, 0.1, \, 0.2]$
along the desired trajectory depicted in Fig.~\ref{fig:q_T_1_5s_slowTraj} for different couples of $(k,T)$ as described in Section~\ref{sec:param_estim},
obtaining the stability region depicted in Fig.~\ref{fig:stab_region_sim}.
Then, 
for each value of $T$, 
we chose the gain $k$ which ensures
the highest convergence rate while keeping the system stable,
by choosing the $ x $ coordinate corresponding to $T$ on the curve $\tau(k_o)$ in Fig.~\ref{fig:stab_region_sim}.
Table~\ref{tab:sim} reports the observed mean tracking error norms for each strategy. 
The resulting feedback gain--sampling time pairs $ (k_\text{off},T) $ were $ (2,0.5) $, $ (1.28,0.75) $ and $ (0.67,1.5) $.
Point-stabilization PS yields the largest error which increases quickly with the sampling time.
All feedback-feedforward strategies are comparable for short values of $ T $,
while SIKM-C and SIKM-D yield the lowest errors for large $ T $.
However, we stress that \textit{SIKM-D is distributed}.
Hence, the results shown in Table~\ref{tab:sim} shall be intended in an even stronger way:
our proposed distributed controller not only outperforms centralized implementations of standard techniques (point-stabilization and naive feedforward),
but is even comparable with its centralized version with online gain adaptation.

\begin{table}[t]
	\centering
	\setlength\belowcaptionskip{-0.4cm}
	\begin{tabular}{ c | c | c | c | c }
		\multirow{2}{*}{Sampling time $T[\rm s]$} & \multicolumn{4}{c}{Mean tracking error norm} \\
		\cline{2-5}		 
		& SIKM-D & SIKM-C & FF & PS\\
		\hline	
		0.5 & {0.06} & 0.05 & {0.05} & 0.07\\  
		0.75 & 0.06 & 0.07 & {0.06} & 0.10\\  
		1.5 & {0.10} & 0.09 & 0.15 & 0.16 \\  
	\end{tabular}
	\caption{Mean tracking error norms obtained in simulation. 
	}
	\label{tab:sim}
\end{table}

To further validate our method under realistic conditions, 
\revision{we also performed simulations with high position measurement noise,
comparable to a GPS-based positioning system.}
As expected, the tracking error remains bounded,
proving our design robust.
Results of such additional simulations are provided in Appendix~\ref{app:simulations}.

\section{Conclusions and Future Work}\label{sec:conclusion}
In this paper, we proposed a decentralized controller for multi-robot systems
where feedback measurements are transmitted via wireless.
We showed that computing the feedforward term is nontrivial, 
and proposed a strategy to compute the feedback gain
with provable stability and convergence guarantees.
The more multi-robot systems will become popular and employed for real application, the more sampled-based control strategies will be needed. 
In this sense, this work paves the way to future developments and practical implementations.\\
A natural evolution of this work is the implementation of experiments on a real system. 
Furthermore, an analysis including packets loss and latency, which are typical issues of wireless communication, should be developed.

	\appendix

\section{Useful Lemmas}\label{app:lemmas}
\newtheorem{lemA}{Lemma}[section]
\begin{lemA}
	\label{lem:up_bound}
	Given $\vect{x} \in \mathcal{X}$, where $\mathcal{X} \subset \mathbb{R}^n$ is a compact set and given a continuously differentiable function $\vect{f(x)}:\mathbb{R}^n \rightarrow \mathbb{R}^m$ s.t. $\vect{f}(\vect{0})=\vect{0}$, then there exists $\alpha>0$ s.t.:
	\begin{align*}
	\|\vect{f(x)}\|\leq \alpha\|\vect{x}\| \ \forall \vect{x} \in \mathcal{X}.
	\end{align*}
\end{lemA}
\begin{pf}
	Since $\vect{f(x)}$ is a continuously differentiable function on a compact set, then it is Lipschitz continuous and it holds
	$$\vect{f}(\vect{x}_1-\vect{x}_2)\leq \alpha \|\vect{x}_1-\vect{x}_2\|,$$
	for some $\alpha>0$. Now let us call $\vect{x}:=\vect{x}_1$ and $\vect{x}_2=\vect{0}$ and we obtain the result of the Lemma.
\end{pf}

\begin{lemA}
	\label{lem:up_bound_2_vars}
	Given $\vect{x} \in \mathcal{X}$ and $\vect{y} \in \mathcal{Y}$, where $\mathcal{X} \subset \mathbb{R}^n$, $\mathcal{Y} \subset \mathbb{R}^p$ are compact sets and given a function that is continuously differentiable on the second argument $\vect{f(x,y)}:\mathbb{R}^n \times \mathbb{R}^p \rightarrow \mathbb{R}^m$ s.t. $\vect{f}(\vect{x},\vect{0})=\vect{0} \, \forall \vect{x} \in \mathcal{X}$, then there exists $\alpha>0$ s.t.:
	\begin{align*}
	\|\vect{f(x,y)}\|\leq \alpha\|\vect{y}\| \ \forall (\vect{x},\vect{y}) \in \mathcal{X} \times \mathcal{Y}.
	\end{align*}
\end{lemA}
\begin{pf}
	The proof is similar to the one of Lemma~\ref{lem:up_bound}: being the function continuously differentiable in the second argument, we apply that reasoning on $\vect{y}$:
	$$\|\vect{f}(\vect{x},\vect{y}_1-\vect{y}_2)\|\leq \alpha \|\vect{y}_1-\vect{y}_2\|,$$
	for some $\alpha>0$. Now let us call $\vect{y}:=\vect{y}_1$ and $\vect{y}_2=\vect{0}$ and we obtain the result of the Lemma.
\end{pf}

\begin{lemA}
	\label{lem:up_bound_2_fcn}
	Given $\vect{x} \in \mathcal{X}$, where $\mathcal{X} \subset \mathbb{R}^n$ is a compact set and given two continuously differentiable functions $\vect{f(x)}:\mathbb{R}^n \rightarrow \mathbb{R}^m$, $\vect{g(x)}:\mathbb{R}^n \rightarrow \mathbb{R}^m$ s.t. $\vect{f}(\vect{0})=\vect{g}(\vect{0})=\vect{0}$, then there exists $\alpha>0$ s.t.:
	\begin{align*}
	\|\vect{f(x)}\cdot \vect{g(x)}\|\leq \alpha\|\vect{x}\|^2 \ \forall \vect{x} \in \mathcal{X}.
	\end{align*}
\end{lemA}
\begin{pf}
	The proof is similar to the one of Lemma~\ref{lem:up_bound}. Since both the functions are continuously differentiable on a compact set, then they are also Lipschitz and there exist $a,b>0$ s.t.
	$$\|\vect{f}(\vect{x})\|\leq a\|\vect{x}\|, \quad \|\vect{g}(\vect{x})\|\leq b\|\vect{x}\|,$$
	hence
	$$\|\vect{f}(\vect{x})\cdot \vect{g}(\vect{x})\| \leq \|\vect{f}(\vect{x})\|\cdot \|\vect{g}(\vect{x})\|\leq \alpha \|\vect{x}\|^2,$$
	where $\alpha:=a b>0$ and we obtain the result of the lemma.
\end{pf}

\section{Error Bound for Point-Stabilization Control}
\label{app:point-stabilization}
In~\cite{8743454}, the authors discussed how, given a desired constant configuration $\vect{q}^r $,
if $ \vect{q}(0)\in\mathcal{B}_d(\vect{q}^r) $,
the gain $k$ can be designed offline s.t. $\vect{u}_k$ in \eqref{eq:u_k} drives the system state to $ \vect{q}^r $ exponentially fast.\\
To perform trajectory tracking, the same idea is replicated by assigning a sequence of points as a time-varying reference to the system.
In particular, the reference trajectory $ \vect{q}^r(t) $ is sampled with period $T$, and the sample points of the sequence $(\vect{q}^r(hT), \vect{\dot{q}}^r(hT))$ are given as reference configurations one after the other,
obtaining controller~\eqref{eq:u_k_hT}.
Such a strategy 
is indeed commonly used in robotic applications in virtue of its simplicity.

When $k$ is assigned offline, the tracking error is characterized as follows.
\begin{prop}\label{prop:bounded_error}
	Consider system \eqref{eq:jacobian1}--\eqref{eq:controllability} with $\vect{u}(t)$ given by~\eqref{eq:u_k_hT}.
	There exist $0<\rho<1$ and $k^*$ such that,
	if $\vect{q}(0)\in \mathcal{B}_d(\vect{q}^r(0))$,
	$\|\vect{q}^r((h+1)T)- \vect{q}^r(hT)\| \leq \beta$ for all $h$ with $ 0 < \beta < (1-\rho)d $,
	and $0<k< k^*$,
	then the tracking error $\vect{e}(hT)\overset{\text{def}}{=}\vect{q}(hT)-\vect{q}^r(hT)$ is bounded, i.e.,
	$$\|\vect{e}(hT)\|\leq \rho d + \frac{\beta}{1-\rho}, \quad \forall h\in\mathbb{N},$$ 
	and in particular
	$$\lim_{h\rightarrow+\infty}\|\vect{e}(hT)\|\leq \frac{\beta}{1-\rho}.$$
\end{prop}
\begin{pf}
	As shown in Fig.~\ref{fig:point_control_tracking}, at time $t=hT$ the system state is $\vect{q}_h = \vect{q}(hT)$ and the reference point is $\vect
	{q}^r_h$; then, at time $t=(h+1)T$, the new system state is $\vect{q}_{h+1} = \vect{q}((h+1)T)$ and a new reference $\vect
	{q}^r_{h+1}$ is assigned. 
	This strategy is applied every $T$ seconds.  
	In~\cite[Proposition~6]{8743454} the authors showed that, given a configuration $\vect{q}_h$ and the desired one $\vect{q}_h^r$,
	there exists a feedback gain s. t.
	\begin{align}\label{eq:error_decrease}
		\|\vect{e}((h+1)T)\|\leq \rho_h\|\vect{e}(hT)\|, 
	\end{align}
	where $\rho_h \in (0,1)$ is defined in~\cite[Proposition~5]{8743454} and
	depends on the set $\mathcal{B}_d(\vect{q}^r(hT))$.
	{In virtue of~Assumption~1.\ref{ass:squareInvertibleSystem:1},
	we consider $ \rho =\max_h\rho_h $ as an upper bound for the convergence rate over all the trajectory samples.}
	Equation~\eqref{eq:error_decrease} implies that the error decreases by at least a factor $ \rho $ between two consecutive sampling times.
	Further, it is shown in~\cite{8743454} that the error also decreases at each time instant $t \in (hT,(h+1)T)$.
	\begin{figure}[t]
		\centering
		{\includegraphics[width=0.4\columnwidth]{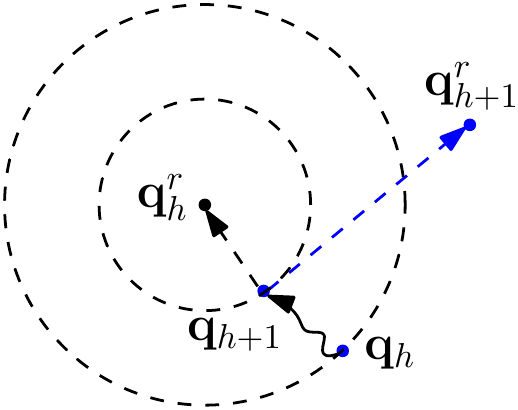}}
		\caption{Trajectory tracking by using the point stabilization strategy.}
		\label{fig:point_control_tracking}
	\end{figure}	
	Now, we define the quantities
	\begin{equation*}
		\vect{e}^+_{h+1}\overset{\text{def}}{=}\vect{q}_{h+1}-\vect{q}^r_{h+1}, \quad \vect{e}^-_{h+1}\overset{\text{def}}{=}\vect{q}_{h+1}-\vect{q}^r_{h} .\label{eq:def_eh_plus_min}
	\end{equation*}
	In words, $\vect{e}^-_{h+1}$ is the error between configuration $\vect{q}_{h+1}$ and the desired one $\vect{q}^r_{h}$,
	while $\vect{e}^+_{h+1}$ is the error between $\vect{q}_{h+1}$ and the next desired configuration $\vect{q}^r_{h+1}$.
	In view of~\eqref{eq:error_decrease} and assumption $\|\vect{q}_{h+1}^r- \vect{q}_h^r\| \leq \beta$, it holds
	\begin{align*}
		\|\vect{e}_{h+1}^-\|&=\|\vect{q}_{h+1}-\vect{q}^r_{h}\| \leq \rho\|\vect{q}_{h}-\vect{q}^r_{h}\|=\rho\|\vect{e}^+_{h}\|\\
		\|\vect{e}^+_{h+1}\|&=\|\vect{q}_{h+1}-\vect{q}^r_h+\vect{q}^r_h-\vect{q}^r_{h+1}\|\\
		&\leq \|\vect{q}_{h+1}-\vect{q}^r_h\|+\|\vect{q}^r_h-\vect{q}^r_{h+1}\|\\
		&=\|\vect{e}^-_{h+1}\|+\|\vect{q}^r_h-\vect{q}^r_{h+1}\|\leq \rho \|\vect{e}^+_{h}\|+\beta.
	\end{align*}
	{Being $ \beta < (1-\rho)d $, it follows $ \|\vect{e}^+_{h+1}\| < d $
	and therefore Proposition~6 in~\cite{8743454} can be applied s.t. the error decrease~\eqref{eq:error_decrease} holds 
	for reference point $ \vect{q}_{h+1}^r $ in an iterative fashion.}
	Now, we find an upper bound to the steady-state tracking error:
	iterating the last inequality, it holds
	\begin{align*}
		\|\vect{e}_1^+\|&\leq \rho \|\vect{e}_0^+\| +\beta\\
		\|\vect{e}_2^+\|&\leq \rho \|\vect{e}_1^+\| +\beta \leq\rho^2\|\vect{e}_0^+\|+\rho\beta+\beta\\
		&\hspace{2mm} \vdots \\
		\|\vect{e}_h^+\|&\leq \rho^h \|\vect{e}_0^+\|+\beta \sum_{i=0}^{h-1}{\rho^i}= \rho^h \|\vect{e}_0^+\| +\beta \frac{1-\rho^h}{1-\rho}.
	\end{align*}
	{The overall upper bound for the error norm can be deduced in virtue of the assumptions
	$ \|\vect{e^+_0}\| < d $ and $ \rho < 1 $. 
	In particular, at the limit it follows
	$$\lim_{h\rightarrow+\infty}\|\vect{e}_h^+\|\leq \lim_{h\rightarrow+\infty} \beta \frac{1-\rho^h}{1-\rho} = \frac{\beta}{1-\rho}. \qed$$}
\end{pf}	

\section{Proofs of Propositions~\ref{prop:tau_s}--\ref{prop:k_o_tau_fixed}}\label{app:proof_props}

\subsection{Proof of Proposition~\ref{prop:tau_s}}\label{app:proof_prop_tau_s}

Recall that the stability time $\tau_s(k)$ is s.t. $z(k,\tau)<1$ for $k>\alpha,0<\tau<\tau_s(k)$. Hence the goal now is to find an expression for $\tau_s(k)$ by studying the inequality 
$$z(k,\tau)<1,$$
From the definition of the function $z(k,\tau)$, we can distinguish three cases: $k\tau <1$, $k\tau=1$, $k\tau>1$.
\begin{enumerate}
	\item $k\tau <1$ : In this case 
	$$z(k,\tau) = z^-(k,\tau)=1+\tau(\alpha-k)+\tau^2(k^2\mu+k\gamma_1+\gamma_2).$$
	We are interested to understand when $z^-(k,\tau)<1$, that is to find the maximum $\tau$ for which $\|\vect{e}(hT+\tau)\|<\|\vect{e}(hT)\|$.
	\begin{figure}[h!]
		\centering
		\includegraphics[width=0.35\textwidth]{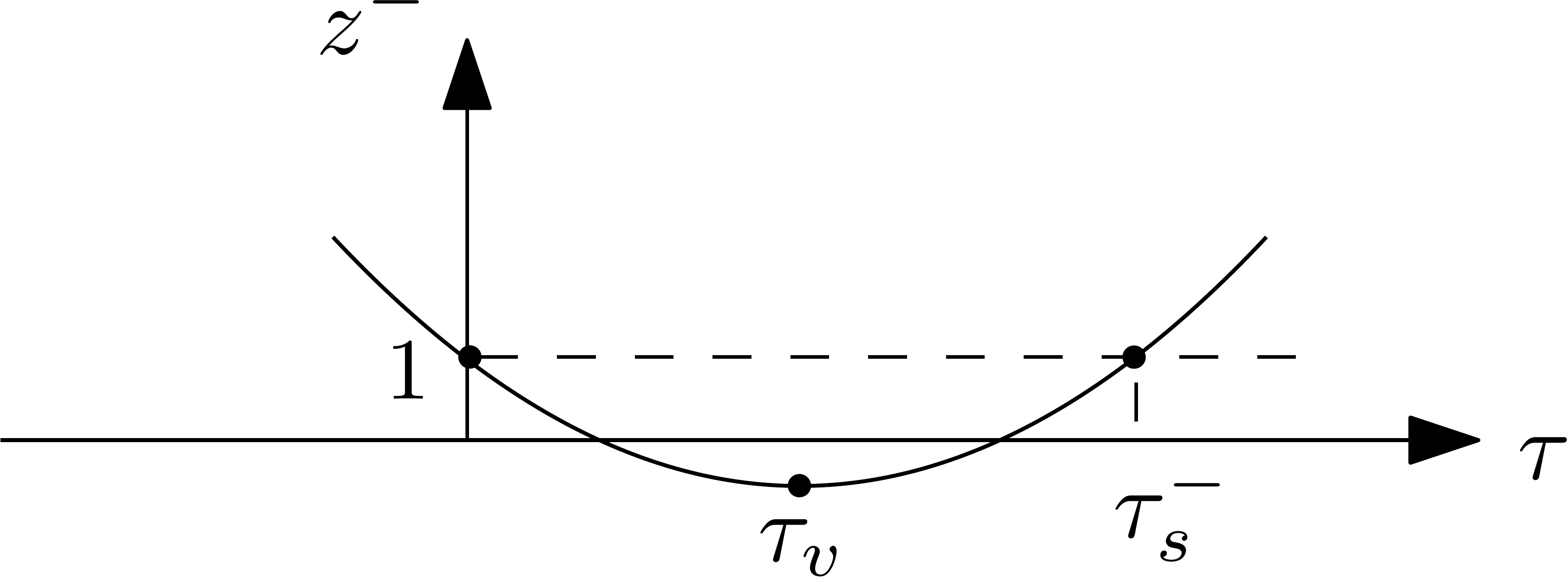}
		\caption{Representation of $z(k,\tau)$ in the case $k\tau<1$ and $k>\alpha$.}
		\label{fig:parabola_ktau_smaller_1}
	\end{figure}
	\begin{align*}
	&z^-(k,\tau)<1 \\
	&\Leftrightarrow
	1+\tau(\alpha-k)+\tau^2(k^2\mu+k\gamma_1+\gamma_2) <1\\
	&\Leftrightarrow  p_s^-(k):=\tau(\alpha-k)+\tau^2(k^2\mu+k\gamma_1+\gamma_2)<0.
	\end{align*}
	$p_s^-(k)$ represents an upward parabola with solutions
	$$\tau_{s_1}^-=0, \ \ \tau_{s_2}^-=\frac{k-\alpha}{k^2\mu+k\gamma_1+\gamma_2}.$$
	Hence $p_s^-(k)<0$ if $0<\tau<\tau_{s_2}^-$; as a consequence $z^-(k,\tau)<1$ if $0<\tau<\tau_{s_2}^-$.
	Let us define 
	$$\tau_{s}^-:=\tau_{s_2}^-.$$ 
	Now let us check when $\tau_{s}^-<\frac{1}{k}$, that is 
	\begin{align*}
	&\frac{k-\alpha}{k^2\mu+k\gamma_1+\gamma_2} < \frac{1}{k}\\
	\Leftrightarrow p_{s_2}^-(k)&:=k^2(1-\mu)-k(\alpha+\gamma_1)-\gamma_2 <0.
	\end{align*}
	The solutions are
	$$k_{s_{1,2}}^-=\frac{\alpha+\gamma_1\pm \sqrt{(\alpha+\gamma_1)^2+4\gamma_2(1-\mu)}}{2(1-\mu)}.$$
	\begin{figure}[h!]
		\centering
		\includegraphics[width=0.45\textwidth]{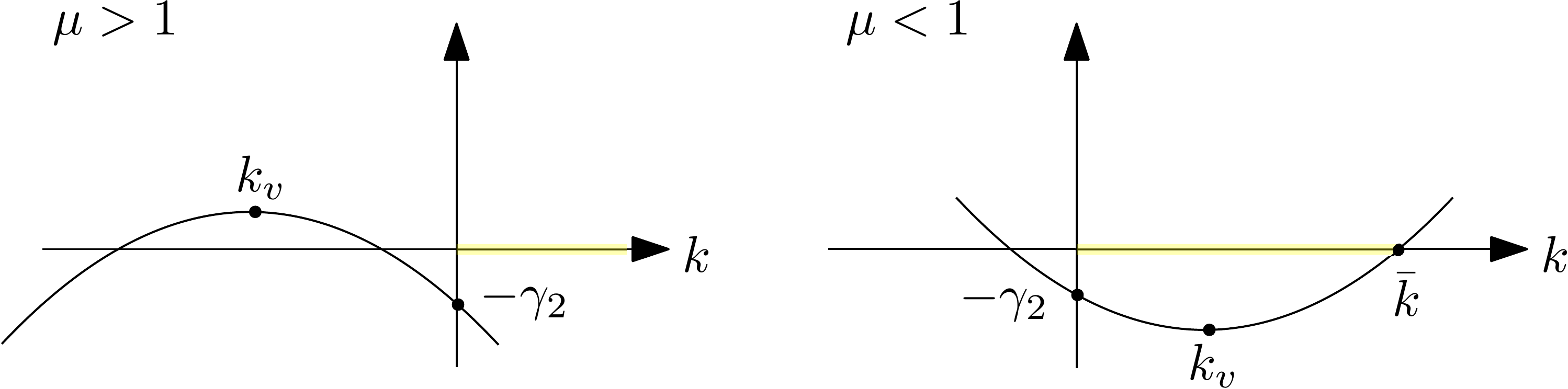}
		\caption{Representation of $p_{s_2}^-(k)$.}
		\label{fig:ps_min}
	\end{figure}
	\begin{itemize}
		\item if $\mu<1$, then $p_{s_2}^-(k)$ is an upward parabola such that the solutions and the vertex $k_v$ satisfy
		$$\sum \text{sol}=\frac{\alpha+\gamma_1}{1-\mu}>0, \quad \prod \text{sol}=\frac{-\gamma_2}{1-\mu}<0$$
		$$k_v=\frac{\alpha+\gamma_1}{2(1-\mu)}>0.$$
		Hence, if $\mu<1$, then $\tau_{s}^-<\frac{1}{k} $ for $\alpha<k<\bar{k}$, where
		\begin{align}
		&\bar{k}=\max\{k_{s_1}^-,k_{s_2}^-\}=\nonumber\\
		&\left\{\begin{array}{ll}
		\frac{\alpha+\gamma_1+ \sqrt{(\alpha+\gamma_1)^2+4\gamma_2(1-\mu)}}{2(1-\mu)} &\text{ if }\mu<1 \\
		\frac{\alpha+\gamma_1- \sqrt{(\alpha+\gamma_1)^2+4\gamma_2(1-\mu)}}{2(1-\mu)} &\text{ if }\mu>1 \\
		\end{array}\right.  .
		\label{eq:def_k_bar}
		\end{align}
		In this case $\mu<1$, hence $\bar{k}=\frac{\alpha+\gamma_1+ \sqrt{(\alpha+\gamma_1)^2+4\gamma_2(1-\mu)}}{2(1-\mu)}$.
		\item if $\mu>1$, then $p_{s_2}^-(k)$ is a downward parabola such that
		$$\sum \text{sol}<0, \quad \prod \text{sol}>0$$
		$$k_v<0.$$
		Hence, if $\mu>1$, then $\tau_{s}^-<\frac{1}{k} \ \ \forall \ \ k>\alpha$.
	\end{itemize}
	In conclusion, 
	\[
	\begin{aligned}
	z^-(k,\tau)<1 &\text{ if } 0<\tau<\tau_{s}^-\\
	\tau_s^-<\frac{1}{k} &\text{ if } \left\{\begin{array}{lr}
	\alpha < k < \bar{k}  & \text{ if }  \mu<1\\
	k > \alpha & \text{ if }  \mu>1 \\
	\end{array}\right.
	\end{aligned}
	\]
	\item $k\tau =1$ : In this case 
	\begin{align*}
	z\bigg(k,\frac{1}{k}\bigg)=\mu + \frac{1}{k}(\alpha+\gamma_1) + \frac{\gamma_2}{k^2}.
	\end{align*}
	For which values of $k$ is $z\big(k,\frac{1}{k}\big)<1$?
	\begin{align}
	&\mu + \frac{1}{k}(\alpha+\gamma_1) + \frac{\gamma_2}{k^2} < 1 \nonumber\\
	& p_s^=(k):=k^2(\mu-1)+k(\alpha+\gamma_1)+\gamma_2<0.
	\end{align}
	\begin{figure}[h!]
		\centering
		\includegraphics[width=0.45\textwidth]{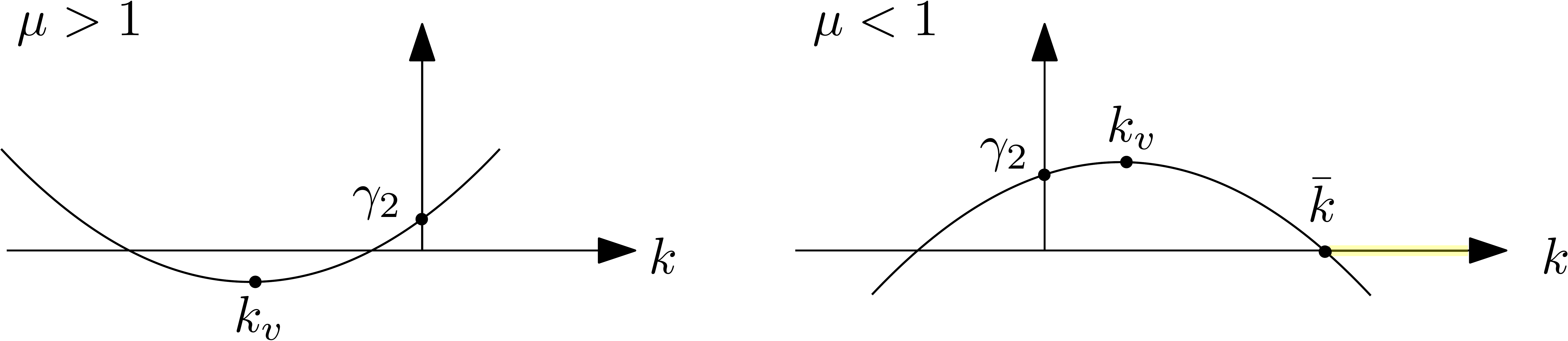}
		\caption{Representation of $p_s^=$.}
		\label{fig:p_s_eq}
	\end{figure}
	\begin{itemize}
		\item $\mu<1$: $p_s^=$ is a concave downward parabola. In $k=0$ the parabola intersects the y-axis in $\gamma_2>0$. Now the vertex of the parabola is in $k_v^=>0$, the sum of the solutions is $\sum \text{sol}>0$ and product $\prod \text{sol}<0$.
		Finally, $z\big(k,\frac{1}{k}\big)<1$ where $p_s^=(k)<0$, that is for $k>\bar k$, where
		$\bar{k}:=\max\{k_{s_{1}}^=,k_{s_{2}}^=\}$ has the same value that we have found in \eqref{eq:def_k_bar}.
		\item $\mu>1$: $p_s^=$ is a concave upward parabola. In $k=0$ the parabola intersects the y-axis in $\gamma_2>0$. The vertex of the parabola is in 
		$$k_v^==\frac{\alpha+\gamma_1}{2(1-\mu)}<0.$$
		The sum and product of the solutions are
		\begin{align*}
		\sum \text{sol}=\frac{\alpha+\gamma_1}{1-\mu}<0, & \quad \prod \text{sol}=\frac{\gamma_2}{\mu-1}>0,
		\end{align*}
		hence the solutions are both negative and $p_s^=(k)$ is always positive, so $z\big(k,\frac{1}{k}\big)>1 \ \forall k>\alpha$ if $\mu>1$.
	\end{itemize}
	In conclusion:\\
	\[z(k,1/k) < 1 \text{ for } \left\{\begin{array}{lr}
	\nexists \ \ k>\alpha & \text{ if }  \mu>1 \\
	k > \bar{k}  & \text{ if }  \mu<1
	\end{array}\right. .\]
	\item $k\tau >1$ : In this case 
	\begin{equation*}
	z\big(k,\tau \big)=z^+\big(k,\tau \big)=-1+\tau(\alpha+k)+\tau^2(k^2\gamma_1+k\gamma_2+\gamma_3).
	\end{equation*}
	This is an upward parabola with vertex
	\begin{equation*}
	\tau_v^+=-\frac{k+\alpha}{2(k^2\gamma_1+k\gamma_2+\gamma_3)}<0
	\end{equation*}
	Notice that $z^+(k,0)=-1<0$ and
	\begin{align*}
	\sum \text{sol}&=-\frac{\alpha+k}{k^2\mu+k\gamma_1+\gamma_2}<0 \\
	\prod \text{sol} &= -\frac{1}{k^2\mu+k\gamma_1+\gamma_2}<0.
	\end{align*}
	\begin{figure}[h!]
		\centering
		\includegraphics[width=0.35\textwidth]{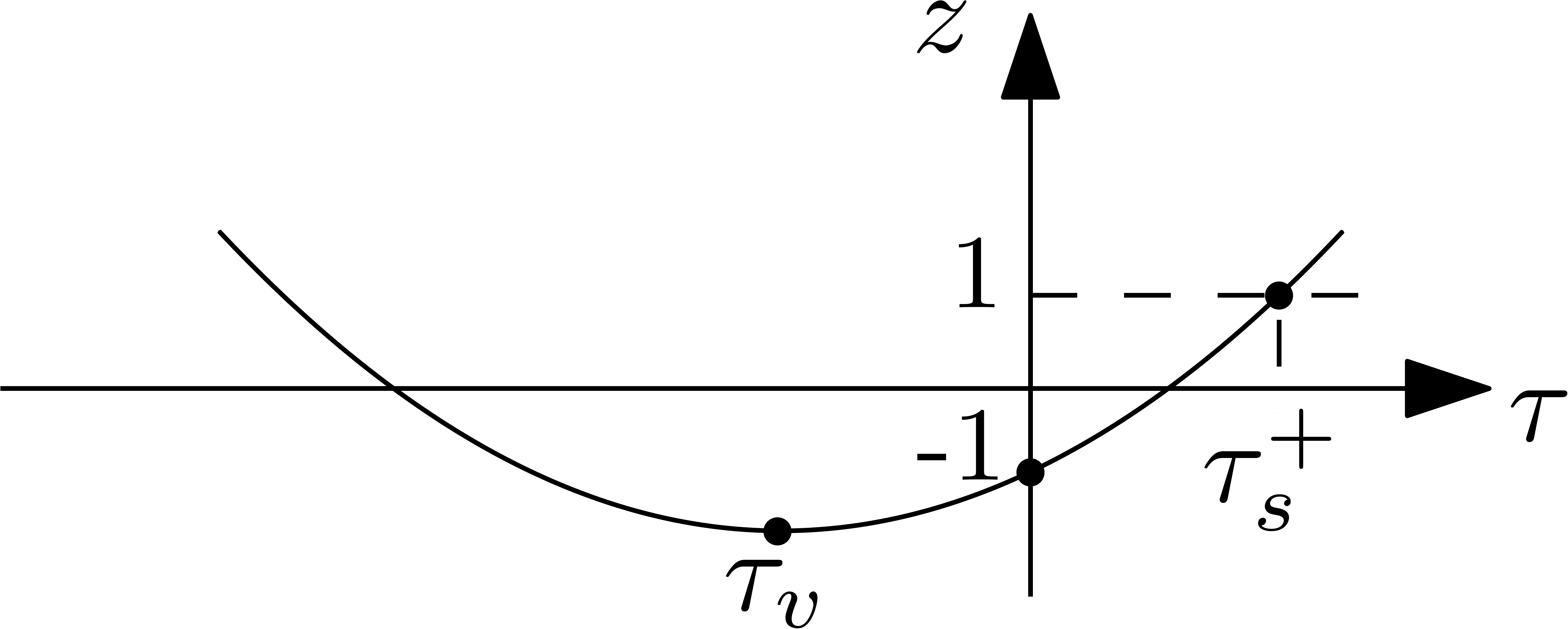}
		\caption{Representation of $z(k,\tau)$ in the case $k\tau>1$.}
		\label{fig:parabola_ktau_greater_1}
	\end{figure}
	For which values of $k$ is $z^+(k,\tau)<1$?
	\begin{align*}
	&z^+(k,\tau)<1 \\
	&\Leftrightarrow
	-1+\tau(\alpha+k)+\tau^2(k^2\mu+k\gamma_1+\gamma_2)<1\\
	&\Leftrightarrow  p_s^+(k):=\tau^2(k^2\mu+k\gamma_1+\gamma_2)+\tau(\alpha+k)-2<0.
	\end{align*}
	$p_s^+(k)$ represents an upward parabola with solutions s.t.
	\begin{align*}
	\sum \text{sol}&=-\frac{\alpha+k}{k^2\mu+k\gamma_1+\gamma_2}<0 \\
	\prod \text{sol} &= -\frac{2}{k^2\mu+k\gamma_1+\gamma_2}<0
	\end{align*}
	\begin{align*}
	\tau_{s_{1}}^+&=\frac{-(\alpha+k)-\sqrt{(\alpha+k)^2+8(k^2\mu+k\gamma_1+\gamma_2)}}{2(k^2\mu+k\gamma_1+\gamma_2)} \\ \tau_{s_{2}}^+&=\frac{-(\alpha+k)+\sqrt{(\alpha+k)^2+8(k^2\mu+k\gamma_1+\gamma_2)}}{2(k^2\mu+k\gamma_1+\gamma_2)}
	\end{align*}
	We define 
	$$\tau_{s}^+:=\max\{\tau_{s_{1,2}}^+\}=\tau_{s_2}^+$$
	and $z^+(k,\tau)<1$ for $0<\tau<\tau_{s}^+$. Notice that in this case we have to check for which $k$ it holds $\tau_{s}^+>\frac{1}{k}$ since we are analyzing the case $k\tau>1$:
	\begin{align*}
	&\tau_{s}^+	>\frac{1}{k}\\
	&\frac{-(\alpha+k)+\sqrt{(\alpha+k)^2+8(k^2\mu+k\gamma_1+\gamma_2)}}{2(k^2\mu+k\gamma_1+\gamma_2)}>\frac{1}{k}\\
	&-k(\alpha+k)+k\sqrt{(\alpha+k)^2+8(k^2\mu+k\gamma_1+\gamma_2)}>\\
	& >2(k^2\mu+k\gamma_1+\gamma_2)\\
	&k^2 ((\alpha+k)^2+8(k^2\mu+k\gamma_1+\gamma_2)) >k^2(\alpha+k)^2+\\
	&+4(k^2\mu+k\gamma_1+\gamma_2)^2+4k(\alpha+k)(k^2\mu+k\gamma_1+\gamma_2)\\
	&8k^2(k^2\mu+k\gamma_1+\gamma_2)>4(k^2\mu+k\gamma_1+\gamma_2)^2+\\
	&+4k(\alpha+k)(k^2\mu+k\gamma_1+\gamma_2)\\
	&2k^2>(k^2\mu+k\gamma_1+\gamma_2)+k(\alpha+k)\\
	&k^2(1-\mu)-k(\alpha+\gamma_1)-\gamma_2>0\\
	&p_{\tau_s^+}(k):=k^2(\mu-1)+k(\alpha+\gamma_1)+\gamma_2<0
	\end{align*}
	\begin{figure}[h!]
		\centering
		\includegraphics[width=0.45\textwidth]{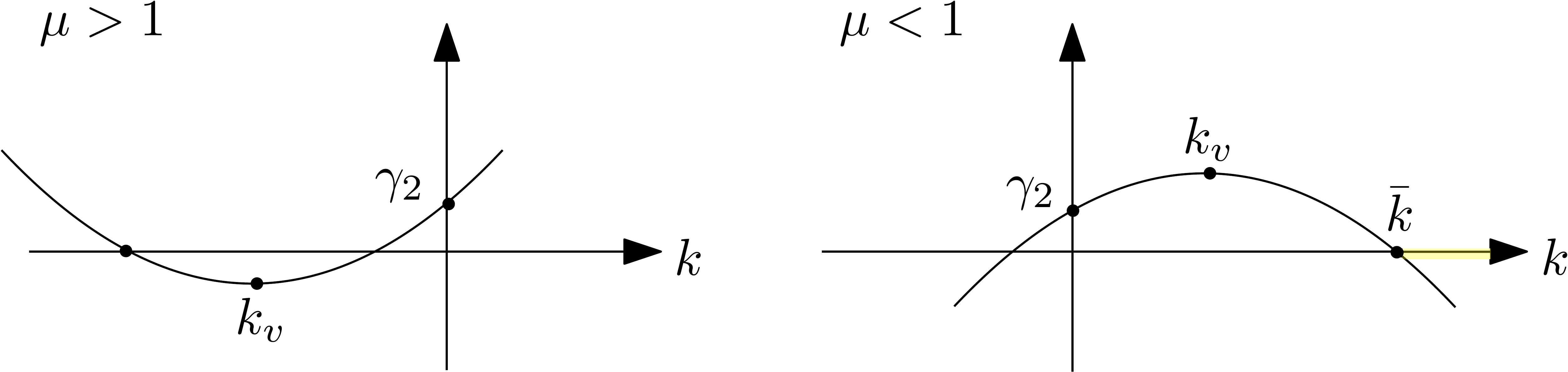}
		\caption{Representation of $p_{\tau_s^+}(k)$.}
		\label{fig:ps_plus}
	\end{figure}
	The solutions are
	\begin{align*}
	k_{\pm}=\frac{\alpha+\gamma_1\pm \sqrt{(\alpha+\gamma_1)^2+4\gamma_2(1-\mu)}}{2(1-\mu)}
	\end{align*}
	\begin{itemize}
		\item $\mu>1$: $p_{\tau_s^+}(k)$ is an upward parabola s.t. for $k=0$ it intersects the y-axis in $\gamma_2$ and 
		\begin{align*}
		\sum \text{sol}=-\frac{\alpha+\gamma_1}{\mu-1}<0, \quad \prod \text{sol} = \frac{\gamma_2}{\mu-1}<0
		\end{align*}
		As a consequence, this parabola is never negative for $k>0$, hence $\tau_{s}^+<\frac{1}{k} \ \forall k>0$.
		\item  $\mu<1$: $p_{\tau_s^+}(k)$ is a downward parabola and 
		\begin{align*}
		\sum \text{sol}>0, \quad \prod \text{sol}>0
		\end{align*}
		As a consequence, it is negative for $k>\bar{k}$, hence $\tau_{s}^+>\frac{1}{k}$ for $k>\bar{k}=\max\{k_{\pm}\}$, that is the same value found in \eqref{eq:def_k_bar}.
	\end{itemize}
	In conclusion, 
	\[
	\begin{aligned}
	z^+(k,\tau)<1 &\text{ if } 0<\tau<\tau_{s}^+\\
	\tau_s^+& \left\{\begin{array}{lr}
	<\frac{1}{k}  & \text{for } k>\alpha \quad \text{ if }  \mu>1\\
	>\frac{1}{k}  & \text{for } k>\bar{k} \quad \text{ if }  \mu<1 \\
	\end{array}\right.
	\end{aligned}
	\]
	Notice that only if $\mu<1$, then $\tau_{s}^+$ belongs to the region $k\tau>1$ that we are considering. Otherwise, we have to consider $\tau_{s}^-$ instead.
\end{enumerate}
Hence $\tau_s=\tau_s^-$ if $\mu>1$ or if $\mu<1$ and $\alpha<k<\bar{k}$, whereas $\tau_s=\tau_s^+$ if $\mu<1$ and $k>\bar{k}$, that is
\[{
	\begin{aligned}
	&\tau_s(k; \mu, \gamma_1,\gamma_2)=\frac{k-\alpha}{k^2\mu+k\gamma_1+\gamma_2} \text{ if } \mu>1\\
	&\tau_s(k; \mu, \gamma_1,\gamma_2)=\\
	&=\left\{\begin{array}{ll}
	\frac{k-\alpha}{k^2\mu+k\gamma_1+\gamma_2} &\text{ for }\alpha < k < \bar{k} \\
	\frac{-(\alpha+k)+\sqrt{(\alpha+k)^2+8(k^2\mu+k\gamma_1+\gamma_2)}}{2(k^2\mu+k\gamma_1+\gamma_2)} &\text{for }k > \bar{k} \\
	\end{array}\right. \text{ if }  \mu<1
	\end{aligned}}
\]

\subsection{Proof of Proposition~\ref{prop:tau_o_k_fixed}}\label{app:proof_prop_tau_o}
Recall that the optimal time $\tau_o(k)$ is s.t. $z(k,\tau_o(k))$ reaches its minimum value. Hence the goal now is to find an expression for $\tau_o(k)$.\\
From the definition of the function $z(k,\tau)$, we can distinguish two cases: $k\tau <1$, $k\tau>1$.
\begin{enumerate}
	\item $k\tau <1$ : In this case 
	$$z(k,\tau) = z^-(k,\tau)=1+\tau(\alpha-k)+\tau^2(k^2\mu+k\gamma_1+\gamma_2)$$
	$z(k,\tau)$ is an upward parabola (see Fig.~\ref{fig:parabola_ktau_smaller_1}) with vertex at 
	\begin{equation}
	\tau_v^-=\frac{k-\alpha}{2(k^2\mu+k\gamma_1+\gamma_2)}
	\end{equation}
	Notice that $\tau_v^->0$ since $k>\alpha$. 
	The value of the function at the vertex corresponds to its minimum:
	\begin{align*}
	z(k,\tau_v)&=1-\frac{(k-\alpha)^2}{2(k^2\mu+k\gamma_1+\gamma_2)}+\\
	&+\frac{(k-\alpha)^2}{4(k^2\mu+k\gamma_1+\gamma_2)^{\cancel{2}}}\cancel{(k^2\mu+k\gamma_1+\gamma_2)}=\\
	&=1-\frac{1}{4}\frac{(k-\alpha)^2}{(k^2\mu+k\gamma_1+\gamma_2)}<1
	\end{align*}
	Now we want to check if $\tau_v^-<\frac{1}{k}$, that is if the minimum value belongs to the region we are analyzing:
	\begin{align}
	\tau_v^-&=\frac{k-\alpha}{2(k^2\mu+k\gamma_1+\gamma_2)}<\frac{1}{k}\nonumber\\
	&\Leftrightarrow k^2-k\alpha <2\mu k^2+2\gamma_1k+2\gamma_2 \nonumber\\
	&\Leftrightarrow p_v^-(k):=k^2(1-2\mu)-k(\alpha+2\gamma_1)-2\gamma_2<0
	\end{align}
	where the solutions of $p_v^-(k)$ 
	\begin{align*}
	k_{\pm}=\frac{\alpha+2\gamma_1\pm \sqrt{(\alpha+2\gamma_1)^2+8\gamma_2(1-2\mu)}}{2(1-2\mu)}
	\end{align*}
	are s.t.
	$$\sum \text{sol}=\frac{\alpha+2\gamma_1}{1-2\mu}, \quad \prod \text{sol}=\frac{-2\gamma_2}{1-2\mu}$$
	{where there exist two different solutions if $(\alpha+2\gamma_1)^2+8\gamma_2(1-2\mu)>0$, that is
		\begin{align*}
		\mu<\frac{(\alpha+2\gamma_1)^2}{16\gamma_2}+\frac{1}{2}
		\end{align*}
		otherwise $\tau_v^-$ is always greater than $1/k$.}\\
	\begin{figure}[h!]
		\centering
		\includegraphics[width=0.45\textwidth]{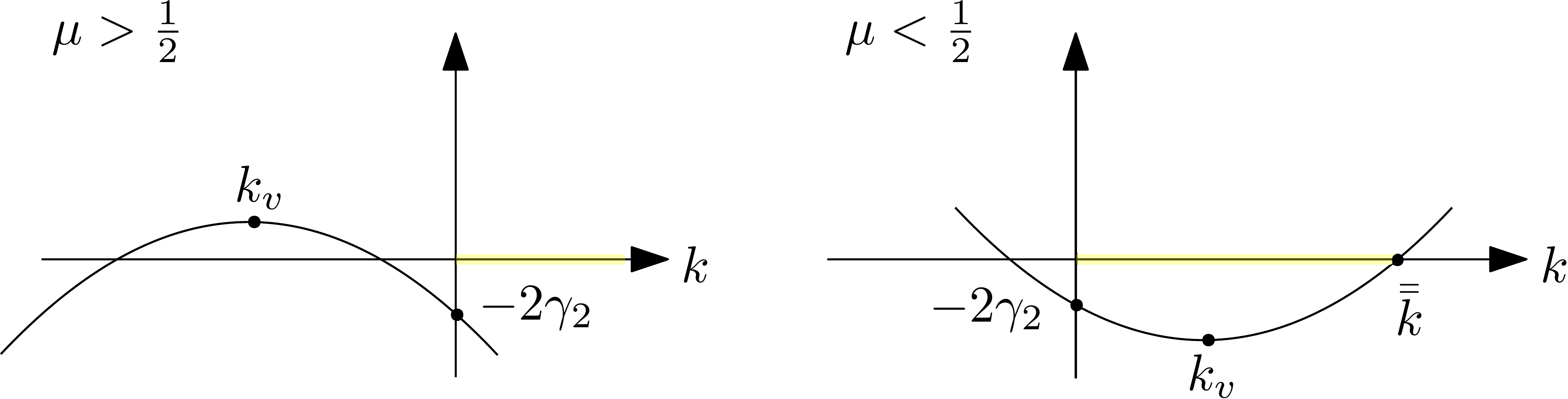}
		\caption{Representation of $p_v^-(k)$.}
		\label{fig:pv_min}
	\end{figure}
	\begin{itemize}
		\item if $1-2\mu>0$, that is $\mu<\frac{1}{2}$, then $p_v^-(k)$ is an upward parabola with solutions $k_{\pm}$ s.t. $\sum \text{sol}>0$, $\prod \text{sol}<0$ and $\tau_v<\frac{1}{k}$ if $\alpha<k<\bar{\bar{k}}$, where
		\begin{align}
		&\bar{\bar{k}}:=\max\{k_-,k_+\}=\nonumber\\
		&=\left\{\begin{array}{lcr}
		\frac{\alpha+2\gamma_1+ \sqrt{(\alpha+2\gamma_1)^2+8\gamma_2(1-2\mu)}}{2(1-2\mu)} &  \text{ if }  &\mu<\frac{1}{2}\\
		\frac{\alpha+2\gamma_1> \sqrt{(\alpha+2\gamma_1)^2+8\gamma_2(1-2\mu)}}{2(1-2\mu)} & \text{ if }  & \mu	>\frac{1}{2}
		\end{array}\right.
		\end{align}
		\item if $1-2\mu<0$, that is $\mu>\frac{1}{2}$, then $p_v^-(k)$ is a downward parabola with solutions $k_{\pm}$ s.t. $\sum \text{sol}<0$, $\prod \text{sol}>0$  and $\tau_v<\frac{1}{k}$ if $k>\alpha$.
	\end{itemize}
	In conclusion, 
	\[\tau_v^-<\frac{1}{k} \text{ if } 
	\left\{\begin{array}{lcr}
	\alpha < k < \bar{\bar{k}} &  \text{ if }  &\mu<\frac{1}{2}\\
	k > \alpha & \text{ if }  & \mu	>\frac{1}{2}
	\end{array}\right.
	\]
	So $\tau_v^-$ represents the optimal time in the case $k\tau<1$ for $\alpha < k < \bar{\bar{k}} $ if $\mu<\frac{1}{2}$ and for $k > \alpha$ if $\mu	>\frac{1}{2}$.
	\item $k\tau>1$. In this case 
	\begin{equation*}
	z\big(k,\tau \big)=z^+\big(k,\tau \big)=-1+\tau(\alpha+k)+\tau^2(k^2\gamma_1+k\gamma_2+\gamma_3)
	\end{equation*}
	we still have an upward parabola with vertex
	\begin{equation*}
	\tau_v^+=-\frac{k+\alpha}{2(k^2\gamma_1+k\gamma_2+\gamma_3)}<0
	\end{equation*}
	Since $\tau_v^+<0$ it cannot be a valid optimal time and, since $z(k,\tau)$ is a monotonically increasing function in $k$ and $\tau$, then the minimum can be detected in $\tau=1/k$.
\end{enumerate}
Finally $\tau_o(k)=\tau_v^-(k)$ for $\alpha < k < \bar{\bar{k}} $ if $\mu<\frac{1}{2}$ and for $k > \alpha$ if $\mu	>\frac{1}{2}$, whereas $\tau_o(k)=1/k$ for $k>\bar{\bar{k}}$
if $\mu<\frac{1}{2}$, that is
\[{
	\begin{aligned}
	&\tau_o(k; \mu, \gamma_1,\gamma_2)=\frac{k-\alpha}{2(k^2\mu+k\gamma_1+\gamma_2)} \text{ if } \mu>\frac{1}{2}\\
	&\tau_o(k; \mu, \gamma_1,\gamma_2)=\left\{\begin{array}{ll}
	\frac{k-\alpha}{2(k^2\mu+k\gamma_1+\gamma_2)} &\text{ for }\alpha < k < \bar{\bar{k}} \\
	\frac{1}{k} &\text{for }k > \bar{\bar{k}} \\
	\end{array}\right. \text{ if }  \mu<\frac{1}{2}
	\end{aligned}}
\]
\vspace{0.3cm}\\
\textit{Computation of the convergence rate}	\vspace{0.3cm}\\
The convergence rate, defined in \eqref{def:rho}, corresponds to the function $z(k,\tau)$ evaluated in $\tau=\tau_o(k)$: 
\begin{align*}
\rho(k)=z(k,\tau_o)
\end{align*}
We have already seen that, depending on the value of the parameter $\mu$, then $\tau_o(k)$ has different values and $z(k,\tau_o(k))$ too:
\begin{itemize}
	\item $0<\mu<\frac{1}{2}$:
	\[
	\tau_o= \left\{\begin{array}{lr}
	\tau_v^-  & \text{ if }  \alpha<k<\bar{\bar{k}}\\
	\frac{1}{k}  & \text{ if }  k>\bar{\bar{k}} \\
	\end{array}\right.
	\]
	and
	\begin{align*}
	&\rho(k)=z^-(k,\tau_o)=\\
	&=\left\{\begin{array}{lr}
	z^-(k,\tau_v^-)=1-\frac{(\alpha-k)^2}{4(k^2\mu+k\gamma_1+\gamma_2)}  & \text{ if }  \alpha<k<\bar{\bar{k}}\\
	z(k,\frac{1}{k})=\mu+\frac{1}{k}(\alpha+\gamma_1)+\frac{1}{k^2}\gamma_2  & \text{ if }  k>\bar{\bar{k}} \\
	\end{array}\right.
	\end{align*}
	Notice that $z^-(k,\tau_o)<1$ always and $z(k,\frac{1}{k})<1$ in the case it is considered ($\mu<\frac{1}{2}$). Moreover, note that for $\gamma_1=\gamma_2=0$, $z^-(k,\frac{1}{k})=\mu +\frac{\alpha}{k}$ that tends to $\mu$ for large values of $k$:
	\[
	\rho(k)\rightarrow\left\{\begin{array}{lr}
	1-\frac{1}{4\mu}  & \text{ if }  \alpha<k<\bar{\bar{k}}\\
	\mu  & \text{ if }  k>\bar{\bar{k}} \\
	\end{array}\right.
	\]
	\item $\mu>\frac{1}{2}$: in this case 
	$$\rho=z^-(k,\tau_v^-)=1-\frac{(\alpha-k)^2}{4(k^2\mu+k\gamma_1+\gamma_2)} \ \ \forall \ \ k>\alpha$$
	and for $\gamma_1=\gamma_2=0$ and big values of $k$:
	$$\rho(k)\rightarrow 1-\frac{1}{4\mu} \ \ \forall \ \ k>\alpha$$
\end{itemize}

\subsection{Proof of Proposition~\ref{prop:k_o_tau_fixed}}\label{app:proof_prop_k_o}
\begin{enumerate}
	\item $k\tau<1$:\\
	Let us rewrite the function $z^-(k,\tau)$ s.t. it depends on k, since $\tau$ is assumed to be fixed now.
	\begin{align}
	z^-(k,\tau)=\tau^2\mu k^2+\tau(\tau \gamma_1-1)k+1+\tau\alpha +\tau^2 \gamma_2
	\label{eq:zmin_tau_fixed}
	\end{align}
	The function represents an upward parabola with vertex
	\begin{align*}
	k^-_v = \frac{1-\tau \gamma_1}{2\tau \mu}>0 \text{ if } \tau<\frac{1}{\gamma_1}
	\end{align*}
	Let us check when $k^-_v<\frac{1}{\tau}$:
	\begin{align*}
	\frac{1-\tau \gamma_1}{2\tau \mu} < \frac{1}{\tau} \quad \Leftrightarrow \quad \tau >\frac{1-2\mu}{\gamma_1}
	\end{align*}
	Observe that if $\mu>\frac{1}{2}$, then $\frac{1-2\mu}{\gamma_1}<0$, hence $\tau >\frac{1-2\mu}{\gamma_1}$ always and $k^-_v<\frac{1}{\tau}$.\\
	In conclusion, 
	\[{k_v^-<\frac{1}{\tau} \text{ if } 
		\tau>\frac{1-2\mu}{\gamma_1}
	}\]
	Now we find out when $z^-(k^-_v,\tau)<1$. First of all we compute the expression $z^-(k^-_v,\tau)$:
	\begin{align*}
	z^-(k^-_v,\tau)	&=\cancel{\tau^2 \mu} \frac{(1-\tau \gamma_1)^2}{4\cancel{\tau^2}\mu^{\cancel{2}}}+\cancel{\tau}(\tau \gamma_1-1)\frac{1-\tau \gamma_1}{2\cancel{\tau}\mu}+\tau^2 \gamma_2+\\
	&+\tau \alpha+1=\\
	&= \frac{(1-\tau \gamma_1)^2}{4\mu}-\frac{(1-\tau \gamma_1)^2}{2\mu}+\tau^2 \gamma_2+\tau \alpha+1\\
	&= -\frac{(1-\tau \gamma_1)^2}{4\mu}+\tau^2 \gamma_2+\tau \alpha+1\\
	&= \frac{(-\gamma_1^2+4\gamma_2 \mu)\tau^2+2(\gamma_1+2\alpha \mu)\tau+4\mu-1}{4 \mu}
	\end{align*}
	Now, under which conditions does it hold $z^-(k^-_v,\tau)<1$?
	\begin{align}
	&\frac{(-\gamma_1^2+4\gamma_2 \mu)\tau^2+2(\gamma_1+2\alpha \mu)\tau+4\mu-1}{4 \mu}<1 \nonumber\\
	& \Leftrightarrow (-\gamma_1^2+4\gamma_2 \mu)\tau^2+2(\gamma_1+2\alpha \mu)\tau+4\mu-1<4\mu \nonumber\\
	& p^-_v(\tau):=(-\gamma_1^2+4\gamma_2 \mu)\tau^2+2(\gamma_1+2\alpha \mu)\tau-1<0 \label{eq:p_v_min_tauFixed}
	\end{align}
	$p^-_v(\tau)$ is a parabola in $\tau$ with vertex in
	\begin{align*}
	\tau^-_v=-\frac{\gamma_1+2\alpha \mu}{-\gamma_1^2 +4\gamma_2 \mu}>0 \quad \text{ if }\quad \mu < \frac{\gamma_1^2}{4 \gamma_2}
	\end{align*}
	\begin{figure}[h!]
		\centering
		\includegraphics[width=0.45\textwidth]{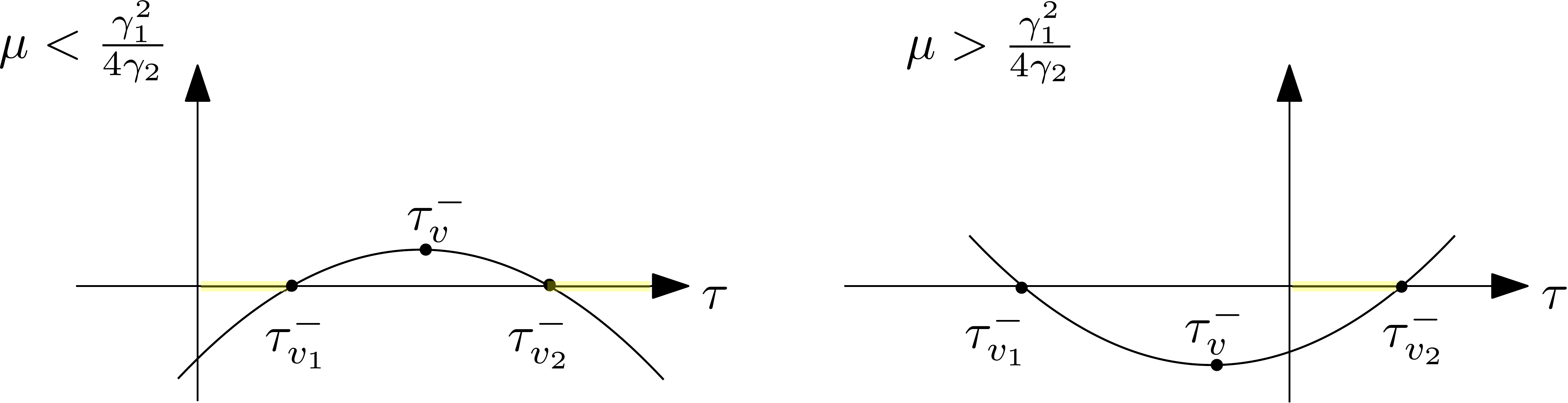}
		\caption{Representation of $p_v^-(\tau)$.}
		\label{fig:p_v_min_tauFixed}
	\end{figure}
	The solutions of \eqref{eq:p_v_min_tauFixed} are
	\begin{align*}
	\tau_{v_{1}}^-&=\frac{-(\gamma_1+\alpha \mu)- \sqrt{(\gamma_1+\alpha \mu)^2+(-\gamma_1^2+4\gamma_2\mu)}}{-\gamma_1^2+4\gamma_2\mu} \\ \tau_{v_{2}}^-&=\frac{-(\gamma_1+\alpha \mu)+ \sqrt{(\gamma_1+\alpha \mu)^2+(-\gamma_1^2+4\gamma_2\mu)}}{-\gamma_1^2+4\gamma_2\mu}
	\end{align*}
	Notice that 
	\begin{align*}
	\sum \text{sol}&=\frac{\gamma_1+\alpha \mu}{\gamma_1^2-4\gamma_2\mu}>0 \quad \text{if } \quad \mu<\frac{\gamma_1^2}{4\gamma_2}\\
	\prod \text{sol} &= \frac{1}{\gamma_1^2-4\gamma_2\mu}>0 \quad \text{if } \quad \mu<\frac{\gamma_1^2}{4\gamma_2}
	\end{align*}
	Moreover the argument of the square root is always positive and  $\tau_{v_{1}}^-<\tau_{v_{2}}^-$ if $\mu >\frac{\gamma_1^2}{4\gamma_2}$. \\
	Finally, the concavity of $p_v^-(\tau)$ is upward if $\mu>\frac{\gamma_1^2}{4\gamma_2}$, otherwise it is downward. You can see these results in Fig.~\ref{fig:p_v_min_tauFixed}.\\
	In conclusion, $z^-(k^-_v,\tau)$:
	\[
	\left\{\begin{array}{llr}
	<1  & \text{for } &0<\tau<\tau^-_{v_2} \ \  \vee \ \  \tau>\tau^-_{v_1} \quad \text{ if }  \mu<\frac{\gamma_1^2}{4\gamma_2}\\
	<1  & \text{for }& 0<\tau<\tau^-_{v_2} \quad \text{ if }  \mu>\frac{\gamma_1^2}{4\gamma_2} \\
	\end{array}\right.	
	\]
	\item ${k\tau=1}$\\
	Even if the stability analysis has been performed in the scenario where $k$ was fixed, we will analyze the stability on the curve $k=\frac{1}{\tau}$ because the result will be useful when we will compute the convergence rate.
	\begin{align*}
	z(\frac{1}{\tau},\tau)=\gamma_2\tau^2+(\alpha+\gamma_1)\tau+\mu
	\end{align*}
	Let us check when $z(\frac{1}{\tau},\tau)<1$:
	\begin{align*}
	&\gamma_2\tau^2+(\alpha+\gamma_1)\tau+\mu<1\\
	&\Leftrightarrow p^=(\tau):=\gamma_2\tau^2+(\alpha+\gamma_1)\tau+\mu-1<0
	\end{align*}
	$p^=(\tau)$ is an upward parabola with vertex
	\begin{align*}
	\tau_v^==\frac{-(\alpha+\gamma_1)}{2\gamma_2}<0
	\end{align*}
	and solutions
	\begin{figure}[h!]
		\centering
		\includegraphics[width=0.45\textwidth]{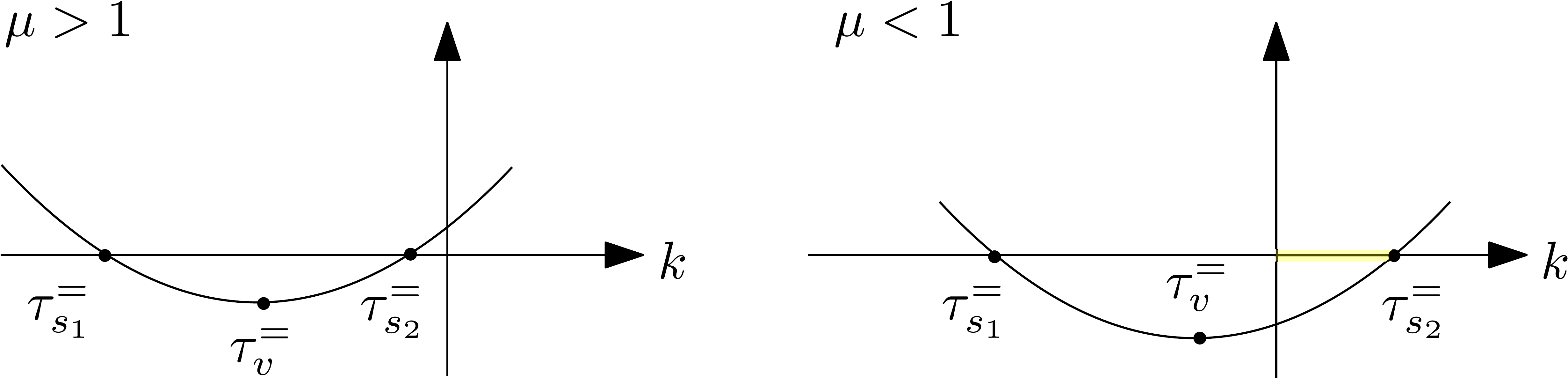}
		\caption{Representation of $p_s^=(k,\tau)$.}
		\label{fig:p_s_eq_tauFixed}
	\end{figure}
	\begin{align*}
	\tau_{s_{1,2}}^==\frac{-(\alpha+\gamma_1)\pm \sqrt{(\alpha+\gamma_1)^2-4\gamma_2(\mu-1)}}{2\gamma_2}
	\end{align*}
	s.t.
	\begin{align*}
	\sum \text{sol}&=-\frac{\alpha+\gamma_1}{\gamma_2}<0 \\
	\prod \text{sol} &= \frac{\mu-1}{\gamma_2}>0 \quad \text{if } \quad \mu>1
	\end{align*}
	In conclusion:\\
	\[z\big(\frac{1}{\tau},\tau\big) < 1 \text{ for } \left\{\begin{array}{lr}
	\nexists \ \tau>0 & \text{ if }  \mu>1 \\
	0<\tau<\tau_{s_2}^=  & \text{ if }  \mu<1
	\end{array}\right.\]
	\item ${k\tau >1}$:\\
	Let us rewrite the function $z^+(k,\tau)$ s.t. it depends on k, since $\tau$ is assumed to be fixed now.
	\begin{align}
	z^+(k,\tau)=\tau^2\mu k^2+\tau(\tau \gamma_1+1)k+\tau^2 \gamma_2 k +\tau\alpha-1
	\label{eq:zplus_tau_fixed}
	\end{align}
	The function represents an upward parabola with vertex
	\begin{align*}
	k^+_v = -\frac{1+\tau \gamma_1}{2\tau \mu}<0 \text{ for } \tau>0
	\end{align*}
	Hence $k^+_v$ never belongs to the region $k\tau>1$.
\end{enumerate}
In conclusion $k_o(\tau)=k_v^-$ for $0<\tau <\frac{1}{\gamma_1}$ if $\mu>\frac{1}{2}$ and for $\frac{1-2\mu}{\gamma_1} < \tau < \frac{1}{\gamma_1}$ if $\mu<\frac{1}{2}$; instead, $k_o(\tau)=\frac{1}{\tau}$ for $0<\tau<\frac{1-2\mu}{\gamma_1}$ if $\mu<\frac{1}{2}$, that is
\[{
	\begin{aligned}
	&k_o(\tau; \mu, \gamma_1,\gamma_2)=\frac{1-\tau \gamma_1}{2\tau \mu} \ \ \text{ for } \tau <\frac{1}{\gamma_1} \ \ \text{ if } \mu>\frac{1}{2}\\
	&k_o(\tau; \mu, \gamma_1,\gamma_2)-=\left\{\begin{array}{ll}
	\frac{1-\tau \gamma_1}{2\tau \mu} &\text{ for } \frac{1-2\mu}{\gamma_1} < \tau < \frac{1}{\gamma_1} \\
	\frac{1}{\tau} &\text{for }0<\tau<\frac{1-2\mu}{\gamma_1} \\
	\end{array}\right. \text{ if }  \mu<\frac{1}{2}
	\end{aligned}}
\]
\vspace{0.3cm}
\textit{Computation of the convergence rate}\vspace{0.3cm}\\
The convergence rate, defined in \eqref{def:rho_tau_fixed}, corresponds to the function $z(k,\tau)$ evaluated in $\tau=\tau_o(k)$: 
\begin{align*}
\rho(\tau)=z(k_o,\tau)
\end{align*}
We have already seen that, depending on the value of the parameter $\mu$, then $k_o(\tau)$ has different values and $z(k_o(\tau),\tau)$ too:
\begin{itemize}
	\item $0<\mu<\frac{1}{2}$:
	\[
	k_o= \left\{\begin{array}{lr}
	k_v^-  & \text{ for }  \frac{1-2\mu}{\gamma_1}<\tau<\frac{1}{\gamma_1}\\
	\frac{1}{\tau}  & \text{ for }  0<\tau< \frac{1-2\mu}{\gamma_1} \\
	\end{array}\right.
	\]
	and
	\begin{align*}
	&\rho(\tau)=z^-(k_o,\tau)=\\
	&=\left\{\begin{array}{lr}
	z^-(k_v^-,\tau) &  \text{ for }  \frac{1-2\mu}{\gamma_1}<\tau<\frac{1}{\gamma_1}\\
	z(\frac{1}{\tau},\tau) &  \text{ for }  0<\tau< \frac{1-2\mu}{\gamma_1} \\
	\end{array}\right.\\
	&=\left\{\begin{array}{lr}
	\frac{(-\gamma_1^2+4\gamma_2 \mu)\tau^2+2(\gamma_1+2\alpha \mu)\tau+4\mu-1}{4 \mu}  &  \text{ for }  \frac{1-2\mu}{\gamma_1}<\tau<\frac{1}{\gamma_1}\\
	\gamma_2\tau^2+(\alpha+\gamma_1)\tau+\mu  &  \text{ for }  0<\tau< \frac{1-2\mu}{\gamma_1} \\
	\end{array}\right.
	\end{align*}
	
	\item $\mu>\frac{1}{2}$: in this case 
	\begin{align*}
	&\rho(\tau)=z^-(k_v^-,\tau)=\\
	&=\frac{(-\gamma_1^2+4\gamma_2 \mu)\tau^2+2(\gamma_1+2\alpha \mu)\tau+4\mu-1}{4 \mu} \\
	& \text{ for }  \ 0<\tau<\frac{1}{\gamma_1}
	\end{align*}
\end{itemize}

\section{Online Gain Design Proposed in~\cite{8743454}}\label{app:auxiliary_system}

Let us introduce the following auxiliary system,
\begin{align}\label{eqn:NL}
	\dot{\mathbf{q}}'  (\tau;\vect{q}_h) &=- \vect{A}^{-1}_{\vect{q}'(\tau;\vect{q}_h)} \vect{A}_{\vect{q}_h} \vect{q}_h=:\vect{f}(\vect{q}'(\tau);\vect{q}_h)\\
	\vect{q}'(0;\vect{q}_h)&=\vect{q}_h; \ \  \vect{q}_h \in \mathcal{B}_d(\vect{q}^r) \nonumber,
\end{align}
with $\vect{q}_h\overset{\text{def}}{=} \vect{q}(hT)$. 
Being $\vect{f}(\vect{q}; \vect{q})=-\vect{q}$, it holds $\| \mathbf{q}'(0^+;\vect{q}_h)  - \vect{q}^r\| <  \| \vect{q}_h - \vect{q}^r \| $.
Also, we define
\begin{align}
	\tau_s(\vect{q}_h) &\overset{\text{def}}{=}
	\min_{\tau } \{\tau < 0  \, | \,  \|\vect{q}'(\tau;\vect{q}_h)\|=\|\vect{q}_h\|\},\\
	\tau_o(\vect{q}_h) &\overset{\text{def}}{=} \underset{0 \leq \tau \leq \tau_s(\vect{q}_h)}{\text{arginf}} \|\vect{q}'(\tau;\vect{q}_h)\|,
\end{align}	
where $\tau_s(\vect{q}_h)=\infty$ if $\|\vect{q}'(\tau;\vect{q}_h) - \vect{q}^r\| < \|\vect{q}_h - \vect{q}^r\|, \forall \tau$.
Then, $k_h$ is designed as (see~\cite[Proposition~7]{8743454})
\begin{equation}\label{eq:k-time-varying}
	k_h= \frac{\tau_o(\vect{q}_h)}{T}.
\end{equation}

\section{Additional Simulations}\label{app:simulations}

\begin{figure*}
	\centering
	\includegraphics[width=\linewidth]{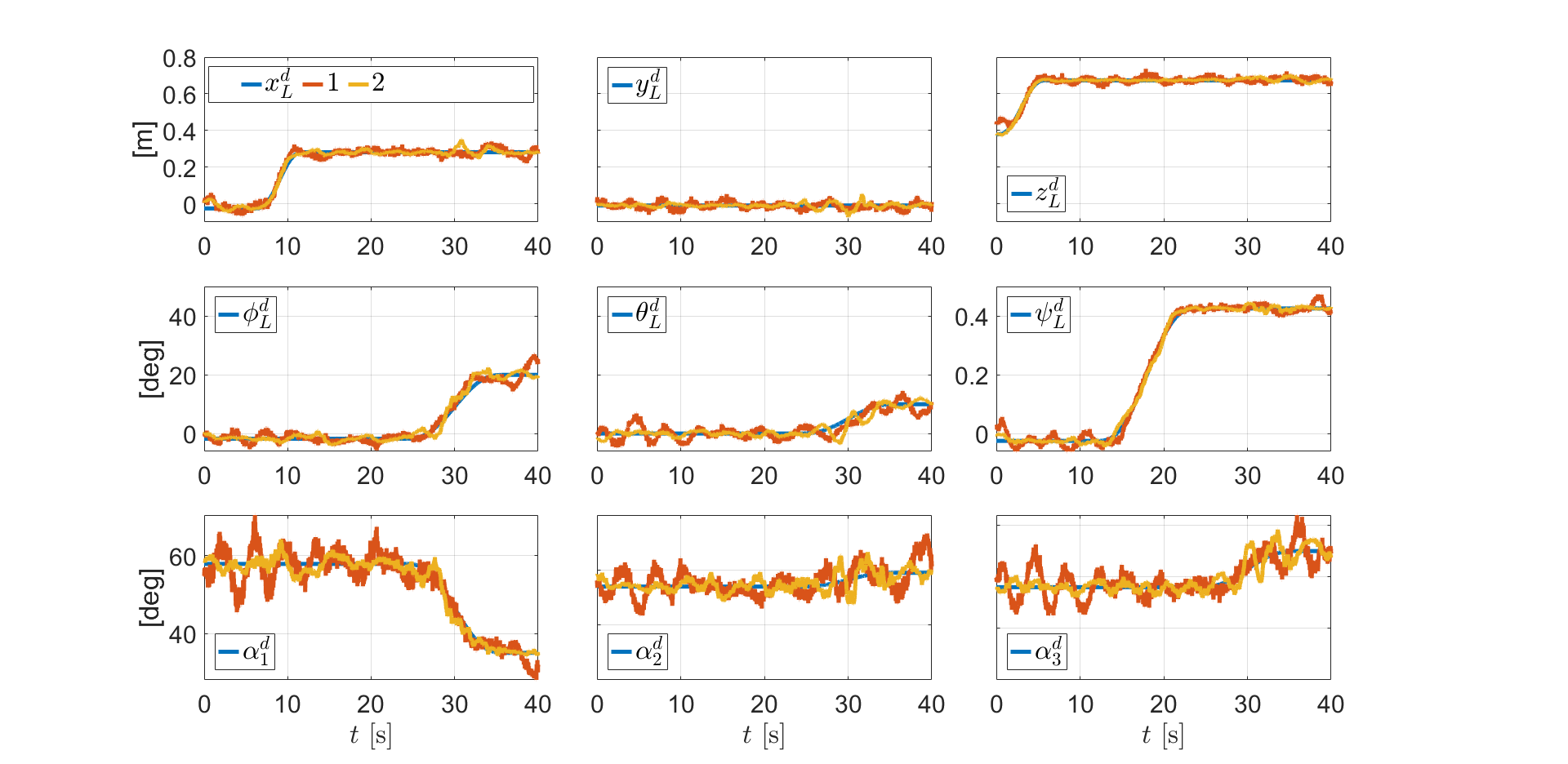}
	\caption{Comparison of the variables $\vect{q}(t)$ for two different values $\sigma_1=0.01$ and $\sigma_2=0.003$ of noise variance of the sensor vision (the values 1 and 2 in the legend refer to $\sigma_{1,2}$): the tracking strategy SIKM-D is implemented with sampling time $T=0.75$ [s] and feedback gain $ k_\text{off} = 1.28 $. The first two rows represent respectively the position $x_L,y_L,z_L$ and orientation $\phi_L,\theta_L, \psi_L$ (roll, pitch and yaw) of the load. On the last row the angles $\alpha_i\overset{\text{def}}{=}q_i, \, i=1,2,3$ between the cables and the load are depicted. Notice that the oscillations around the reference trajectory become bigger as the noise variance increases.}
	\label{fig:q_var_01_003_k1_28_T_0_72}
\end{figure*}

In Fig.~\ref{fig:q_var_01_003_k1_28_T_0_72}, we show  the behavior of the system state when increasing the noise variance of the sensor that collects data from the MoCap system. $\sigma_2 = 0.003$ is the variance used for the previous simulations, while $\sigma_1=0.01$ is the one used for this simulation. 
You can notice that, as expected, the performance become worse when increasing the noise variance and the oscillations around the reference trajectory become bigger. This result shows that our strategy can handle uncertain measurements of the state $\vect{q}$ while guaranteeing stability of the system.
	

\begin{thebibliography}{10}

\bibitem{antonelli2006kinematic}
G.~Antonelli and S.~Chiaverini.
\newblock Kinematic control of platoons of autonomous vehicles.
\newblock {\em IEEE Trans. Robot.}, 22(6):1285--1292, 2006.

\bibitem{arai2002advances}
T.~Arai, E.~Pagello, L.~E. Parker, et~al.
\newblock Advances in multi-robot systems.
\newblock {\em IEEE Trans. Robot. Autom.}, 18(5):655--661, 2002.

\bibitem{conti2015innovative}
R.~Conti, E.~Meli, A.~Ridolfi, and B.~Allotta.
\newblock An innovative decentralized strategy for i-auvs cooperative
  manipulation tasks.
\newblock {\em Robot. Auton. Syst.}, 72:261--276, 2015.

\bibitem{2017-CorEge}
J.~Cort{\'e}s and M.~Egerstedt.
\newblock Coordinated control of multi-robot systems: A survey.
\newblock {\em SICE JCMSI}, 10(6):495--503, 2017.

\bibitem{2006-DimKyr}
D.~V. Dimarogonas and K.~J. Kyriakopoulos.
\newblock Distributed cooperative control and collision avoidance for multiple
  kinematic agents.
\newblock In {\em IEEE Conf. Decis. Control}, pages 721--726. IEEE, 2006.

\bibitem{erhart2015internal}
S.~Erhart and S.~Hirche.
\newblock Internal force analysis and load distribution for cooperative
  multi-robot manipulation.
\newblock {\em IEEE Trans. Robot.}, 31(5):1238--1243, 2015.

\bibitem{farivarnejad2018stability}
H.~Farivarnejad and S.~Berman.
\newblock Stability and convergence analysis of a decentralized
  proportional-integral control strategy for collective transport.
\newblock In {\em American Control Conf.}, pages 2794--2801. IEEE, 2018.

\bibitem{7468455}
A.~Franchi and P.~R. Giordano.
\newblock Online leader selection for improved collective tracking and
  formation maintenance.
\newblock {\em IEEE Control Netw. Syst.}, 5(1):3--13, 2018.

\bibitem{doi:10.1177/0278364912462493}
A.~Franchi, C.~Masone, V.~Grabe, M.~Ryll, H.~H. Bülthoff, and P.~{Robuffo
  Giordano}.
\newblock Modeling and control of uav bearing formations with bilateral
  high-level steering.
\newblock {\em The Int. J. of Robot. Research}, 31(12):1504--1525, 2012.

\bibitem{2018q-FraPetRiz}
A.~Franchi, A.~Petitti, and A.~Rizzo.
\newblock Distributed estimation of state and parameters in multi-agent
  cooperative load manipulation.
\newblock {\em IEEE Control Netw. Syst.}, 6(2):690--701, 2019.

\bibitem{2019l-GabTogPalFra}
C.~Gabellieri, M.~Tognon, D.~Sanalitro, L.~Palottino, and A.~Franchi.
\newblock A study on force-based collaboration in swarms.
\newblock {\em Swarm Intelligence}, 14:57--82, 2020.

\bibitem{hayakawa2006formation}
T.~{Hayakawat}, T.~{Matsuzawat}, and S.~{Harat}.
\newblock Formation control of multi-agent systems with sampled information.
\newblock In {\em IEEE Conf. Decis. Control}, pages 4333--4338, 2006.

\bibitem{jimenez2022precise}
A.~Jim{\'e}nez-Cano, D.~Sanalitro, M.~Tognon, A.~Franchi, and J.~Cort{\'e}s.
\newblock Precise cable-suspended pick-and-place with an aerial multi-robot
  system.
\newblock {\em J. Intell. Robot. Syst.}, 105(3):1--13, 2022.

\bibitem{2010-LeeLeoMcc}
T.~Lee, M.~Leoky, and N.~H. McClamroch.
\newblock Geometric tracking control of a quadrotor {UAV} on {SE}(3).
\newblock In {\em IEEE Conf. Decis. Control}, pages 5420--5425, Atlanta, GA,
  Dec. 2010.

\bibitem{liu2016formation}
Z.~Liu, W.~Chen, J.~Lu, H.~Wang, and J.~Wang.
\newblock Formation control of mobile robots using distributed controller with
  sampled-data and communication delays.
\newblock {\em IEEE Trans. Control Syst. Technol.}, 24(6):2125--2132, 2016.

\bibitem{2013-ManDevRosCor}
M.~Manubens, D.~Devaurs, L.~Ros, and J.~Cort\'es.
\newblock Motion planning for {6-D} manipulation with aerial towed-cable
  systems.
\newblock In {\em Robot.: Sci. Syst.}, Berlin, Germany, May 2013.

\bibitem{7759262}
C.~{Masone}, H.~H. {Bülthoff}, and P.~{Stegagno}.
\newblock Cooperative transportation of a payload using quadrotors: A
  reconfigurable cable-driven parallel robot.
\newblock In {\em IEEE/RSJ Int. Conf. Intell. Robots Syst.}, pages 1623--1630,
  Oct 2016.

\bibitem{2010-MazKonBerOll}
I.~Maza, K.~Kondak, M.~Bernard, and A.~Ollero.
\newblock {Multi-UAV} cooperation and control for load transportation and
  deployment.
\newblock {\em J. Intell. Robot. Syst.}, 57(1-4):417--449, 2010.

\bibitem{2013-MelShoMicKum}
D.~Mellinger, M.~Shomin, N.~Michael, and V.~Kumar.
\newblock Cooperative grasping and transport using multiple quadrotors.
\newblock In {\em Distr. Auton. Robot. Syst.}, pages 545--558. Springer, 2013.

\bibitem{pereira2004decentralized}
G.~A. Pereira, M.~F. Campos, and V.~Kumar.
\newblock Decentralized algorithms for multi-robot manipulation via caging.
\newblock {\em Int. J. Rob. Res.}, 23(7-8):783--795, 2004.

\bibitem{9213842}
A.~Petitti, D.~Sanalitro, M.~Tognon, A.~Milella, J.~Cortés, and A.~Franchi.
\newblock Inertial estimation and energy-efficient control of a cable-suspended
  load with a team of uavs.
\newblock In {\em Int. Conf. Unmanned Aircraft Syst.}, pages 158--165, 2020.

\bibitem{ren2020fully}
Y.~Ren, S.~Sosnowski, and S.~Hirche.
\newblock Fully distributed cooperation for networked uncertain mobile
  manipulators.
\newblock {\em IEEE Trans. Robot.}, 36(4):984--1003, 2020.

\bibitem{2021arXiv211201107R}
E.~{Rossi}, M.~{Tognon}, R.~{Carli}, A.~{Franchi}, and L.~{Schenato}.
\newblock {Control of over-redundant cooperative manipulation via sampled
  communication}.
\newblock {\em arXiv e-prints}, page arXiv:2112.01107, Dec. 2021.

\bibitem{8743454}
E.~{Rossi}, M.~{Tognon}, R.~{Carli}, L.~{Schenato}, J.~{Cortés}, and
  A.~{Franchi}.
\newblock Cooperative aerial load transportation via sampled communication.
\newblock {\em IEEE Contr. Syst. Lett.}, 4(2):277--282, 2020.

\bibitem{2013-SabSecChoGas}
L.~Sabattini, C.~Secchi, N.~Chopra, and A.~Gasparri.
\newblock Distributed control of multirobot systems with global connectivity
  maintenance.
\newblock {\em IEEE Trans. Robot.}, 29(5):1326--1332, 2013.

\bibitem{75}
D.~Sanalitro, H.~J. Savino, M.~Tognon, J.~Cort{\'e}s, and A.~Franchi.
\newblock Full-pose manipulation control of a cable-suspended load with
  multiple uavs under uncertainties.
\newblock {\em IEEE Robot. Autom. Lett.}, 5:2185--2191, January 2020.

\bibitem{5959179}
M.~{Schwager}, B.~J. {Julian}, M.~{Angermann}, and D.~{Rus}.
\newblock Eyes in the sky: Decentralized control for the deployment of robotic
  camera networks.
\newblock {\em Proc. IEEE}, 99(9):1541--1561, 2011.

\bibitem{2009-SicSciVilOri}
B.~Siciliano, L.~Sciavicco, L.~Villani, and G.~Oriolo.
\newblock {\em Robotics: Modelling, Planning and Control}.
\newblock Springer, 2009.

\bibitem{sieber2018human}
D.~Sieber and S.~Hirche.
\newblock Human-guided multirobot cooperative manipulation.
\newblock {\em IEEE Trans. Control Syst. Technol.}, 27(4):1492--1509, 2019.

\bibitem{simetti2016manipulation}
E.~Simetti and G.~Casalino.
\newblock Manipulation and transportation with cooperative underwater vehicle
  manipulator systems.
\newblock {\em IEEE J. Ocean. Eng.}, 42(4):782--799, 2016.

\bibitem{2013-SreKum}
K.~Sreenath and V.~Kumar.
\newblock Dynamics, control and planning for cooperative manipulation of
  payloads suspended by cables from multiple quadrotor robots.
\newblock In {\em Robot.: Sci. Syst.}, Berlin, Germany, June 2013.

\bibitem{stramigioli2005sampled}
S.~Stramigioli, C.~Secchi, A.~J. van~der Schaft, and C.~Fantuzzi.
\newblock Sampled data systems passivity and discrete port-hamiltonian systems.
\newblock {\em IEEE Trans. Robot.}, 21(4):574--587, 2005.

\bibitem{doi:10.1177/0278364919854131}
A.~Tagliabue, M.~Kamel, R.~Siegwart, and J.~Nieto.
\newblock Robust collaborative object transportation using multiple mavs.
\newblock {\em Int. J. Rob. Res.}, 38(9):1020--1044, 2019.

\bibitem{2018h-TogGabPalFra}
M.~Tognon, C.~Gabellieri, L.~Pallottino, and A.~Franchi.
\newblock Aerial co-manipulation with cables: The role of internal force for
  equilibria, stability, and passivity.
\newblock {\em IEEE Robot. Autom. Lett., Spec. Issue Aerial Manip.}, 3(3):2577
  -- 2583, 2018.

\bibitem{tsiamis2015cooperative}
A.~Tsiamis, C.~K. Verginis, C.~P. Bechlioulis, and K.~J. Kyriakopoulos.
\newblock Cooperative manipulation exploiting only implicit communication.
\newblock In {\em IEEE/RSJ Int. Conf. Intell. Robots Syst.}, pages 864--869.
  IEEE, 2015.

\bibitem{verginis2018communication}
C.~K. Verginis, A.~Nikou, and D.~V. Dimarogonas.
\newblock Communication-based decentralized cooperative object transportation
  using nonlinear model predictive control.
\newblock In {\em European Control Conf.}, pages 733--738. IEEE, 2018.

\bibitem{2016-WaSch}
Z.~Wang and M.~Schwager.
\newblock Force-amplifying n-robot transport system (force-ants) for
  cooperative planar manipulation without communication.
\newblock {\em Int. J. Rob. Res.}, 35(13):1564--1586, 2016.

\end{thebibliography}
\end{document}